  \RequirePackage{fix-cm}
  \documentclass[twocolumn]{svjour3}          % twocolumn
  \smartqed  % flush right qed marks, e.g. at end of proof
  \usepackage{graphicx}
  \usepackage[authoryear]{natbib}
  \usepackage{tabu,multirow}
  \usepackage[leftcaption]{sidecap}
  \usepackage{amsmath}
  \usepackage{amssymb}
  \usepackage{color}
  \usepackage{enumitem}
  \usepackage{verbatim}
  \usepackage[colorlinks,citecolor=blue]{hyperref}
  
  \usepackage{graphicx}
  \usepackage{subfigure}
  \usepackage{float}  % forge figure position
  \usepackage{textcomp}

  \usepackage{booktabs}

  \usepackage{eso-pic}
  \usepackage{xspace}
  \makeatletter
  \DeclareRobustCommand\onedot{\futurelet\@let@token\@onedot}
  \def\@onedot{\ifx\@let@token.\else.\null\fi\xspace}

  \makeatother
  
  \usepackage{mathtools}

  \usepackage{multirow}
  \usepackage{array}
  \usepackage[ruled,linesnumbered]{algorithm2e}

\begin{document}
  \sloppy
  
  \title{Towards Task Sampler Learning for Meta-Learning}
  
  %%%%
  % \subtitle{Do you have a subtitle?\\ If so, write it here}
  
  % \titlerunning{Short form of title}        % if too long for running head
  
  \author{ Jingyao~Wang      \and
           Wenwen~Qiang \and
           Xingzhe~Su \and
           Changwen~Zheng \and
           Fuchun~Sun \and
           Hui~Xiong
  }
  
 \institute{Jingyao Wang, Wenwen Qiang, Xingzhe Su, Changwen Zheng \at
              National Key Laboratory of Space Integrated Information System, Institute of Software Chinese Academy of Sciences, Beijing, China; University of Chinese Academy of Sciences, Beijing, China\\
           \and
           Fuchun Sun \at
              National Key Laboratory of Space Integrated Information System, Institute of Software Chinese Academy of Sciences, Beijing, China; Department of Computer Science and Technology, Tsinghua University, Beijing, China\\
          \and
          Hui Xiong \at
              Thrust of Artificial Intelligence, Hong Kong University of Science and Technology, Guangzhou, China;
              Department of Computer Science and Engineering, the Hong Kong University of Science and Technology, Hong Kong SAR, China\\
          \and    
          Corresponding author: Wenwen Qiang, \email{qiangwenwen@iscas.ac.cn}
}

\date{Received: date / Accepted: date}

\def\ourconv{RIConv++\xspace}
\def\smallgap{\vspace{0.05in}}
  
  \maketitle

  \begin{abstract}
Meta-learning aims to learn general knowledge with diverse training tasks conducted from limited data, and then transfer it to new tasks. It is commonly believed that increasing task diversity will enhance the generalization ability of meta-learning models. However, this paper challenges this view through empirical and theoretical analysis. We obtain three conclusions: (i) there is no universal task sampling strategy that can guarantee the optimal performance of meta-learning models; (ii) over-constraining task diversity may incur the risk of under-fitting or over-fitting during training; and (iii) the generalization performance of meta-learning models are affected by task diversity, task entropy, and task difficulty. Based on this insight, we design a novel task sampler, called Adaptive Sampler (ASr). ASr is a plug-and-play module that can be integrated into any meta-learning framework. It dynamically adjusts task weights according to task diversity, task entropy, and task difficulty, thereby obtaining the optimal probability distribution for meta-training tasks. Finally, we conduct experiments on a series of benchmark datasets across various scenarios, and the results demonstrate that ASr has clear advantages. The code is publicly available at~\href{https://github.com/WangJingyao07/Adaptive-Sampler}{https://github.com/WangJingyao07/Adaptive-Sampler}.

  \keywords{Meta-Learning \and Task Sampling \and Few-Shot Learning \and Transfer Learning \and Bi-Level Optimization}
  \end{abstract}

%%%%%%%%%%%%%%%%%%%%%%%%%%%%%%%%%%%%%%%%%%%%%%%%%%%%%%%%%%%%%%%%%%%%%%%%%%%%%%%%%%%%%%%%%%%%%%%%%%%
%                                         1
%%%%%%%%%%%%%%%%%%%%%%%%%%%%%%%%%%%%%%%%%%%%%%%%%%%%%%%%%%%%%%%%%%%%%%%%%%%%%%%%%%%%%%%%%%%%%%%%%%%
\section{Introduction}
\label{sec:1}
Meta-learning, also known as “learning to learn”, enables machines to learn general priors with limited supervision and quickly adapt to new scenarios \citep{vanschoren2018meta}. It can effectively learn rich knowledge from a series of tasks when data acquisition is difficult or expensive \citep{rajeswaran2019meta}. Therefore, meta-learning is considered a feasible solution for few-shot learning problems \citep{parnami2022learning}. It mainly adopts a bi-level optimization process: it learns general priors, i.e., meta-parameters, by adjusting task-specific parameters in the inner loop and minimizing the average loss of multiple tasks in the outer loop.

The optimization process of meta-learning requires a series of randomly sampled tasks. They are divided into training tasks and test tasks, with the same number of classes but different class labels. The distribution of all tasks is uniform, and the task sampler is called Uniform sampler. Recent studies have shown that the task distribution in the meta-training stage \citep{wang2020generalizing} has an important impact on model performance. Therefore, it is necessary to restrict the task sampler to meet some priors of meta-learning. The traditional view is that the higher diversity of training tasks will lead to a model with better generalization ability \citep{jamal2019task, hospedales2021meta}. This assumption seems reasonable \citep{wang2020generalizing}, because learning from diverse types of data can help acquire more knowledge, which is also consistent with human cognition.

\begin{figure*}
    \centering
    \includegraphics[width=\textwidth]{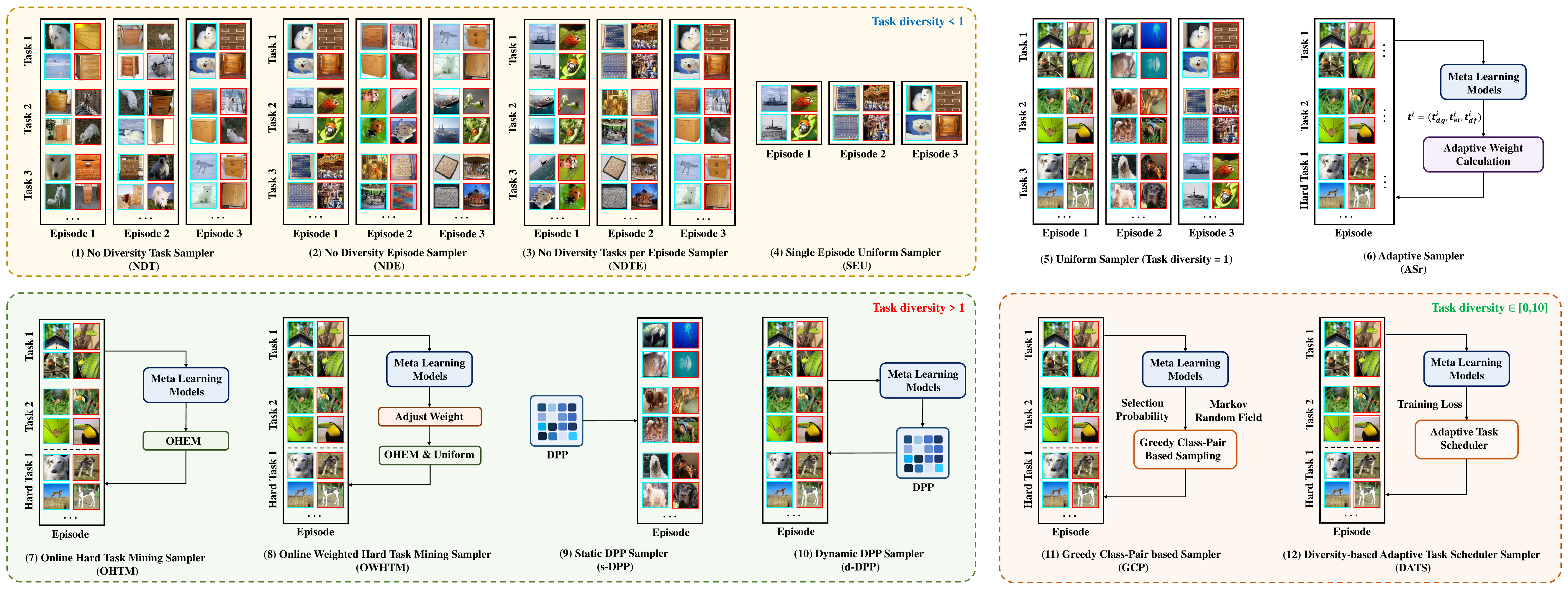}
    \caption{Task Samplers. These twelve samplers are divided into five categories, i.e., (i) standard sampler: Uniform sampler; (ii) low-diversity task samplers: NDT, NDE, NDTE, and SEU; (iii) high-diversity task samplers: OHTM, OWHTM, s-DPP, and d-DPP; (iv) previously proposed adaptive task samplers: GCP and DATS; and (v) our proposed Adaptive Sampler (ASr). More details are shown in Section \ref{sec:3.3}.}
    \label{fig:1}
\end{figure*}

To verify the reliability of this view, we conduct nine task samplers with different diversity levels as shown in Figure \ref{fig:1}. We divide them into three groups based on their task diversity scores as shown in Table \ref{tab:1score}. Next, we use MAML \citep{finn2017model} and ProtoNet \citep{snell2017prototypical} as backbones and evaluate their performance after applying different task samplers on two benchmark datasets, i.e., miniImagenet \citep{vinyals2016matching} and Omniglot \citep{lake2019omniglot}. Contrary to expectations, the results shown in Table \ref{table:intro} demonstrate that increasing task diversity does not significantly enhance performance, and even limiting task diversity leads to better results.

To further examine whether there are common requirements for task diversity, we conduct experiments on various meta-learning problems, including standard few-shot learning, cross-domain few-shot learning, multi-domain few-shot learning, etc. As shown in Figure \ref{fig:2}, we find that not only does task diversity fail to bring consistent improvements to all datasets, but over-constraining task diversity may lead to under-fitting or over-fitting during training. For example, in cross-domain few-shot learning, the training curve with a low-diversity sampler, NDTB, converges before reaching the average converge value, reflecting under-fitting; and the training curve with a high-diversity sampler, OHTM, has an inflection point where the model performance rises and then falls, reflecting over-fitting. The results provided in Tables \ref{table:1}-\ref{table:multi-domain} also support the above conclusions. Therefore, the traditional view that increasing task diversity can improve model generalization is limited. 
Meanwhile, we introduce the previously proposed adaptive task samplers for evaluation \citep{liu2020adaptive,yao2021meta}, which also rely on task diversity for sampling. Despite achieving great improvement compared to Uniform sampler, they are difficult to perform well on all datasets while incurring large computational overhead.
Thus, we cannot only rely on task diversity to find the optimal sampling strategy.

To find a universal optimal sampling strategy, we try to explore more reasonable measurements to evaluate the quality of meta-learning tasks. Through theoretical analysis, we reveal that a high-quality task should have four properties: intra-class compaction, inter-class separability, feature space enrichment, and causal invariance. Based on this insight, we propose three measurements: task diversity, task entropy, and task difficulty. To validate their effectiveness, we construct theoretical and empirical analyses. The results demonstrate that the proposed three measurements can well reflect the properties required for high-quality tasks, and each measurement is indispensable.

Based on the above analysis, we propose the Adaptive Sampler (ASr) for episodic meta-learning. ASr is a distribution generation function implemented using a multilayer perceptron. It takes task diversity, task entropy, and task difficulty as inputs, and dynamically weights the randomly sampled tasks to obtain the optimal distribution for any meta-learning model. To optimize ASr, we propose a simple and general meta-learning algorithm. Our contributions can be summarized as follows:
\begin{itemize}
    \item We draw three conclusions through theoretical and empirical analysis: (i) there is no universal task sampling strategy to ensure the performance of meta-learning models; (ii) over-constraining task diversity can lead to either under-fitting or over-fitting during training; and (ii) the meta-learning model's generalization ability is affected by task diversity, task entropy, and task difficulty.
    \item We propose ASr, an adaptive sampler that adjusts the weights of tasks in a training episode according to task diversity, task entropy, and task difficulty. ASr is a plug-and-play module that can be applied to any meta-learning scenario and framework.
    \item We present a simple and general meta-learning algorithm that uses the model’s performance in a single iteration as the feedback signal to update ASr’s parameters for the next iteration. This optimization mechanism can incorporate prior constraints with higher weights for high-quality tasks and lower weights for low-quality tasks into meta-learning.
    \item We conduct extensive experiments on various meta-learning settings, and the results demonstrate the superiority and compatibility of our ASr. We also perform ablation studies and further analysis to understand how ASr works.
\end{itemize}

%%%%%%%%%%%%%%%%%%%%%%%%%%%%%%%%%%%%%%%%%%%%%%%%%%%%%%%%%%%%%%%%%%%%%%%%%%%%%%%%%%%%%%%%%%%%%%%%%%%
%                                         2
%%%%%%%%%%%%%%%%%%%%%%%%%%%%%%%%%%%%%%%%%%%%%%%%%%%%%%%%%%%%%%%%%%%%%%%%%%%%%%%%%%%%%%%%%%%%%%%%%%%
\section{Related Work}
\label{sec:2}
Meta-learning enables a model to learn how to adapt to new tasks quickly with only a few training samples, instead of requiring complete retraining from scratch. This technique addresses the problems that have limited or challenging data \citep{vanschoren2018meta, hospedales2021meta}. Meta-learning methods can be classified into three main categories: optimization-based, metric-based, and Bayesian-based methods.

\begin{table}
\begin{center}
\caption{Task diversity scores of task samplers. The calculation of task diversity scores is illustrated in Section \ref{sec:3.2}. The score of Uniform sampler is normalized to 1, and the other samplers' scores are processed based on Uniform sampler.}
\label{tab:1score}
\begin{tabular}{l|c}
\toprule[1.2pt]
\textbf{Samplers}& \textbf{Diversity}\\
\midrule
\textbf{NDT sampler} & 0.00 \\
\textbf{NDE sampler} & 0.00 \\
\textbf{NDTE sampler} & $\approx $ 0.00 \\
\textbf{SEU sampler} & 0.00 \\
\midrule
\textbf{Uniform sampler} & 1.00 \\
\midrule
\textbf{OHTM sampler} & 1.71 \\
\textbf{OWHTM sampler} & 2.36 \\
\textbf{s-DPP sampler} & 12.43 \\
\textbf{d-DPP sampler} & 12.86 \\
\midrule
\textbf{GCP sampler} & $ \left [ 0.65,9.82 \right]$ \\
\textbf{DATS sampler} & $ \left [ \approx 0.00,8.01 \right]$ \\
\midrule
\textbf{Adaptive sampler} & $ \left [ 0,13.97 \right]$ \\
\bottomrule
\end{tabular}
\end{center}
\end{table}

\begin{table}
\begin{center}
\caption{The effects of different samplers on MAML \citep{finn2017model} and ProtoNet \citep{snell2017prototypical}). The $\downarrow $ and $\uparrow $ indicate that the result using the corresponding sampler is lower or higher than using Uniform Sampler. The best results are highlighted in \textbf{bold}.} 
\label{table:intro}
% \resizebox{\linewidth}{!}{
\begin{tabular}{l|l|c|c}
\toprule[1.2pt]
& \textbf{Samplers}& \textbf{MAML} &  \textbf{ProtoNet}\\
\midrule
\multirow{5}{*}{\rotatebox{90}{miniImagenet}}
& NDE & \textbf{49.34}$\uparrow $ & 47.86$\uparrow $ \\
& SEU & 39.45$\downarrow $ & 42.14$\downarrow $ \\
& Uniform \textbf{*} & 49.27\textbf{*} & 47.29\textbf{*} \\
& OHTM & 47.14$\downarrow $ & \textbf{47.89}$\uparrow $ \\
& s-DPP & 48.83$\downarrow $ & 46.30$\downarrow $ \\
\midrule
\multirow{5}{*}{\rotatebox{90}{Omniglot}}
& NDE & 98.42$\downarrow $ & 97.37$\downarrow $ \\
& SEU & 93.91$\downarrow $ & 95.78$\downarrow $ \\
& Uniform \textbf{*} & \textbf{98.98}\textbf{*} & \textbf{97.93}\textbf{*} \\
& OHTM & 98.14$\downarrow $ & 97.42$\downarrow $ \\
& s-DPP & 98.55$\downarrow $ & 96.78$\downarrow $ \\
\bottomrule
\end{tabular}
\end{center}
\end{table}

Optimization-based meta-learning methods learn a set of optimal initialization parameters that enable a model to quickly converge when learning new tasks. Classic methods include MAML \citep{finn2017model}, Reptile \citep{nichol2018reptile}, and MetaOptNet \citep{lee2019meta}. MAML trains a model that can adapt to various tasks by sharing the initial parameters across different tasks and performing multiple gradient updates \citep{abbas2022sharp,jeong2020ood}. Reptile also shares initial parameters, but uses an approximate update strategy, i.e., fine-tuning the model through multiple iterations to approach the optimal parameters. MetaOpt focuses on how to choose the optimizer and learning rate to quickly adapt to new tasks, without involving direct adjustment of model parameters. Recently, more optimization-based meta-learning methods have been proposed to solve various problems, such as fault diagnosis \citep{feng2022semi} and catastrophic forgetting \citep{chi2022metafscil}.

Metric-based meta-learning methods learn embedding functions that map instances from different tasks, where they can be easily classified by non-parametric methods. This idea has been explored by various methods that differ in how they learn the embedding functions and how they define the similarity or distance metrics in the feature space. In chronological order, the influential methods include Siamese Network \citep{koch2015siamese}, Matching Network \citep{vinyals2016matching}, Prototypical Network \citep{snell2017prototypical,cheng2022holistic}, Relation Network \citep{sung2018learning}, and Graph Neural Network \citep{hospedales2021meta}. Among them, Siamese Network \citep{chen2021exploring} maximizes the similarity between two augmentations of one instance to perform non-parametric learning, Graph Neural Network \citep{gao2023survey} aims to study meta-learning with the prism of inference on a partially observed graphical model, while the other three conventional techniques \citep{vinyals2016matching,lang2023base,sung2018learning,lang2022learning} and their variations \citep{zhu2022convolutional,hu2022unsupervised,bartler2022mt3,lang2023few} generally aim to establish a metric space where classification can be performed by calculating distances to prototype representations.

Bayesian-based meta-learning methods use conditional probability as the basis of meta-learning computations. The classic Bayesian-based methods include CNAPs \citep{zhang2021shallow} and Simple CNAPS (SCNAP) \citep{bateni2020improved}. They learn a feature extractor whose parameters are modulated by an adaptation network that takes the current task’s dataset as input. There are few meta-learning methods in this category compared to the aforementioned two categories, but still have some achievements in recent years. BOOM \citep{grant2018recasting} uses Gaussian processes to model the objective function and acquisition functions, which guide the search for optimal meta-parameters. VMGP \citep{myers2021bayesian} uses Gaussian processes with learned deep kernel and mean functions to model the predictive label distribution for each task.

Note that some meta-learning methods proposed task diversity-based sampling strategies \citep{kumar2022effect,liu2020adaptive} to conduct tasks in each episode and achieve good performance in various environments. However, they only focused on how to sample more diverse tasks, and the rationale behind this sampling strategy is often overlooked who simply adopt the current setting without questioning it. The training process is treated as a black box, where high performance is celebrated, but a deeper understanding of this phenomenon is lacking. Moreover, there are numerous meta-learning strategies that may interact with different sampling mechanisms. The previously proposed methods used fixed strategies and only focused on feature space enrichment to sample tasks, which we proved limited in Section \ref{sec:4.1} and Section \ref{sec:8}. Therefore, in this paper, we take data distribution as the entry point, investigate how tasks affect model performance, and attempt to find a general optimal sampling strategy.

%%%%%%%%%%%%%%%%%%%%%%%%%%%%%%%%%%%%%%%%%%%%%%%%%%%%%%%%%%%%%%%%%%%%%%%%%%%%%%%%%%%%%%%%%%%%%%%%%%%
%                                         3
%%%%%%%%%%%%%%%%%%%%%%%%%%%%%%%%%%%%%%%%%%%%%%%%%%%%%%%%%%%%%%%%%%%%%%%%%%%%%%%%%%%%%%%%%%%%%%%%%%%
\section{Preliminaries}
\label{sec:3}
In this section, we present the background of task sampling for meta-learning. We first introduce the foundations of episodic meta-learning. Then, we illustrate the calculation of task diversity scores mentioned in Table \ref{tab:1score}. Finally, we introduce the details of the constructed task samplers shown in Figure \ref{fig:1}.

%%%%%%%%%%%%%%%%%%%%%%%%%%%%%%%%%%
%            3.1
%%%%%%%%%%%%%%%%%%%%%%%%%%%%%%%%%%
\subsection{Episodic Meta-Learning}
\label{sec:3.1}
Meta-learning aims to learn a model $f$ that can quickly adapt to new tasks with limited data. Formally, we have a meta-training dataset $\mathcal{D}_{train} $ and a meta-testing dataset $\mathcal{D}_{test} $. The classes for the two datasets are designated as $\mathcal{C}_{train} = \{1, \dots, c_{tr}\}$ and $\mathcal{C}_{test} = \{c_{tr+1}, \dots, c_{tr+te}\}$, respectively, with no overlap. Existing meta-learning methods rely on the episodic training paradigm. Each episode contains $N^{pool}$ randomly selected tasks, denoted as $\mathcal{T}=\left \{ \mathcal{T}_{i} \right \}_{i=1}^{N^{pool}}$. The data in task $\mathcal{T}_{i}$ is then split into a support set $\mathcal{D}^{s}_{i}$ and a query set $\mathcal{D}^{q}_{i}$, thus, we have ${\mathcal{T}_i} = \{ {\mathcal{D}_i^s,\mathcal{D}_i^q} \}$. % The selection of the classes in each task $\mathcal{T}_i$ is uniform and random.

Meta-learning can be seen as a bi-level optimization problem. In the inner loop, the task-specific model $f^{*}_i$ is learned with the support set $\mathcal{D}^{s}_{i}$ using the meta-learning model $f$. In the outer loop, $f$ is updated by minimizing the average loss across multiple tasks with the query sets $\mathcal{D}^{q}$ using the learned task-specific models. The objective can be organized as follows:
\begin{equation}
    \begin{array}{l}
\mathop {\min }\limits_f \sum\limits_{i = 1,{\mathcal{T}_i} \in \mathcal{T}}^{N^{pool}} {\mathcal{L}(\mathcal{D}_i^q,{f^ *_i })} \\[10pt]
s.t.\quad f^ *_i = f - \lambda {\nabla _f}{\mathcal{L}(\mathcal{D}_i^s,f)} 
\end{array}
\end{equation}
where $\lambda$ is the learning rate during the optimization phase, $\mathcal{L}(\cdot)$ is the training loss.

%%%%%%%%%%%%%%%%%%%%%%%%%%%%%%%%%%
%            3.2
%%%%%%%%%%%%%%%%%%%%%%%%%%%%%%%%%%

\subsection{Task Diversity}
\label{sec:3.2}
The concept of task diversity in meta-learning is heuristic and intuitive, and few works provide a clear definition. In this study, we introduce an interesting measurement (T-DPP) that uses Determinantal Point Process (DPP) to quantify task diversity \citep{kumar2022effect}. T-DPP represents task diversity as the square of the parallelepiped volume spanned by task feature vectors, which is calculated using the determinant of the matrix formed by these feature vectors \citep{daw2023matrix}. The more diverse the task is, the more dissimilar these vectors are to each other, resulting in larger angles between the feature vectors and larger volumes of the parallelepipeds they form.

According to T-DPP, we first define the candidate pool of tasks in each episode as ${\mathcal{T}}=\{ {{\mathcal{T}_1},...,{\mathcal{T}_{N^{pool}}}} \}$, where $N^{pool}$ is the total number of tasks. Then, we encode each task $\mathcal{T}_i$ with a pre-trained ProtoNet model, and obtain a set of feature vectors that form a matrix $\mathcal{M} _{\mathcal{T}_i}\in {\mathbb{R}^{d \times d}}$ and $\mathcal{M} _{\mathcal{T}_i}=\Psi _{\mathcal{T}_i}^\mathrm {T}\Psi _{\mathcal{T}_i}$, where $\Psi _{\mathcal{T}_i}$ is the matrix corresponds to the feature vectors of $\mathcal{T}_i$, $d$ and $N_i$ represent the length of the feature vectors and the number of samples in the task $\mathcal{T}_i$, respectively. Then, for task $\mathcal{T}_i$, its diversity score $S_{td}$ is:
\begin{equation}
\label{equDPP}
    S_{td}(\mathcal{T}_i)= det(\mathcal{M} _{\mathcal{T}_i}) =vol^{2}(\Psi _{\mathcal{T}_i} ) 
\end{equation}
where a higher diversity score is achieved when the feature vectors corresponding to the task are more orthogonal, resulting in a larger squared volume (\emph{vol}) of the parallelepiped spanned by $\Psi _{\mathcal{T}_i}$. 

T-DPP can measure task diversity in different modes, i.e., class, task, and episode, by obtaining the matrix corresponding to the feature vectors \citep{kumar2022effect}. Table \ref{tab:1score} lists the task diversity scores of the task samplers shown in Figure \ref{fig:1}. We use the average score of the tasks obtained from ten samplings as the diversity score of the task sampler, and set the score of the Uniform sampler to 1 for numerical processing.

%%%%%%%%%%%%%%%%%%%%%%%%%%%%%%%%%%
%            3.3
%%%%%%%%%%%%%%%%%%%%%%%%%%%%%%%%%%
\subsection{Task Samplers}
\label{sec:3.3}
We introduce eleven task samplers for evaluation in this study, as shown in Figure \ref{fig:1}. These samplers are conducted following \citep{kumar2022effect, shrivastava2016training,liu2020adaptive,yao2021meta}. According to the task diversity scores shown in Table \ref{tab:1score}, we classify these samplers into four categories: standard task sampler, low-diversity task sampler, high-diversity task sampler, and previously proposed adaptive task sampler. The details are illustrated as follows.

The standard task sampler is Uniform sampler that is used in most meta-learning methods (see Figure \ref{fig:1} (5)). It creates a new task by sampling classes uniformly without favoring a specific type of task.

The low-diversity task sampler is the sampler whose task diversity score is lower than Uniform sampler (see Figure \ref{fig:1} (1)-(4)), including: (i) No Diversity Task Sampler (NDT), which samples one set of tasks with the same classes for episode one and propagate these tasks in all other episodes; (ii) No Diversity Episode Sampler (NDE), which samples one set of tasks with the different classes for episode one and propagate these tasks in all other episodes; (iii) No Diversity Tasks per Episode Sampler (NDTE), which samples different sets of tasks for different episodes, and the classes of the tasks in the same episode are same; and (iv) Single Episode Uniform Sampler (SEU), which samples one task for each episode, and the sampled tasks are different.

The high-diversity task sampler is the sampler whose task diversity score is higher than Uniform sampler (see Figure \ref{fig:1} (7)-(10)), including: (i) Online Hard Task Mining Sampler (OHTM), which samples tasks based on online hard example mining (OHEM); (ii) Online Weighted Hard Task Mining Sampler (OWHTM), which samples tasks by adjusting the ratio of OHEM-based sampling and Uniform sampling; (iii) Static DPP Sampler (s-DPP), which samples tasks based on Determinantal Point Process (DPP) before training; and (iv) Dynamic DPP Sampler (d-DPP), which also samples tasks based on DPP but resamples tasks at each episode dynamically instead of fixing them.

The precious proposed adaptive task sampler uses an adaptive strategy to sample tasks (see Figure \ref{fig:1} (11)-(12)). These samplers also only rely on task diversity for sampling and their task diversity scores are mainly above 1. This type of samplers include (i) Greedy Class-Pair based sampler (GCP) \citep{liu2020adaptive}, which samples difficult tasks using a greedy strategy; and (ii) Diversity-based Adaptive Task Scheduler Sampler (DATS), which samples tasks based on the training loss following \citep{yao2021meta}. 

\begin{figure*}[t]
     \centering
     \subfigure[Standard]{\includegraphics[width=0.245\textwidth]{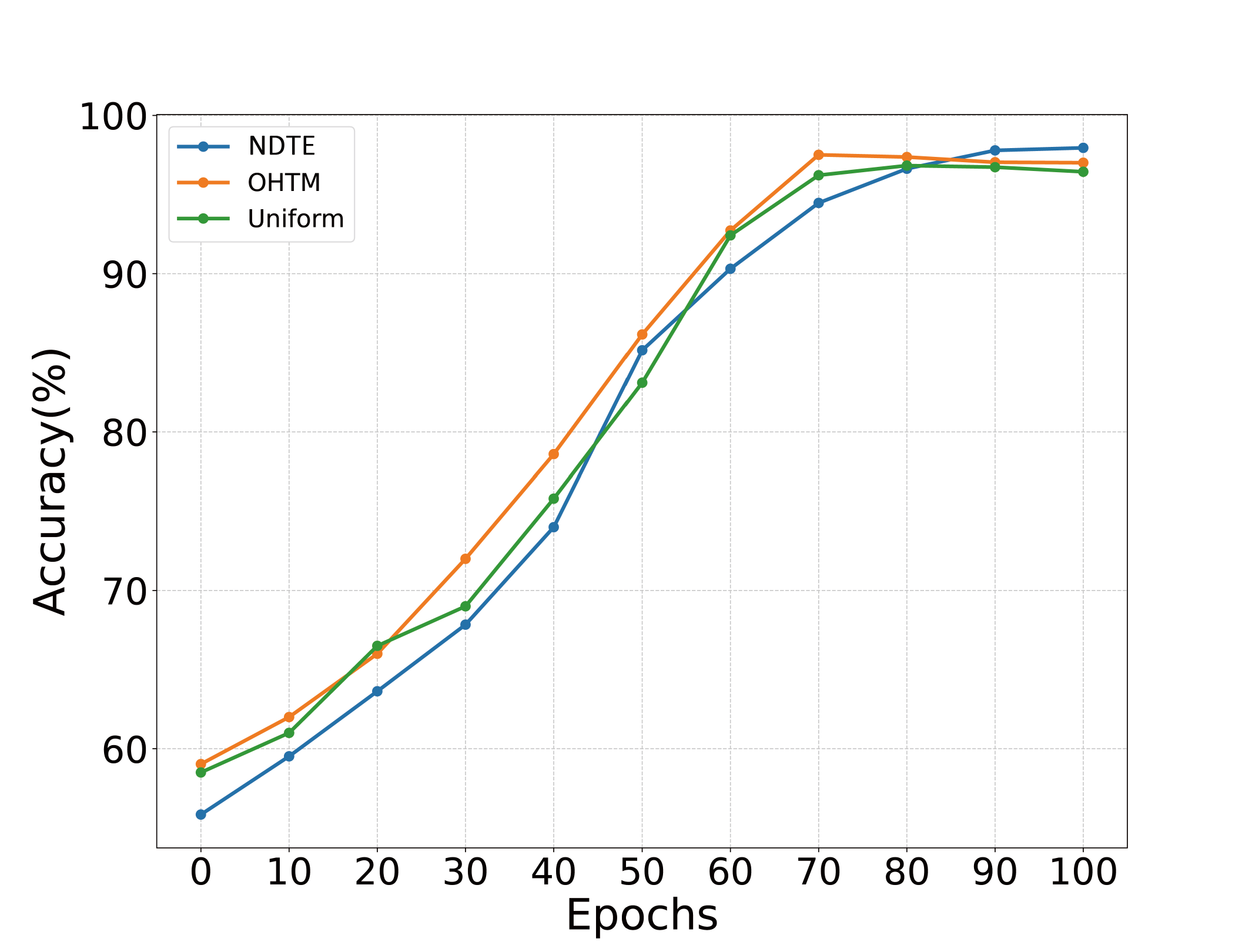}\label{fig:standard}}
     \subfigure[Cross-domain]{\includegraphics[width=0.245\textwidth]{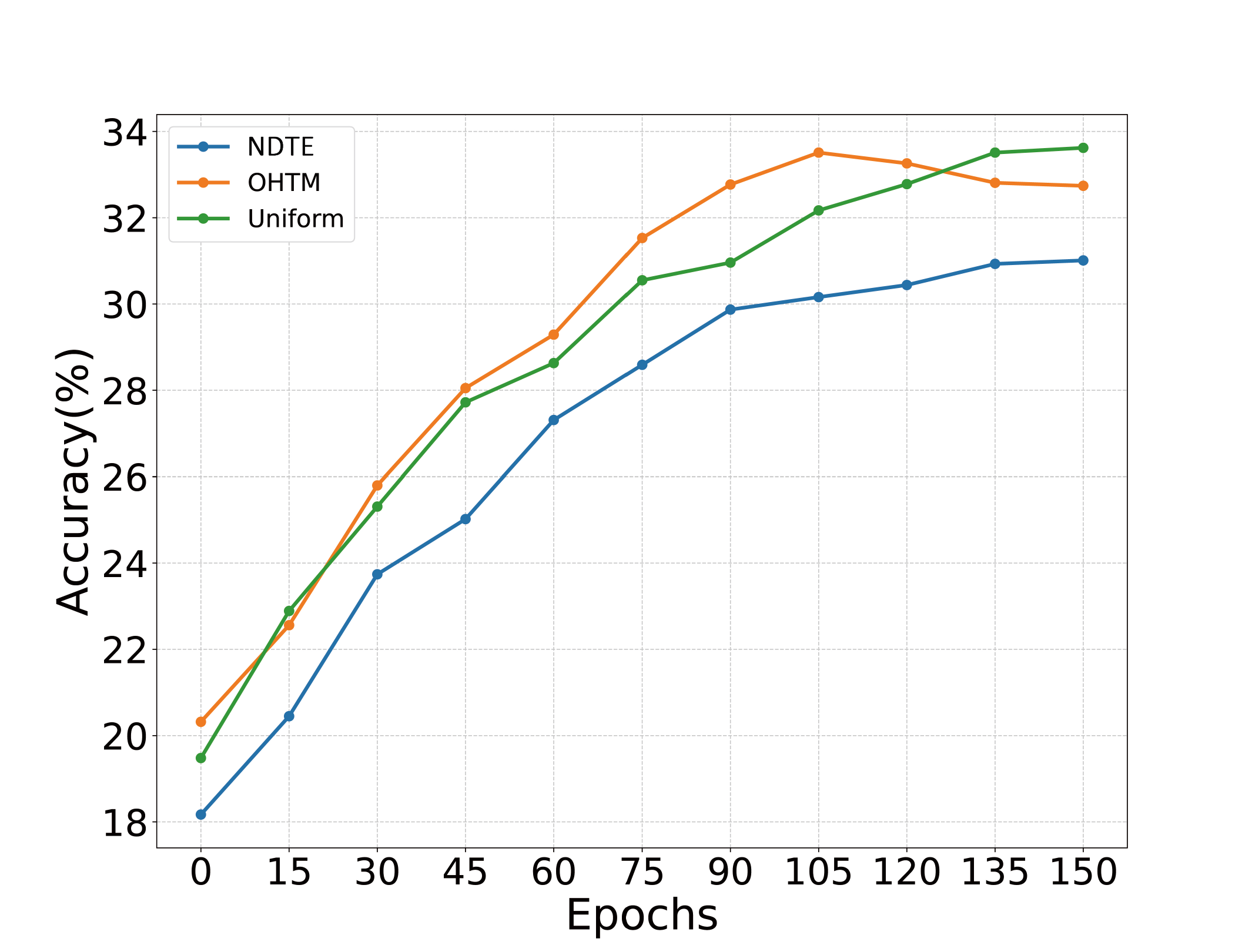}\label{fig:cross_domain}}
     \subfigure[Multi-domain]{\includegraphics[width=0.245\textwidth]{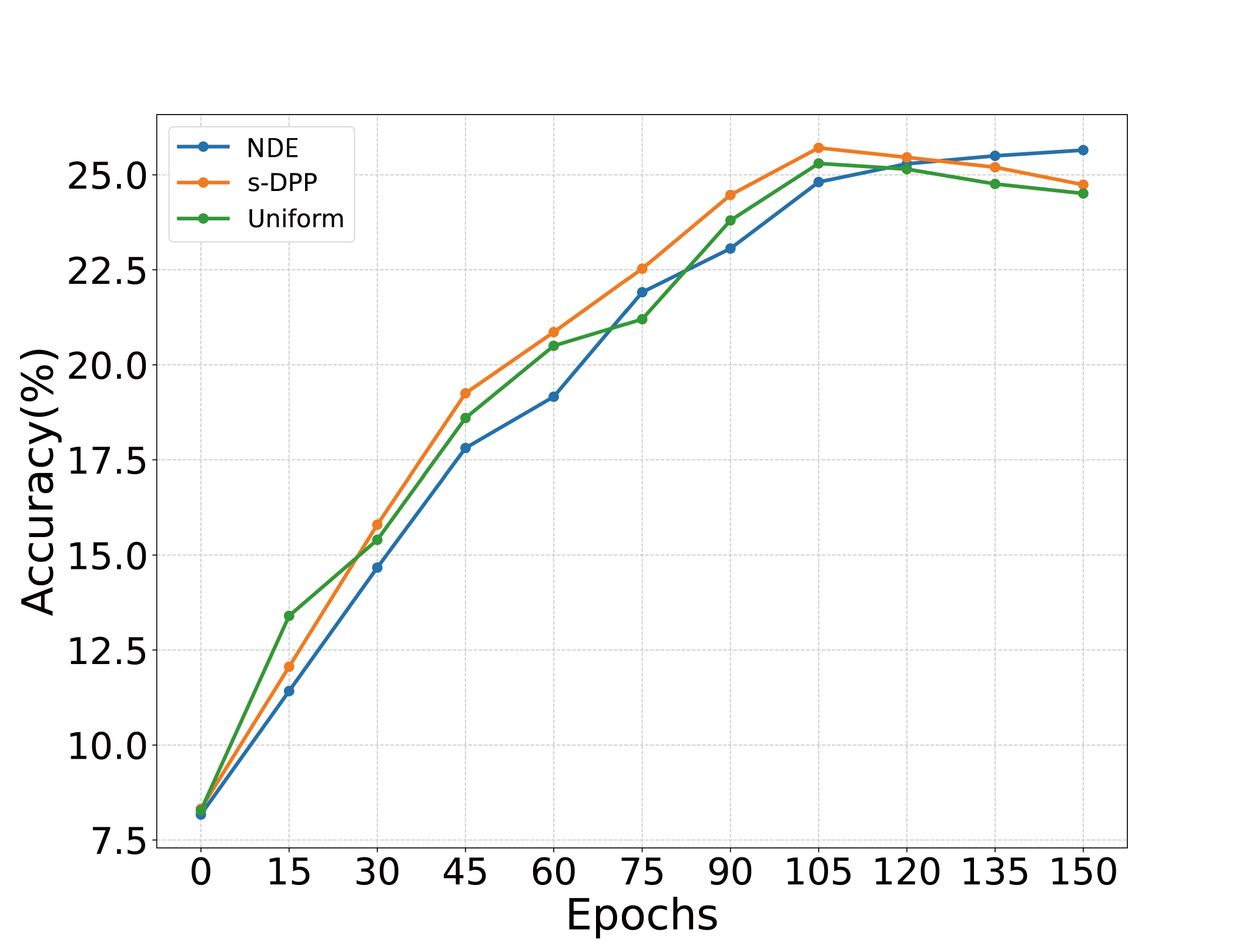}\label{fig:multi_domain}}
     \subfigure[Regression]{\includegraphics[width=0.245\textwidth]{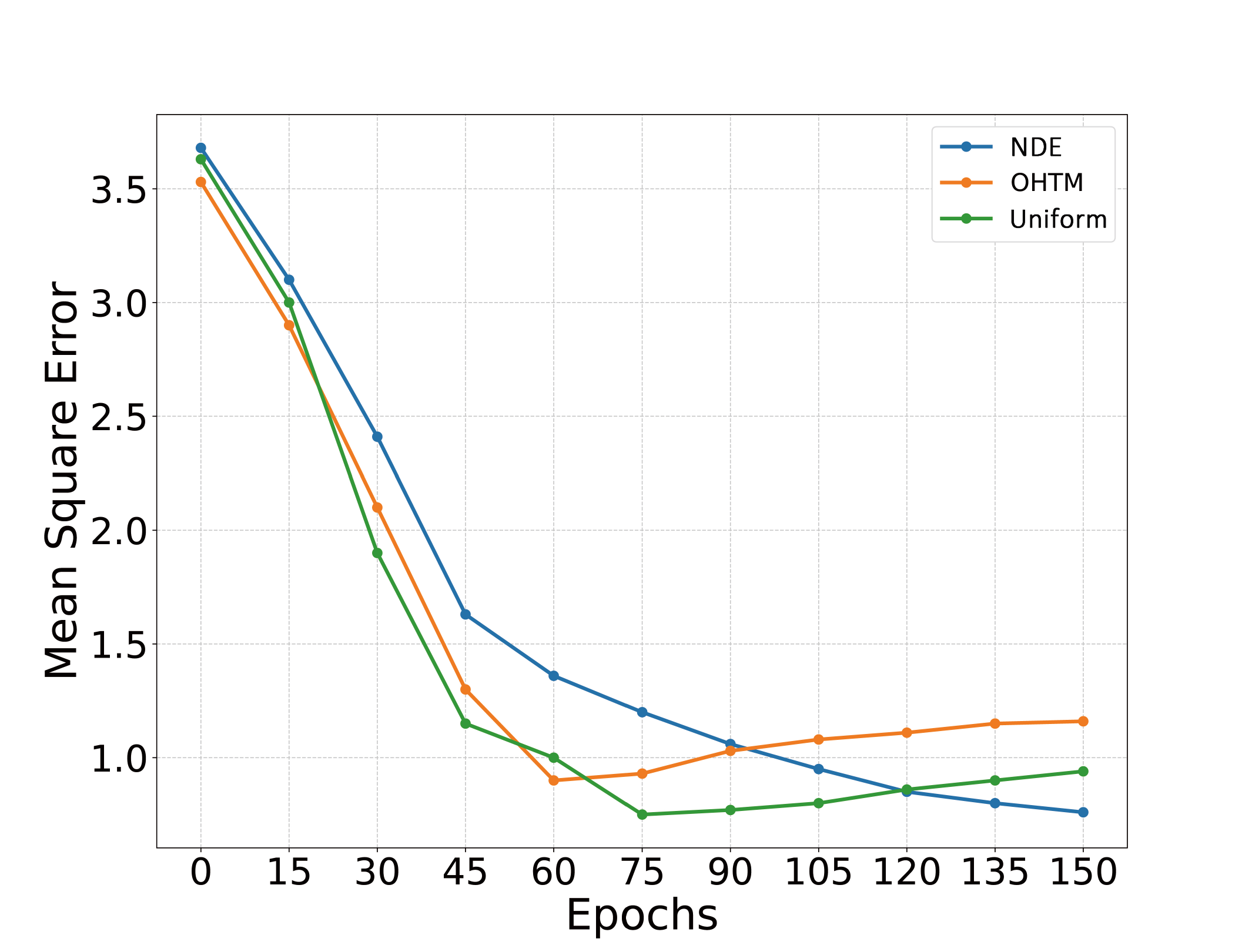}\label{fig:Regression}}
    \caption{The training process of MAML on various meta-learning settings, including standard few-shot classification (Omniglot), cross-domain few-shot classification (miniImagenet $\to$ CUB), multi-domain few-shot classification (Meta-Dataset), and few-shot regression (Sinusoid).}
    \label{fig:2}
\end{figure*}

\begin{figure*}
\begin{center}
\includegraphics[width=\textwidth]{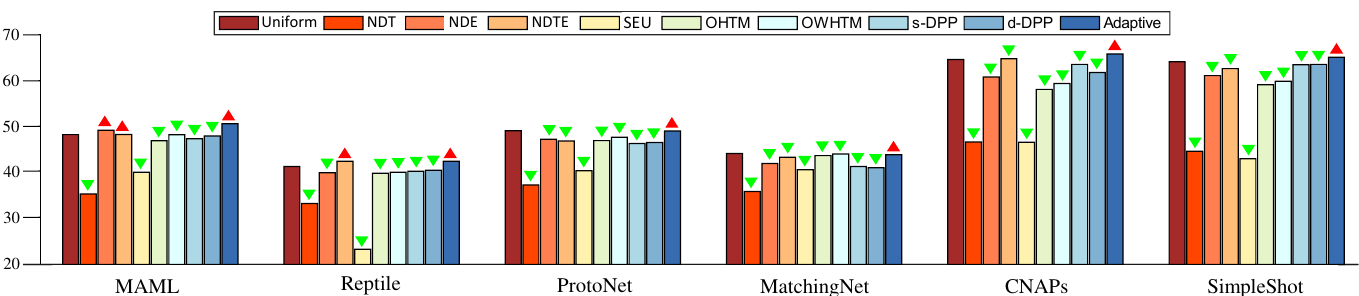}
\end{center}
   \caption{Average accuracy of meta-learning models with 10 task samplers on miniImagenet. The \textcolor{green}{$\bigtriangledown $} (\textcolor{red}{$\bigtriangleup $}) indicate that the accuracy of using the current sampler is lower (higher) than that of using Uniform Sampler.}
\label{fig:3}
\end{figure*}

\begin{table*}
\begin{center}
\caption{Standard few-shot learning results (accuracy $\pm $ 95\% confidence interval) on image classification. The $\downarrow $ ($\uparrow $) indicates that the result is lower (higher) than using Uniform sampler. The optimal value for each row in this table is highlighted in \textbf{bold}. The ``-'' denotes that the result is not reported.}
\label{table:1}
\resizebox{\linewidth}{!}{
\begin{tabular}{l|l|c|cccc|cccc|cc|c}
\toprule[1.2pt]
& \multirow{2}{*}{\textbf{Model}} & \multicolumn{10}{c}{\textbf{Samplers}}\\  \cline{3-14}
    & & \textbf{Uniform} & \textbf{NDT} & \textbf{NDE} & \textbf{NDTE} & \textbf{SEU} & \textbf{OHTM} & \textbf{OWHTM} & \textbf{s-DPP} & \textbf{d-DPP} & \textbf{GCP} & \textbf{DATS} & \textbf{ASr}\\
\midrule
\multirow{11}{*}{\rotatebox{90}{miniImagenet}}
& MAML & 48.51 $\pm$ 0.32 & 35.47 $\pm$ 0.30 $\downarrow$ & 49.49 $\pm$ 0.31 $\uparrow$ & 48.53 $\pm$ 0.30 $\uparrow$ & 40.24 $\pm$ 0.27 $\downarrow$ & 47.15 $\pm$ 0.25 $\downarrow$ & 48.49 $\pm$ 0.29 $\downarrow$ & 47.62 $\pm$ 0.30 $\downarrow$ & 48.21 $\pm$ 0.28 $\downarrow$ & 49.13 $\pm$ 0.31 $\uparrow$ & 50.01 $\pm$ 0.29 $\uparrow$
 & \textbf{51.91 $\pm$ 0.29} $\uparrow$\\
& Reptile & 41.49 $\pm$ 0.28 & 33.40 $\pm$ 0.27 $\downarrow$ & 40.15 $\pm$ 0.25 $\downarrow$ & 42.65 $\pm$ 0.25 $\uparrow$ & 23.32 $\pm$ 0.19 $\downarrow$ & 40.03 $\pm$ 0.18 $\downarrow$ & 40.27 $\pm$ 0.24 $\downarrow$ & 40.48 $\pm$ 0.22 $\downarrow$ & 40.70 $\pm$ 0.20 $\downarrow$ & 41.02 $\pm$ 0.28 $\downarrow$ & 41.99 $\pm$ 0.28 $\uparrow$  &  \textbf{43.65 $\pm$ 0.24} $\uparrow$ \\
& MetaOptNet & 58.71 $\pm$ 0.20 & 34.58 $\pm$ 0.18 $\downarrow$ & 56.39 $\pm$ 0.26 $\downarrow$ & 52.24 $\pm$ 0.24 $\downarrow$ & 42.21 $\pm$ 0.24 $\downarrow$ & 50.47 $\pm$ 0.28 $\downarrow$ & 51.86 $\pm$ 0.25 $\downarrow$  & 56.01 $\pm$ 0.23 $\downarrow$ & 51.45 $\pm$ 0.24 $\downarrow$ & 59.74$\pm$ 0.25 $\uparrow$ & 58.70 $\pm$ 0.25 $\downarrow$ & \textbf{60.20 $\pm$ 0.26} $\uparrow$ \\
& ProtoNet & 49.29 $\pm$ 0.24 & 37.45 $\pm$ 0.18 $\downarrow$ & 47.48 $\pm$ 0.19 $\downarrow$ & 47.10 $\pm$ 0.23 $\downarrow$ & 40.58 $\pm$ 0.23 $\downarrow$ & 47.19 $\pm$ 0.26 $\downarrow$ & 47.95 $\pm$ 0.28 $\downarrow$ & 46.55 $\pm$ 0.25 $\downarrow$ & 46.78 $\pm$ 0.27 $\downarrow$ & 48.32$\pm$ 0.23 $\downarrow$ & 49.55 $\pm$ 0.20 $\uparrow$ & \textbf{49.37 $\pm$ 0.28} $\uparrow$\\
& MatchingNet & 44.10 $\pm$ 0.22 & 36.02 $\pm$ 0.20 $\downarrow$ & 42.19 $\pm$ 0.25 $\downarrow$ & 43.56 $\pm$ 0.21 $\downarrow$ & 40.78 $\pm$ 0.22 $\downarrow$ & 43.90 $\pm$ 0.25 $\downarrow$ & 44.24 $\pm$ 0.24 $\downarrow$ & 41.51 $\pm$ 0.22 $\downarrow$ & 41.24 $\pm$ 0.25 $\downarrow$ & 44.99 $\pm$ 0.23 $\uparrow$ & 44.21 $\pm$ 0.24 $\uparrow$ & \textbf{46.38 $\pm$ 0.22} $\uparrow$\\
& RelationNet & 50.44 $\pm$ 0.30 & 42.24 $\pm$ 0.28 $\downarrow$ & 46.17 $\pm$ 0.27 $\downarrow$ & 47.02 $\pm$ 0.27 $\downarrow$ & 45.98 $\pm$ 0.28 $\downarrow$ & 46.54 $\pm$ 0.24 $\downarrow$ & 46.67 $\pm$ 0.26 $\downarrow$ & 48.48 $\pm$ 0.25 $\downarrow$ & 47.23 $\pm$ 0.28 $\downarrow$ & 50.66 $\pm$ 0.28 $\uparrow$ & 50.50 $\pm$ 0.27 $\uparrow$ & \textbf{52.24 $\pm$ 0.24} $\uparrow$ \\
& CNAPs & 65.03 $\pm$ 0.31 & 46.87 $\pm$ 0.24 $\downarrow$ & 61.15 $\pm$ 0.27 $\downarrow$ & 65.20 $\pm$ 0.27 $\downarrow$ & 46.82 $\pm$ 0.35 $\downarrow$ & 58.45 $\pm$ 0.30 $\downarrow$ & 59.74 $\pm$ 0.30 $\downarrow$ & 63.94 $\pm$ 0.21 $\downarrow$ & 62.17 $\pm$ 0.25 $\downarrow$ & 63.68 $\pm$ 0.25 $\downarrow$ & 64.23 $\pm$ 0.24 $\downarrow$ & \textbf{66.24 $\pm$ 0.24} $\uparrow$\\
& SCNAP & 70.61 $\pm$ 0.28 & 47.50 $\pm$ 0.24 $\downarrow$ & 62.48 $\pm$ 0.26 $\downarrow$ & 64.42 $\pm$ 0.26 $\downarrow$ & 52.50 $\pm$ 0.23 $\downarrow$ & 62.21 $\pm$ 0.23 $\downarrow$ & 62.84 $\pm$ 0.21 $\downarrow$ & 67.47 $\pm$ 0.25 $\downarrow$ & 68.16 $\pm$ 0.27 $\downarrow$ & 71.48 $\pm$ 0.29 $\uparrow$ & 70.96 $\pm$ 0.24 $\uparrow$ & \textbf{73.52 $\pm$ 0.22} $\uparrow$ \\
& SimpleShot & 64.51 $\pm$ 0.30 & 44.84 $\pm$ 0.29 $\downarrow$ & 61.47 $\pm$ 0.35 $\downarrow$ & 63.01 $\pm$ 0.34 $\downarrow$ & 43.21 $\pm$ 0.38 $\downarrow$ & 59.45 $\pm$ 0.29 $\downarrow$ & 60.23 $\pm$ 0.35 $\downarrow$ & 63.86 $\pm$ 0.38 $\downarrow$ & 63.94 $\pm$ 0.31 $\downarrow$ & 64.18 $\pm$ 0.22 $\downarrow$ & 64.84 $\pm$ 0.26 $\downarrow$ & \textbf{66.52 $\pm$ 0.31} $\uparrow$\\
& SUR & 64.92 $\pm$ 0.29 & 50.24 $\pm$ 0.28 $\downarrow$ & 62.83 $\pm$ 0.31 $\downarrow$ & 57.43 $\pm$ 0.30 $\downarrow$ & 47.98 $\pm$ 0.30 $\downarrow$ & 62.04 $\pm$ 0.27 $\downarrow$ & 63.39 $\pm$ 0.33 $\downarrow$ & 64.02 $\pm$ 0.25 $\downarrow$ & 64.45 $\pm$ 0.30 $\downarrow$ & 65.01 $\pm$ 0.24 $\uparrow$ & 64.95 $\pm$ 0.27 $\uparrow$ & \textbf{67.01 $\pm$ 0.31} $\uparrow$ \\
& PPA & 59.61 $\pm$ 0.33 & 46.74 $\pm$ 0.27 $\downarrow$ & 56.91 $\pm$ 0.25 $\downarrow$ & 57.40 $\pm$ 0.25 $\downarrow$ & 56.89 $\pm$ 0.28 $\downarrow$ & 59.82 $\pm$ 0.31 $\uparrow$ & 60.01 $\pm$ 0.24 $\uparrow$ & 57.26 $\pm$ 0.27 $\uparrow$ & 55.92 $\pm$ 0.33 $\downarrow$ & 60.18 $\pm$ 0.24 $\uparrow$ & 59.88 $\pm$ 0.26 $\uparrow$ & \textbf{62.39 $\pm$ 0.26} $\uparrow$ \\
\midrule
\multirow{11}{*}{\rotatebox{90}{Omniglot}}
& MAML & 97.64 $\pm$ 0.19 & 82.30 $\pm$ 0.17 $\downarrow$ & 97.32 $\pm$ 0.25 $\downarrow$ & 97.95 $\pm$ 0.21 $\uparrow$ & 89.69 $\pm$ 0.23 $\downarrow$ & 97.01 $\pm$ 0.23 $\downarrow$ & 96.91 $\pm$ 0.22 $\downarrow$ & 94.56 $\pm$ 0.24 $\downarrow$ & 94.19 $\pm$ 0.20 $\downarrow$ & 98.84 $\pm$ 0.23 $\uparrow$  & 98.01 $\pm$ 0.22 $\uparrow$ & \textbf{98.58 $\pm$ 0.17} $\uparrow$ \\
& Reptile & 93.68 $\pm$ 0.18 & 79.02 $\pm$ 0.12 $\downarrow$ & 93.15 $\pm$ 0.14 $\downarrow$ & 93.37 $\pm$ 0.14 $\downarrow$ & 94.06 $\pm$ 0.13 $\uparrow$ & 92.38 $\pm$ 0.09 $\downarrow$ & 92.69 $\pm$ 0.17 $\downarrow$ & 94.42 $\pm$ 0.15 $\uparrow$ & 94.01 $\pm$ 0.14 $\uparrow$ & 93.26 $\pm$ 0.27 $\downarrow$ & 92.99 $\pm$ 0.21 $\downarrow$ & \textbf{95.82 $\pm$ 0.14} $\uparrow$ \\
& MetaOptNet & 97.01 $\pm$ 0.15 & 84.04 $\pm$ 0.14 $\downarrow$ & 96.24 $\pm$ 0.08 $\downarrow$ & 97.47 $\pm$ 0.10 $\uparrow$ & 94.58 $\pm$ 0.11 $\downarrow$ & 96.23 $\pm$ 0.12 $\downarrow$ & 95.07 $\pm$ 0.13 $\downarrow$ & 95.52 $\pm$ 0.14 $\downarrow$ & 97.41 $\pm$ 0.12 $\uparrow$ & 97.11 $\pm$ 0.23 $\uparrow$ & 96.89 $\pm$ 0.29 $\downarrow$ & \textbf{98.77 $\pm$ 0.10} $\uparrow$\\
& ProtoNet & 97.51 $\pm$ 0.14 & 85.04 $\pm$ 0.20 $\downarrow$ & 95.84 $\pm$ 0.19 $\downarrow$ & 97.01 $\pm$ 0.12 $\downarrow$ & 96.12 $\pm$ 0.17 $\downarrow$ & 97.91 $\pm$ 0.17 $\uparrow$ & 98.03 $\pm$ 0.15 $\uparrow$ & 96.79 $\pm$ 0.14 $\downarrow$ & 98.10 $\pm$ 0.11 $\uparrow$ & 97.23 $\pm$ 0.26 $\downarrow$ & 97.64 $\pm$ 0.29 $\uparrow$ & \textbf{98.95 $\pm$ 0.09} $\uparrow$ \\
& MatchingNet & 95.84 $\pm$ 0.12 & 70.41 $\pm$ 0.11 $\downarrow$ & 78.84 $\pm$ 0.17 $\downarrow$ & 89.42 $\pm$ 0.15 $\downarrow$ & 90.54 $\pm$ 0.14 $\downarrow$ & 93.40 $\pm$ 0.14 $\downarrow$ & 95.15 $\pm$ 0.16 $\downarrow$ & 95.27 $\pm$ 0.21 $\downarrow$ & 96.10 $\pm$ 0.19 $\uparrow$ & 95.23 $\pm$ 0.26 $\downarrow$ & 95.88 $\pm$ 0.24 $\uparrow$ & \textbf{97.06 $\pm$ 0.14} $\uparrow$\\
& RelationNet & 95.42 $\pm$ 0.07 & 64.17 $\pm$ 0.11 $\downarrow$ & 75.04 $\pm$ 0.12 $\downarrow$ & 91.55 $\pm$ 0.09  $\downarrow$ & 92.48 $\pm$ 0.13 $\downarrow$ & 93.21 $\pm$ 0.06 $\downarrow$ & 94.19 $\pm$ 0.11 $\downarrow$ & 93.70 $\pm$ 0.12 $\downarrow$ & 94.23 $\pm$ 0.13 $\downarrow$ & 95.37 $\pm$ 0.12 $\downarrow$ & 95.03 $\pm$ 0.17 $\downarrow$ & \textbf{95.71 $\pm$ 0.10} $\uparrow$ \\
& CNAPs & 96.41 $\pm$ 0.18 & 64.51 $\pm$ 0.19 $\downarrow$ & 90.17 $\pm$ 0.23 $\downarrow$ & 89.25 $\pm$ 0.17 $\downarrow$ & 69.03 $\pm$ 0.10 $\downarrow$ & 91.74 $\pm$ 0.17 $\downarrow$ & 91.46 $\pm$ 0.12 $\downarrow$ & 95.94 $\pm$ 0.13 $\downarrow$ & 95.52 $\pm$ 0.14 $\downarrow$ & 96.57 $\pm$ 0.27 $\uparrow$ & 96.02 $\pm$ 0.30 $\downarrow$ & \textbf{96.67 $\pm$ 0.13} $\uparrow$ \\
& SCNAP & 95.48 $\pm$ 0.14 & 62.03 $\pm$ 0.15 $\downarrow$ & 91.52 $\pm$ 0.12 $\downarrow$ & 91.24 $\pm$ 0.17 $\downarrow$ & 75.74 $\pm$ 0.19 $\downarrow$ & 90.57 $\pm$ 0.16 $\downarrow$ & 91.46 $\pm$ 0.11 $\downarrow$ & 95.62 $\pm$ 0.14 $\uparrow$ & 95.38 $\pm$ 0.10 $\downarrow$ & 95.71 $\pm$ 0.13 $\uparrow$ & 95.62 $\pm$ 0.13 $\uparrow$ & \textbf{96.39 $\pm$ 0.14} $\uparrow$\\
& SimpleShot & 98.14 $\pm$ 0.15 & 81.59 $\pm$ 0.21 $\downarrow$ & 94.62 $\pm$ 0.16 $\downarrow$ & 95.58 $\pm$ 0.17 $\downarrow$ & 95.60 $\pm$ 0.15 $\downarrow$ & 91.18 $\pm$ 0.15 $\downarrow$ & 96.05 $\pm$ 0.16 $\downarrow$ & 96.43 $\pm$ 0.14 $\downarrow$ & 97.94 $\pm$ 0.14 $\downarrow$ & 98.00 $\pm$ 0.14 $\uparrow$ & 98.14 $\pm$ 0.16 $\uparrow$ & \textbf{98.79 $\pm$ 0.13} $\uparrow$\\
& SUR & 94.51 $\pm$ 0.14 & 85.44 $\pm$ 0.19 $\downarrow$ & 89.39 $\pm$ 0.21 $\downarrow$ & 90.72 $\pm$ 0.19 $\downarrow$ & 83.05 $\pm$ 0.18 $\downarrow$ & 91.93 $\pm$ 0.22 $\downarrow$ & 90.77 $\pm$ 0.23 $\downarrow$ & 88.13 $\pm$ 0.20 $\downarrow$ & 86.40 $\pm$ 0.17 $\downarrow$ & 89.28 $\pm$ 0.21 $\downarrow$ & 89.23 $\pm$ 0.25 $\downarrow$ & \textbf{95.52 $\pm$ 0.19} $\uparrow$ \\
& PPA & 97.59 $\pm$ 0.19 & 80.24 $\pm$ 0.37 $\downarrow$ & 95.17 $\pm$ 0.28 $\downarrow$ & 96.73 $\pm$ 0.29 $\downarrow$ & 94.66 $\pm$ 0.21 $\downarrow$ & 96.42 $\pm$ 0.24 $\downarrow$ & 97.78 $\pm$ 0.29 $\uparrow$ & 94.70 $\pm$ 0.27 $\downarrow$ & 94.29 $\pm$ 0.20 $\downarrow$ & 96.12 $\pm$ 0.16 $\downarrow$ & 97.70 $\pm$ 0.18 $\uparrow$ & \textbf{97.81 $\pm$ 0.24} $\uparrow$\\
\midrule
\multirow{11}{*}{\rotatebox{90}{tieredImagenet}}
& MAML & 52.38 $\pm$ 0.25 & 32.17 $\pm$ 0.28 $\downarrow$ & 51.75 $\pm$ 0.27 $\downarrow$ & 52.63 $\pm$ 0.27 $\uparrow$ & 41.29 $\pm$ 0.25 $\downarrow$ & 49.76 $\pm$ 0.24 $\downarrow$ & 50.41 $\pm$ 0.20 $\downarrow$ & 51.88 $\pm$ 0.24 $\downarrow$ & 51.06 $\pm$ 0.26 $\downarrow$ & 52.39 $\pm$ 0.24 $\uparrow$ & 52.38 $\pm$ 0.24 $\uparrow$ & \textbf{54.15 $\pm$ 0.19} $\uparrow$ \\
& Reptile & 50.16 $\pm$ 0.24 & 33.37 $\pm$ 0.28 $\downarrow$ & 49.45 $\pm$ 0.24 $\downarrow$ & 51.71 $\pm$ 0.26 $\uparrow$ & 36.29 $\pm$ 0.26 $\downarrow$ & 50.02 $\pm$ 0.25 $\downarrow$ & 49.65 $\pm$ 0.29 $\downarrow$ & 49.30 $\pm$ 0.24 $\downarrow$ & 50.47 $\pm$ 0.20 $\uparrow$ & 50.36 $\pm$ 0.23 $\uparrow$ & 50.74 $\pm$ 0.30 $\uparrow$ & \textbf{52.33 $\pm$ 0.21} $\uparrow$ \\
& MetaOptNet & 49.58 $\pm$ 0.27 & 38.01 $\pm$ 0.32 $\downarrow$ & 48.68 $\pm$ 0.30 $\downarrow$ & 46.40 $\pm$ 0.33 $\downarrow$ & 36.62 $\pm$ 0.28 $\downarrow$ & 45.33 $\pm$ 0.27 $\downarrow$ & 45.58 $\pm$ 0.27 $\downarrow$ & 49.87 $\pm$ 0.30 $\uparrow$ & 49.56 $\pm$ 0.32 $\uparrow$ & 49.32 $\pm$ 0.25 $\downarrow$ & 49.70 $\pm$ 0.28 $\uparrow$ & \textbf{50.29 $\pm$ 0.29} $\uparrow$\\
& ProtoNet & 50.94 $\pm$ 0.25 & 35.19 $\pm$ 0.22 $\downarrow$ & 49.72 $\pm$ 0.24 $\downarrow$ & 47.37 $\pm$ 0.26 $\downarrow$ & 40.83 $\pm$ 0.25 $\downarrow$ & 35.98 $\pm$ 0.21 $\downarrow$ & 35.66 $\pm$ 0.23 $\downarrow$ & 51.83 $\pm$ 0.25 $\uparrow$ & 51.12 $\pm$ 0.20 $\uparrow$ & 51.98 $\pm$ 0.24 $\uparrow$ & 51.37 $\pm$ 0.29 $\uparrow$ & \textbf{53.70 $\pm$ 0.27} $\uparrow$ \\
& MatchingNet & 43.95 $\pm$ 0.19 & 32.74 $\pm$ 0.21 $\downarrow$ & 42.39 $\pm$ 0.24 $\downarrow$ & 43.21 $\pm$ 0.25 $\downarrow$ & 37.06 $\pm$ 0.22 $\downarrow$ & 42.24 $\pm$ 0.27 $\downarrow$ & 43.55 $\pm$ 0.29 $\downarrow$ & 45.14 $\pm$ 0.25 $\uparrow$ & 43.97 $\pm$ 0.31 $\uparrow$ & 44.22 $\pm$ 0.23 $\uparrow$ & 45.02 $\pm$ 0.20 $\uparrow$ & \textbf{46.91 $\pm$ 0.19} $\uparrow$ \\
& RelationNet & 55.17 $\pm$ 0.31 & 38.48 $\pm$ 0.28 $\downarrow$ & 45.71 $\pm$ 0.35 $\downarrow$ & 48.92 $\pm$ 0.37 $\downarrow$ & 41.38 $\pm$ 0.41 $\downarrow$ & 54.65 $\pm$ 0.37 $\downarrow$ & 55.40 $\pm$ 0.42 $\uparrow$ & 54.22 $\pm$ 0.34 $\downarrow$ & 55.01 $\pm$ 0.29 $\downarrow$ & 55.10 $\pm$ 0.19 $\uparrow$ & 55.89 $\pm$ 0.24 $\uparrow$ & \textbf{56.97 $\pm$ 0.33} $\uparrow$ \\
& CNAPs & 49.57 $\pm$ 0.29 & 30.83 $\pm$ 0.30 $\downarrow$ & 44.56 $\pm$ 0.34 $\downarrow$ & 46.21 $\pm$ 0.34 $\downarrow$ & 45.09 $\pm$ 0.30 $\downarrow$ & 47.68 $\pm$ 0.33 $\downarrow$ & 47.30 $\pm$ 0.31$\downarrow$ & 45.11 $\pm$ 0.34 $\downarrow$ & 46.96 $\pm$ 0.33 $\downarrow$ & 48.89 $\pm$ 0.19 $\downarrow$ & 49.86 $\pm$ 0.27 $\uparrow$ & \textbf{49.73 $\pm$ 0.41} $\uparrow$ \\
& SCNAP & 50.34 $\pm$ 0.20 & 38.46 $\pm$ 0.24 $\downarrow$ & 47.91 $\pm$ 0.29 $\downarrow$ & 45.76 $\pm$ 0.27 $\downarrow$ & 40.22 $\pm$ 0.31 $\downarrow$ & 46.15 $\pm$ 0.32 $\downarrow$ & 47.68 $\pm$ 0.30 $\downarrow$ & 47.45 $\pm$ 0.27 $\downarrow$ & 47.70 $\pm$ 0.29 $\downarrow$ & 49.88 $\pm$ 0.24 $\downarrow$ & 50.83 $\pm$ 0.29 $\uparrow$ & \textbf{50.82 $\pm$ 0.26} $\uparrow$ \\
& SimpleShot & 51.19 $\pm$ 0.19 & 31.57 $\pm$ 0.21 $\downarrow$ & 42.17 $\pm$ 0.22 $\downarrow$ & 45.68 $\pm$ 0.22 $\downarrow$ & 39.87 $\pm$ 0.25 $\downarrow$ & 45.20 $\pm$ 0.17 $\downarrow$ & 46.31 $\pm$ 0.20 $\downarrow$ & 41.76 $\pm$ 0.19 $\downarrow$ & 40.05 $\pm$ 0.22 $\downarrow$ & 51.22 $\pm$ 0.16 $\uparrow$ & 50.90 $\pm$ 0.15 $\downarrow$ & \textbf{52.49 $\pm$ 0.22} $\uparrow$ \\
& SUR & 60.51 $\pm$ 0.22 & 42.48 $\pm$ 0.22 $\downarrow$ & 55.18 $\pm$ 0.24 $\downarrow$ & 57.78 $\pm$ 0.23 $\downarrow$ & 51.23 $\pm$ 0.19 $\downarrow$ & 60.83 $\pm$ 0.19 $\uparrow$ & 60.87 $\pm$ 0.26 $\uparrow$ & 60.21 $\pm$ 0.24 $\downarrow$ & 56.02 $\pm$ 0.24 $\downarrow$ & - & - & \textbf{61.72 $\pm$ 0.19} $\uparrow$ \\
& PPA & 59.13 $\pm$ 0.19 & 44.38 $\pm$ 0.24 $\downarrow$ & 55.73 $\pm$ 0.22 $\downarrow$ & 56.96 $\pm$ 0.20 $\downarrow$ & 54.54 $\pm$ 0.29 $\downarrow$ & 56.05 $\pm$ 0.27 $\downarrow$ & 57.45 $\pm$ 0.27 $\downarrow$ & 58.54 $\pm$ 0.28 $\downarrow$ & 54.90 $\pm$ 0.24 $\downarrow$ & 58.01 $\pm$ 0.25 $\downarrow$ & 59.64 $\pm$ 0.24 $\uparrow$ & \textbf{60.45 $\pm$ 0.23} $\uparrow$ \\
\midrule
\multirow{9}{*}{\rotatebox{90}{CIFAR-FS}}
& MAML & 58.97 $\pm$ 0.29 & 39.45 $\pm$ 0.20 $\downarrow$ & 58.78 $\pm$ 0.21 $\downarrow$ & 59.26 $\pm$ 0.25 $\uparrow$ & 52.73 $\pm$ 0.24 $\downarrow$ & 54.40 $\pm$ 0.24 $\downarrow$ & 55.55 $\pm$ 0.21 $\downarrow$ & 57.38 $\pm$ 0.27 $\downarrow$ & 57.84 $\pm$ 0.32 $\downarrow$ & 59.74 $\pm$ 0.24 $\uparrow$ & 60.00 $\pm$ 0.24 $\uparrow$ & \textbf{61.03 $\pm$ 0.29} $\uparrow$ \\
& Reptile & 59.51 $\pm$ 0.19 & 47.48 $\pm$ 0.24 $\downarrow$ & 55.79 $\pm$ 0.20 $\downarrow$ & 56.28 $\pm$ 0.22 $\downarrow$ & 54.31 $\pm$ 0.21 $\downarrow$ & 55.88 $\pm$ 0.18 $\downarrow$ & 54.37 $\pm$ 0.18 $\downarrow$ & 58.72 $\pm$ 0.17 $\downarrow$ & 57.91 $\pm$ 0.19 $\downarrow$ & 59.22 $\pm$ 0.23 $\downarrow$ & 59.82 $\pm$ 0.24 $\uparrow$ & \textbf{60.45 $\pm$ 0.21} $\uparrow$ \\
& MetaOptNet & 54.46 $\pm$ 0.29 & 49.79 $\pm$ 0.24 $\downarrow$ & 52.26 $\pm$ 0.29 $\downarrow$ & 54.84 $\pm$ 0.31 $\uparrow$ & 46.52 $\pm$ 0.28 $\downarrow$ & 51.19 $\pm$ 0.32 $\downarrow$ & 51.06 $\pm$ 0.35 $\downarrow$ & 53.25 $\pm$ 0.27 $\downarrow$ & 54.83 $\pm$ 0.28 $\uparrow$ & 55.16 $\pm$ 0.26 $\uparrow$ & 55.36 $\pm$ 0.29 $\uparrow$ & \textbf{57.94 $\pm$ 0.25} $\uparrow$\\
& ProtoNet & 54.60 $\pm$ 0.27 & 39.47 $\pm$ 0.35 $\downarrow$ & 50.74 $\pm$ 0.35 $\downarrow$ & 54.21 $\pm$ 0.33 $\downarrow$ & 49.75 $\pm$ 0.31 $\downarrow$ & 54.58 $\pm$ 0.27 $\downarrow$ & 55.30 $\pm$ 0.29 $\uparrow$ & 54.11 $\pm$ 0.30 $\downarrow$ & 54.27 $\pm$ 0.26 $\downarrow$ & 55.02 $\pm$ 0.22 $\uparrow$ & 55.97 $\pm$ 0.24 $\uparrow$ & \textbf{57.05 $\pm$ 0.20} $\uparrow$ \\
& MatchingNet & 54.79 $\pm$ 0.28 & 44.57 $\pm$ 0.33 $\downarrow$ & 48.36 $\pm$ 0.34 $\downarrow$ & 49.86 $\pm$ 0.35 $\downarrow$ & 50.10 $\pm$ 0.30 $\downarrow$ & 52.92 $\pm$ 0.27 $\downarrow$ & 52.48 $\pm$ 0.31 $\downarrow$ & 54.81 $\pm$ 0.29 $\downarrow$ & 53.48 $\pm$ 0.34 $\downarrow$ & - & - & \textbf{55.33 $\pm$ 0.28} $\uparrow$ \\
& RelationNet & 56.24 $\pm$ 0.31 & 33.74 $\pm$ 0.29 $\downarrow$ & 40.84 $\pm$ 0.28 $\downarrow$ & 46.41 $\pm$ 0.33 $\downarrow$ & 47.46 $\pm$ 0.27 $\downarrow$ & 56.02 $\pm$ 0.26 $\downarrow$ & 56.48 $\pm$ 0.30 $\uparrow$ & 56.94 $\pm$ 0.24 $\uparrow$ & 53.55 $\pm$ 0.28 $\downarrow$ & 56.01 $\pm$ 0.25 $\downarrow$ & 56.93 $\pm$ 0.17 $\uparrow$ & \textbf{57.85 $\pm$ 0.19} $\uparrow$ \\
& CNAPs & 60.08 $\pm$ 0.30 & 43.72 $\pm$ 0.28 $\downarrow$ & 55.33 $\pm$ 0.31 $\downarrow$ & 55.58 $\pm$ 0.27 $\downarrow$ & 50.91 $\pm$ 0.24 $\downarrow$ & 56.38 $\pm$ 0.28 $\downarrow$ & 56.60 $\pm$ 0.17 $\downarrow$ & 58.01 $\pm$ 0.20 $\downarrow$ & 59.29 $\pm$ 0.19 $\downarrow$ & 59.86 $\pm$ 0.29 $\uparrow$ & 60.37 $\pm$ 0.23 $\uparrow$ & \textbf{61.46 $\pm$ 0.32} $\uparrow$ \\
& SimpleShot & 59.40 $\pm$ 0.31 & 40.56 $\pm$ 0.28 $\downarrow$ & 49.45 $\pm$ 0.35 $\downarrow$ & 52.45 $\pm$ 0.27 $\downarrow$ & 46.04 $\pm$ 0.40 $\downarrow$ & 52.02 $\pm$ 0.33 $\downarrow$ & 53.34 $\pm$ 0.24 $\downarrow$ & 54.58 $\pm$ 0.21 $\downarrow$ & 52.34 $\pm$ 0.25 $\downarrow$ & 59.07 $\pm$ 0.21 $\downarrow$ & 59.44 $\pm$ 0.23 $\uparrow$ & \textbf{60.77 $\pm$ 0.19} $\uparrow$ \\
& SUR  & 63.16 $\pm$ 0.41 & 48.11 $\pm$ 0.38 $\downarrow$ & 56.25 $\pm$ 0.35 $\downarrow$ & 57.80 $\pm$ 0.40 $\downarrow$ & 53.39 $\pm$ 0.27 $\downarrow$ & 60.27 $\pm$ 0.36 $\downarrow$ & 61.66 $\pm$ 0.34 $\downarrow$ & 63.54 $\pm$ 0.34 $\uparrow$ & 60.00 $\pm$ 0.25 $\downarrow$ & - & - & \textbf{64.48 $\pm$ 0.21} $\uparrow$ \\
\bottomrule
\end{tabular}}
\end{center}
\end{table*}

%%%%%%%%%%%%%%%%%%%%%%%%%%%%%%%%%%%%%%%%%%%%%%%%%%%%%%%%%%%%%%%%%%%%%%%%%%%%%%%%%%%%%%%%%%%%%%%%%%%
%                                        4
%%%%%%%%%%%%%%%%%%%%%%%%%%%%%%%%%%%%%%%%%%%%%%%%%%%%%%%%%%%%%%%%%%%%%%%%%%%%%%%%%%%%%%%%%%%%%%%%%%%
\section{Motivating Evidence}
\label{sec:4}
Previous works \citep{kumar2022effect, willemink2020preparing} assume that increasing the diversity of training tasks will improve the generalization of meta-learning models, but we demonstrate the limitations of this view through empirical and theoretical analysis. In this section, we present the details of our analysis.

\subsection{Empirical Evidence}
\label{sec:4.1}
We use MAML and ProteNet as the backbones and evaluate their performance with different task samplers on two benchmark datasets, i.e., miniImagenet and Omniglot. We choose five different task samplers, including the standard sampler (i.e., Uniform sampler), two low-diversity samplers (i.e., NDE and SEU), and two high-diversity samplers (i.e., OHTM and s-DPP). Table \ref{table:intro} shows the classification results. We observe that increasing task diversity does not significantly improve performance compared to Uniform sampler. Even for MAML, the results are completely opposite to expectations: restricting tasks to follow a uniform distribution or reducing task diversity can achieve better results.

To further explore the effect of task diversity on different meta-learning models, we conduct extensive experiments on various meta-learning settings, including standard few-shot classification, cross-domain few-shot classification, multi-domain few-shot classification, and few-shot regression. Figure \ref{fig:3} shows the comparison results on miniImagenet between different samplers. The results demonstrate that task diversity does not lead to a significant boost in the performance of all of the models; that is, the improvement brought about by relying on task diversity is limited. Table \ref{table:1} summarizes the effect of task diversity on four benchmark datasets of standard few-shot classification. We obtain three observations: (i) Uniform sampler achieves outstanding results, surpassing other samplers designed by constraining task diversity; (ii) low-diversity samplers often perform better on meta-learning models including MAML, Reptile, and Bayesian-based models, even outperform Uniform sampler in some datasets; and (iii) high-diversity samplers perform better on datasets that include tieredImagenet and Omniglot, and meta-learning models that include MetaOptNet, ProtoNet, and MatchingNet.  
Therefore, only increasing task diversity does not improve model generalization, on the contrary, without any constraints or reducing task diversity can even enhance model performance. The results of cross-domain few-shot classification in Table \ref{table:cross-domain} indicate that Uniform and high-diversity task samplers usually achieve better results than low-diversity samplers. Similarly, the results of multi-domain few-shot classification and few-shot regression shown in Table \ref{table:multi-domain} and Table \ref{table:regression} exhibit similar performance as in standard few-shot classification.

In addition, we introduce previously proposed adaptive samplers for comparison, i.e., GCP and DATS mentioned in Section \ref{sec:3.3}. These samplers also rely on task diversity sampling tasks, but can dynamically adjust the sampling results according to the training loss of the meta-learning model. From the results in Table \ref{table:1}-Table \ref{table:multi-domain}, we find that although GCP and DATS achieve significant performance improvements on multiple datasets, e.g., miniImagenet and tieredImagenet, they are difficult to perform well in all scenarios. For example, these adaptive samplers perform worse than Uniform sampler in multi-domain few-shot learning scenarios. Therefore, even if dynamic sampling relies on task diversity, it is still difficult to guarantee obtaining a universally optimal task sampling strategy.

From the above observations, we can conclude that (i) different models prefer task sampling strategies with different task diversity scores, and the traditional view that higher task diversity leads to stronger model generalization is limited; (ii) There may be other factors affecting task sampling quality, and we cannot only rely on task diversity to find the optimal sampling strategy.

\subsection{Theoretical Evidence}
\label{sec:4.2}
A good meta-learning model can accurately extract the information related to decision-making within a task. Considering the constraints of data collection and computational complexity, meta-learning is always evaluated on few-shot learning tasks. For few-shot tasks, the performance depends on two factors \citep{rivolli2022meta}: one is discriminability, which is how separable the classes are within the task, and the other is effectiveness, which is how relevant the data are to the task.

For discriminability, the distribution of each class in the task has relatively low-dimensional intrinsic structures so that it can be mapped well \citep{chan2022redunet}. There are three reasons: (i) high-dimensional data are highly redundant; (ii) data from the same class should be similar and related; and (iii) the equivalent structure of the task is invariant to transformations and augmentations.

For effectiveness, different samples from the same class within the task should have a consistent impact on model performance. There are two reasons: (i) in few-shot learning tasks, there are limited data points that may not represent the true data distribution well, so the task-related samples with key information are important; and (ii) according to causal invariance \citep{Cheng2017causal}, the key task-related information, e.g., causal-related information, that affects task decisions is invariant on different data sets.

Therefore, we propose a high-quality meta-learning task should have the following properties:
\begin{itemize}
    \item \emph{Intra-class compaction}: Samples from the same class should have correlated features, as they belong to a low-dimensional linear subspace.
    \item \emph{Inter-class separability}: Samples from different classes should have uncorrelated features, and belong to different low-dimensional linear subspaces.
    \item \emph{Feature space enrichment}: Each task should have large and diverse feature dimensions.
    \item \emph{Causal invariance}: Task-related causal information is invariant on different datasets.
\end{itemize}

Intra-class compaction and inter-class separability measure the discriminative features; feature space enrichment enhances the information content of the task, covering more discriminative features; causal invariance ensures that the features used for learning are relevant to the task decision. The first three properties are quantifications of discriminability, and the fourth one is quantification of effectiveness. 

For example, to classify zebras on the grassland and pandas in the zoo, shape features can make zebras and pandas compact within classes and separated between classes, but discriminative texture features cannot be obtained by models. This will result in limited information learned by the model and reduced generalization \citep{li2017feature}. Therefore, the task needs to have feature space enrichment to ensure that the features contain more information, such as zebras on the grassland and fish in the water, which have discriminative features like texture, shape, and color. Considering that this kind of task may make the model focus on unnecessary background information, the task needs to have causal invariance to guide the model to only focus on useful foreground features and eliminate the background. Therefore, a high-quality task requires all four of the above attributes.

Based on the above analysis, task diversity can be regarded as a manifestation of feature space enrichment. It only reflects one aspect of the discriminative ability of the task, where higher task diversity means a larger feature space, containing more discriminative features. Therefore, task diversity cannot comprehensively evaluate the quality of the task. This also inspires us to design more reasonable measurements from the perspective of all four properties.

%%%%%%%%%%%%%%%%%%%%%%%%%%%%%%%%%%%%%%%%%%%%%%%%%%%%%%%%%%%%%%%%%%%%%%%%%%%%%%%%%%%%%%%%%%%%%%%%%%%
%                                        5
%%%%%%%%%%%%%%%%%%%%%%%%%%%%%%%%%%%%%%%%%%%%%%%%%%%%%%%%%%%%%%%%%%%%%%%%%%%%%%%%%%%%%%%%%%%%%%%%%%%
\section{Proposed Measurement}
\label{sec:5}
Based on the above insight, we propose the following three indicators to measure the quality of meta-learning tasks, including task diversity, task entropy and task difficulty.

\textbf{Task diversity} corresponds to feature space enrichment. Consistent with Subsection \ref{sec:3.2}, we define task diversity as the squared volume of the parallelepiped spanned by the samples in each task. The difference is that, we introduce logarithmic calculation and reconstruct the matrix $\mathcal{M} _{\mathcal{T}_i}$ to improve calculation speed and eliminate error accumulation. Meanwhile, we let the model encoder directly calculate the representation without using a pre-trained ProtoNet. For task ${\mathcal{T}_i}$, its diversity score is:
\begin{equation}
\label{eq:task diversity}
      t_{dg}^i\triangleq \frac{n}{2} \log\mathrm {det}(\mathcal{I} +\frac{d}{n\sigma ^2} Z_iZ_i^*)
\end{equation}
where $\mathcal{I} $ represents a $d \times d$ identity matrix, $Z_i=h(\mathcal{T}_i)$ denotes the representations obtained by encoding task $\mathcal{T}_i$ through encoder $h(\cdot)$, and $Z_i^*$ is the conjugate (transpose) of $Z_i$, $n$ is the number of samples in task $\mathcal{T}_i$, $d$ is the dimension of $Z_i$ where $Z_i\in \mathbb{R}^d$. This equation can be derived by filling $\sigma $-balls as Gaussian sources into the space spanned by $Z_i$ \citep{ma2007segmentation}, which is the same as $\mathcal{M} _{\mathcal{T}_i}$ \citep{devries2017dataset}.

According to \citep{kumar2022effect}, the larger $t_{dg}^i$ is, the better it satisfies feature space enrichment, because the larger the parallelepiped volume spanned by the eigenvectors, the more orthogonal the eigenvectors are. The theoretical analysis is provided in Theorem \ref{theorem:2}.

\textbf{Task entropy} corresponds to intra-class compaction and inter-class separability. Inspired by data coding theory \citep{chan2022redunet}, we define task entropy as the average number of bits required for lossless encoding of data in each class. It well characterizes the rate distortion of the intra-task subspace distribution, which shows how tightly the intra-class data is distributed within a task. For task ${\mathcal{T}_i}$, its entropy score is:
\begin{equation}
\label{eq:task entopy}
t_{et}^i\triangleq \sum_{j=1}^{k} (\frac{\mathrm {tr}(\mathrm {C}_i^j ) }{2n})\log\det(\mathcal{I} +\frac{d}{\mathrm {tr}(\mathrm {C}_i^j ) \epsilon^2}{{Z}_i}\mathrm {C}_i^j{{{Z}_i^*}})
\end{equation}
where $k$ is the number of classes in task $\mathcal{T}_i$, $\epsilon$ is the upper bound of the distortion during encoding, also the error upper limit in meta-learning, $\mathrm {C}_i^j$ denotes a diagonal matrix whose diagonal entries encode the samples of class $j$. The meanings of $Z_i$, $Z_i^*$, $\mathcal{I} $, $n$, and $d$ are the same as mentioned in Equation \ref{eq:task diversity}.

According to \citep{chan2022redunet}, the lower $t_{et}^i$ is, the better it satisfies intra-class compaction and inter-class separability, because the lower the average number of bits needed for lossless data encoding, the tighter the intra-class data distribution, the greater the difference between classes, making classification easier. The theoretical analysis is provided in Theorem \ref{theorem:3}.

\textbf{Task difficulty} corresponds to causal invariance. We define task difficulty as the gradient inconsistency between the query set and the support set in task ${\mathcal{T}_i}$. For task $\mathcal{T}_i$, its difficulty score is:
\begin{equation}
\label{eq:task difficulty}
    t_{df}^i\triangleq \sum\limits_{i,j} {\| {{\nabla _{x_{i,j}^s}}\mathcal{L}(\mathcal{D}_{i}^s,f) - {\nabla _{x_{i,j}^q}}\mathcal{L}(\mathcal{D}_{i}^q,f}) \|_2^2},
\end{equation}
where ${\nabla _{x_{i,j}^s}}\mathcal{L}(\mathcal{D}_{i}^s,f)$ and ${\nabla _{x_{i,j}^q}}\mathcal{L}(\mathcal{D}_{i}^q,f)$ are the gradients on the samples of the support and query sets with respect to the meta-learning objective $\mathcal{L}(\mathcal{D}_i^s,\mathcal{D}_i^q, f)$. ${\| {{\nabla _{x_{i,j}^s}}\mathcal{L}(\mathcal{D}_{i}^s,f) - {\nabla _{x_{i,j}^q}}\mathcal{L}(\mathcal{D}_{i}^q,f}) \|_2^2}$ can be regarded as the cost of distinguishing classes within the task, and the smaller the better, and $t_{df}^i$ is the accumulated cost. 

According to \citep{chen2022pareto}, the smaller $t_{df}^i$ is, the better it satisfies utility invariance, because the smaller the gradient difference between the query set and the support set, the more consistent the effect is, proving that the learned features are more important. The theoretical analysis is shown in Theorem \ref{theorem:4}.

%%%%%%%%%%%%%%%%%%%%%%%%%%%%%%%%%%%%%%%%%%%%%%%%%%%%%%%%%%%%%%%%%%%%%%%%%%%%%%%%%%%%%%%%%%%%%%%%%%%
%                                        6
%%%%%%%%%%%%%%%%%%%%%%%%%%%%%%%%%%%%%%%%%%%%%%%%%%%%%%%%%%%%%%%%%%%%%%%%%%%%%%%%%%%%%%%%%%%%%%%%%%%
\section{Theoretical Analysis}
\label{sec:6}
We conduct theoretical analysis to prove that the three proposed measurements can help select the best tasks, which possess the properties mentioned in Subsection \ref{sec:4.2}. First, we reveal the properties of optimal tasks according to the analysis in Subsection \ref{sec:4.2}. Next, we demonstrate that the proposed three measurements can well reflect these properties.

A meta-learning task $\mathcal{T}_i$ usually contains multiple subspaces, and its feature representation is represented as $Z_i$. We denote the feature representation of the optimal task as $\textbf Z_i=\textbf Z_i^1 \cup ...\cup \textbf Z_i^k$, where $\textbf Z_i^j$ is the representation of class $j$. Then we can show that $\textbf Z_i$ has the following required properties:

\begin{corollary}\label{theorem:1}
    \textit{Suppose $\textbf Z_i=\textbf Z_i^1 \cup ...\cup \textbf Z_i^k$ is the feature representation of the optimal task $\mathcal{T}_i $ that satisfy the properties mentioned in Subsection \ref{sec:4.2}. The distribution of each class within $\mathcal{T}_i$ has a support on a low-dimensional subspace, where the dimension of the subspace of class $j$ is $d_j$. Assume that the optimal solution $\textbf Z_i$ satisfies $\mathrm {rank}(\textbf Z_i)\le d $, we have}:
    \begin{itemize}
        \item Maximally Feature Space: if the subspaces are orthogonal to each other, i.e., $(\textbf Z_i^a)^*(\textbf Z_i^b)=0, a\ne b$, then the space spanned by feature representations is adequately large, i.e., $d\ge \sum_{j=1}^{k}d_j$.
        \item Maximally Discriminability: if the discriminant precision is high enough, i.e., $\epsilon^4< \underset{j}{\min}\left \{ \frac{n_j}{n}\frac{d ^2}{d_j^2} \right \} $, each subspace reaches maximum dimension, i.e., between-class get $(\textbf Z_i^a)^*(\textbf Z_i^b)=0, a\ne b$, within-class get $\mathrm {rank}(\textbf Z_i^j)= d_j$, and the largest $d_{j}-1$ singular values of $Z_{i}^{j}$ are equal.
        \item Minimally Effect Gap: the effect of different datasets for the same task remains unchanged if $\sum\limits_{j} { {{\nabla _{x_{j}^1}}\mathcal{L}(\mathcal{D}_1,f) = \sum\limits_{j}{\nabla _{x_{j}^2}}\mathcal{L}(\mathcal{D}_2,f} )}$, where $\mathcal{D}_1$ and $\mathcal{D}_2$ are two different datasets of $\mathcal{T}_i $.
    \end{itemize}

\end{corollary}

Next, we will prove that the proposed measurements match the above properties and can guide the acquisition of high-quality tasks. 

Theorem \ref{theorem:2} illustrates that task diversity can well reflect \textit{Maximally Feature Space} in Corollary \ref{theorem:1}.

\begin{theorem}\label{theorem:2}
    \textit{Let $Z_i=\left [ Z_i^1,...,,Z_i^k \right ]\in \mathbb{R}^{d\times n}$ be the representation of task $\mathcal{T}_i $, which has $k$ classes and $n= {\textstyle \sum_{j=1}^{k}n_j} $ samples. For any representations $Z_i^j\in \mathbb{R}^{d \times n_j}$ of class $j$ and any $\sigma>0$, we have:}
    \begin{equation}\label{eq:theorem2_1}
    \begin{array}{l}
    \frac{n}{2}\log\mathrm {det}(\mathcal{I} +\frac{d}{n\sigma ^2} {Z_i}^*Z_i) \\[10pt]
    \le \sum_{j=1}^{k}\frac{n}{2}\log\mathrm {det}(\mathcal{I} +\frac{d}{n\sigma ^2} {(Z_i^j)}^*{(Z_i^j)})
    \end{array}
    \end{equation}
    the equality holds if and only if:
    \begin{equation}\label{eq:theorem2_2}
    \begin{array}{l}
    {(Z_i^{j_1})}^*(Z_i^{j_2})=0,\quad s.t.\quad 1\le j_1 \le j_2 \le k
    \end{array}
    \end{equation}
\end{theorem}

Theorem \ref{theorem:3} illustrates that task entropy can well reflect  \textit{Maximally Discriminability} in Corollary \ref{theorem:1}.

\begin{theorem}\label{theorem:3}
    \textit{Let $Z_i=\left [ Z_i^1,..., Z_i^k \right ]$ be the representation of task $\mathcal{T}_i$, $\varsigma_j:= \left [ \varsigma_{1,j},..., \varsigma_{min(n_j,d),j} \right ] $ be the singular values of the representation $Z_i^j$ of class $j$, $\mathrm {C}_i=\left [ \mathrm {C}_i^1,...,\mathrm {C}_i^k \right ]$ is a collection of diagonal matrices, where the diagonal elements encode the $n$ samples into the $k$ classes. Given any $\epsilon >0$ and $d \ge d_j>0$, consider the optimization problem of task entropy:}
    \begin{equation}\label{eq:theorem3}
    \begin{array}{l}
    \underset{Z_i\in \mathbb{R}^{d\times n} }{\arg \max } (t_{et}^i)\\[10pt]
    s.t.\quad \left \| Z_i\mathrm {C}_i^j \right \|^2=\mathrm {tr}(\mathrm {C}_i^j ),\\
    \quad\qquad \mathrm {rank}(Z_i)\le  d_j, \forall j\in \left \{ 1,...,k \right \}  
    \end{array}
    \end{equation}
    Under the conditions where the error upper limit $\epsilon^4< \underset{j}{\min}\left \{ \frac{n_j}{n}\frac{d ^2}{d_j^2} \right \} $, and the dimension $d\ge  {\textstyle \sum_{j=1}^{k}d_j} $, the optimal solution $\textbf Z_i$ satisfies:
    \begin{itemize}
        \item Between-class: The representation $\textbf Z_i^{j_1}$ and $\textbf Z_i^{j_2}$ lie in orthogonal subspaces, i.e., $(\textbf Z_i^{j_1})^*(\textbf Z_i^{j_2})=0$, where $1\le j_1 \le j_2 \le k$.
        \item Within-class: each class $j$ achieves its maximal dimension $d_j$, i.e., $\mathrm {rank}(\textbf Z_i^j)= d_j$, and either $\left [ \varsigma_{1,j},..., \varsigma_{d_j,j} \right ]$ equal to $\frac{\mathrm {tr}(\mathrm {C}_i^j )}{d_j} $, or $\left [ \varsigma_{1,j},..., \varsigma_{d_j-1,j} \right ]$ equal to and have values larger than $\frac{\mathrm {tr}(\mathrm {C}_i^j )}{d_j} $.
    \end{itemize}
\end{theorem}

Theorem \ref{theorem:4} illustrates that task difficulty can well reflect  \textit{Minimally Effect Gap} in Corollary \ref{theorem:1}.

\begin{theorem}\label{theorem:4}
    \textit{The support set $\mathcal{D}_i^s$ and query set $\mathcal{D}_i^q$ are two different datasets of $\mathcal{T}_i$. For any $\mathcal{T}_i$ and $f$, we have:
    \begin{equation}\label{eq:theorem4}
    \begin{array}{l}
    \sum\limits_{i,j} {\| {{\nabla _{x_{i,j}^s}}\mathcal{L}(\mathcal{D}_{i}^s,f) - {\nabla _{x_{i,j}^q}}\mathcal{L}(\mathcal{D}_{i}^q,f}) \|_2^2}\ge 0
    \end{array}
    \end{equation}
    the equality holds if and only if the gradients of the support set $\mathcal{D}_i^s$ and query set $\mathcal{D}_i^q$ are consistent: 
    \begin{equation}
        \sum\limits_{j}{{\nabla _{x_{i,j}^s}}\mathcal{L}(\mathcal{D}_{i}^s,f) = \sum\limits_{j}{\nabla _{x_{i,j}^q}}\mathcal{L}(\mathcal{D}_{i}^q,f}) 
    \end{equation}
    }
\end{theorem}

In summary, the measurements we proposed can well evaluate the quality of meta-learning tasks. All the proofs are provided in Appendix \ref{sec:appendix_1}.

 %%%%%%%%%%%%%%%%%%%%%%%%%%%%%%%%%%%%%%%%%%%%%%%%%%%%%%%%%%%%%%%%%%%%%%%%%%%%%%%%%%%%%%%%%%%%%%%%%%%
%                                         5
%%%%%%%%%%%%%%%%%%%%%%%%%%%%%%%%%%%%%%%%%%%%%%%%%%%%%%%%%%%%%%%%%%%%%%%%%%%%%%%%%%%%%%%%%%%%%%%%%%%
\section{Methodology}
\label{sec:7}
In this section, we propose an adaptive sampler (ASr), which is a plug-and-play module for episodic meta-learning. It uses a learnable task distribution function to weight tasks based on task diversity, task entropy, and task difficulty, and then adaptively obtains the optimal probability distribution for meta-training tasks without introducing extra computational cost. To optimize ASr, we propose a simple and general meta-learning algorithm. In this section, we introduce the details of ASr and the overall objective. The pseudo-code is shown in Algorithm \ref{Algorithm 1}.

%%%%%%%%%%%%%%%%%%%%%%%%%%%%%%%%%%
%            5.1
%%%%%%%%%%%%%%%%%%%%%%%%%%%%%%%%%%
\subsection{Adaptive Sampler}
\label{sec:7.1}
An optimal task sampler is one that can obtain the optimal task pool for meta-learning models. 
One feasible implementation strategy is to choose a task pool containing numerous tasks. Then, we can apply the three proposed measurements to filter the tasks, retaining high-quality ones while removing low-quality tasks. This process results in a subset of tasks, which can be referred to as the optimal task pool. However, the cost of doing so is that it restricts the variousness of the task pool. In this paper, instead of obtaining the optimal task pool, we learn the optimal probability distribution for the task pool $\mathcal{T}$ of each episode in the training process. It weights each task in $\mathcal{T}$ based on the values of task diversity, task entropy, and task difficulty, and then performs training based on the weighted tasks. The probability distribution for each task pool $\mathcal{T}$ is generated by ASr. Also, high-quality tasks will be given higher weights and low-quality tasks will be given lower weights. The advantage of doing this is that it can ensure the model focuses on high-quality tasks while also maintaining the divergence of the task pool.

Specifically, we define the ASr as a learnable function, represented as $g_\varphi$. First, we are given a task pool $\mathcal{T}=\left \{ \mathcal{T}_1,...,\mathcal{T}_{N^{pool}} \right \} $, where $N^{pool}$ is the total number of the elements in it. Then, for each $\mathcal{T}_i$, we calculate three values, i.e., task diversity $t_{dg}^i$, task entropy $t_{et}^i$, and task difficulty $t_{df}^i$, and then combine them into a vector, denoted as $t^i=(t_{dg}^i, t_{et}^i, t_{df}^i)$. The definition and calculation of task diversity $t_{dg}^i$, task entropy $t_{et}^i$, and task difficulty $t_{df}^i$ are illustrated in Section \ref{sec:5}. Next, we input $t^i$ into $g_\varphi$ to get output $w_i$. At last, we obtain $w = \{ {w_1},...,{w_{{N^{pool}}}}\}$, where all elements are normalized by $w_i = w_i/\sum\nolimits_{i = 1}^{N^{pool}} w_i$. We regard $w_i$ as the importance of ${\mathcal{T}_i}$ in $\mathcal{T}$. For convenience, we have $w = {g_\varphi }({\mathcal{T}})$.

%% -Algorithm 1-
\begin{algorithm}[t]
 \caption{Pseudo code of episodic meta-learning with ASr}
  \label{Algorithm 1}
  \tcc{INITIALIZATION}
  Initialize the meta-learning model $f$ and ASr $g_{\varphi}$ \;
  Initialize the importance of tasks $w$\;
  \tcc{EPISODIC META-LEARNING}
  \For{each episode}{
    Select $N^{pool}$ tasks via $w$\;
    \For{$\mathcal{T}_{i}$ in the candidate task pool}{
    \tcc{GET TASK WEIGHT $w = {g_\varphi }({\mathcal{T}})$}
    Compute task diversity $t_{dg}^i$ via Eq.\ref{eq:task diversity}\;
    Compute task entropy $t_{et}^i$ via Eq.\ref{eq:task entopy}\;
    Compute task difficulty $t_{df}^i$ via Eq.\ref{eq:task difficulty}\;
    Obtain $t^i=(t_{dg}^i, t_{et}^i, t_{df}^i)$\;
    Compute the importance of $\mathcal{T}_{i}$ with ${w_i} = g_\varphi(t^i)$\;
    \tcc{INNER LOOP OPTIMIZATION}
    Update task-specific model $f*$ with $w_i$ and the support set $\mathcal{D}_i^s $ via Eq.\ref{eq:meta}\;
    }
    \tcc{OUTER LOOP OPTIMIZATION}
    Obtain $w = \{ {w_1},...,{w_{{N^{pool}}}}\}$\;
    Update meta-learning model $f$ with $w$ via Eq.\ref{eq:meta}\;
    \tcc{UPDATE ASr}
    Fix $f$ and update $g_{\varphi}$ via Eq.\ref{eq:ASr}\;
  }
\end{algorithm}

%%%%%%%%%%%%%%%%%%%%%%%%%%%%%%%%%%
%            5.2
%%%%%%%%%%%%%%%%%%%%%%%%%%%%%%%%%%
\subsection{Overall Objective}
\label{sec:7.2}
The key idea behind the optimization objective of ASr is that the quality of the task distribution generated by ASr can be improved by using the feedback from the meta-learning performance. To achieve this, ASr defines its loss as a function of how well the meta-learning model, which is guided by the task distribution, performs on the primary training tasks.

Given a task pool ${\mathcal{T}}=\{ {{\mathcal{T}_1},...,{\mathcal{T}_{N^{pool}}}} \}$, the training of the ASr-based meta-learning model alongside the ASr $g_\varphi$ has two stages per episode. In the first stage of each episode, the meta-learning model is trained based on the task distribution determined by $g_\varphi$. We can obtain the meta-objective as follows:
\begin{equation}\label{eq:meta}
    \begin{array}{l}
\mathop {\min }\limits_f \sum\limits_{i = 1,\mathcal{T}_i\in \mathcal{T}, {w_i} \in w}^{N^{pool}} {\mathcal{L}(w_i,\mathcal{D}_i^q,{f^ *_i })} \\[15pt]
s.t.\quad f^ *_i = f - \lambda {\nabla _f}{\mathcal{L}(\mathcal{D}_i^s,f)} \\[8pt]
\quad\quad w = {g_\varphi }(\mathcal{T})
\end{array}
\end{equation}
where $\lambda$ is the learning rate during the meta-learning optimization phase.

In the second stage, $g_\varphi$ is then updated by encouraging the task distribution to be chosen such that, if $f$ are to be trained using the task distribution, the performance of $f$ will be maximized on this episode. Leveraging the performance of $f$ to train $g_\varphi$ can be considered a form of meta-learning. At the same time, we theoretically analyze in Section \ref{sec:6} that high-quality tasks will promote model performance. Therefore, this updating mechanism forces the model to automatically learn the correct weight distribution: high-quality tasks have higher weights and low-quality tasks have lower weights. In other words, we use a bi-level optimization mechanism to incorporate the constraints of high-quality tasks with higher weights and low-quality tasks with lower weights into model learning. Therefore, to update $g_\varphi$, we define the objective as follows:
\begin{equation}\label{eq:ASr}
    \begin{array}{l}
\mathop {\min }\limits_{{g_\varphi }} \sum\limits_{i = 1,{\mathcal{T}_i} \in \mathcal{T},{w_i} \in w}^{N^{pool}} {\mathcal{L}(w_i,\mathcal{D}_i^q,{f^*_i })} \\[10pt]
s.t. \quad f^ *_i = f - \lambda {\nabla _f}{\mathcal{L}(\mathcal{D}_i^s,f)} \\[8pt]
\quad\quad w_i = {g_\varphi }(\mathcal{T}_i)
\end{array}
\end{equation}
where $\lambda$ is the learning rate during the meta-learning optimization phase, and the task-specific model $f_i^*$ is the same as illustrated in Equation \ref{eq:meta}. It's worth noting that the optimization process of ASr can be seen as a straightforward and versatile meta-learning algorithm. This means that ASr has the potential to deliver strong generalization performance.

%%%%%%%%%%%%%%%%%%%%%%%%%%%%%%%%%%%%%%%%%%%%%%%%%%%%%%%%%%%%%%%%%%%%%%%%%%%%%%%%%%%%%%%%%%%%%%%%%%%
%                                         8
%%%%%%%%%%%%%%%%%%%%%%%%%%%%%%%%%%%%%%%%%%%%%%%%%%%%%%%%%%%%%%%%%%%%%%%%%%%%%%%%%%%%%%%%%%%%%%%%%%%

\section{Experiments}
\label{sec:8}
In this section, we first introduce some important implementation details in the experiment. Then, we give a brief description of the datasets involved in the experiment and conduct extensive experiments on diverse meta-learning problems to evaluate the effect of ASr, including standard few-shot learning, cross-domain few-shot learning, multi-domain few-shot learning, and few-shot class-incremental learning. Next, we perform ablation studies to evaluate the effects of the three proposed three measurements, as well as their roles in ASr. Finally, we summarize the aforementioned experimental conclusions and conduct visualization on generalization, convergence, and sampling weights for further analysis. It is important to note that all the experimental results reported in all tables are the average of five independent experiments.

%%%%%%%%%%%%%%%%%%%%%%%%%%%%%%%%%%
%            8.1
%%%%%%%%%%%%%%%%%%%%%%%%%%%%%%%%%%
\subsection{Implementation Details}
\label{sec:8.1}
\noindent{\bf Experimental Setup}. Within the framework of meta-learning, we choose the Conv4 \citep{vinyals2016matching} as the structure of the encoder. This backbone remains fixed throughout the training process. Following the convolution and filtering stages, we apply a sequence of operations including batch normalization, ReLU non-linear activation, and $2\times 2$ max pooling, which is implemented through stride convolution. Transitioning to the optimization stage, we employ the Adam optimizer \citep{kingma2014adam} for model training. The momentum and weight decay values are selected as 0.9 and $10^{-4}$, respectively. Across all experiments, the initial learning rate is set at 0.4, offering the flexibility for linear scaling when required. Furthermore, the running epoch size is established at 150, with batch sizes of either 32 or 16. All experimental procedures are executed using NVIDIA RTX 4090 GPUs.

\noindent{\bf Evaluation Metrics}. We employ the widely adopted meta-learning evaluation protocols, wherein our model undergoes training on the base classes, subsequently being subjected to evaluation on the novel classes. For classification tasks, we undertake the random sampling of a limited number of examples (either 1-shot or 5-shot) from each novel class. The ensuing test accuracy is then computed using these samples, enabling us to assess the model's capacity for generalization to unfamiliar classes. For regression tasks, we similarly conduct random sampling of test samples, while adopting the Mean Squared Error (MSE) as the evaluative metric. For some special scenarios, such as drug activity prediction, we use its dedicated evaluation indicators and explain them in the corresponding subsections.

%%%%%%%%%%%%%%%%%%%%%%%%%%%%%%%%%%%%%%%%%%%%%%%%%%%%%%%%%%%%%%%%%%%%%%%%%%%%%%%%%%%%%%%%%%%%%%%%%%%
%                                         6
%%%%%%%%%%%%%%%%%%%%%%%%%%%%%%%%%%%%%%%%%%%%%%%%%%%%%%%%%%%%%%%%%%%%%%%%%%%%%%%%%%%%%%%%%%%%%%%%%%%

%%%%%%%%%%%%%%%%%%%%%%%%%%%%%%%%%%
%            8.2
%%%%%%%%%%%%%%%%%%%%%%%%%%%%%%%%%%
\subsection{Standard Few-shot Learning}
\label{sec:8.2}
We conduct experiments in six standard few-shot learning scenarios to evaluate the performance of our ASr, including image classification, regression, drug activity prediction, pose prediction, object detection, and semantic segmentation.

%%%%%%%%% 8.2.1 %%%%%%%%%%%%%%%
\subsubsection{Image Classification}
\label{sec:8.2.1}
\noindent{\bf Datasets}.
We evaluate the effectiveness of ASr on four benchmark datasets: (i) miniImagenet \citep{vinyals2016matching} consists of 100 classes with 50,000/10,000 training/testing images, split into 64/16/20 classes for meta-training/validation/testing; (ii) Omniglot \citep{lake2019omniglot} contains 1,623 characters from 50 different alphabets; (iii) tieredImagenet \citep{ren2018meta} contains 779,165 images and 391/97/160 classes for meta-training/validation/testing; and (iv) CIFAR-FS \citep{bertinetto2018meta} contains 100 classes with 600 images per class, split into 64/16/20 classes for meta-training/validation/testing.

\noindent{\bf Frameworks}.  
We group the meta-learning frameworks into three categories: (i) optimization-based methods, i.e., MAML \citep{finn2017model}, Reptile \citep{nichol2018reptile}, and MetaOptNet \citep{lee2019meta}; (ii) metric-based methods, i.e., ProtoNet \citep{snell2017prototypical}, MatchingNet \citep{vinyals2016matching}, and RelationNet \citep{sung2018learning}; and (iii) bayesian-based methods, i.e., CNAPs \citep{requeima2019fast} and SCNAP \citep{bateni2020improved}. We also introduce meta-learning methods particularly designed for few-shot learning, i.e., SimpleShot \citep{wang2019simpleshot}, SUR \citep{mangla2020charting}, and PPA \citep{qiao2018few}. More details are provided in Appendix \ref{sec:appendix_2}.

\noindent{\bf Results}.
Figure \ref{fig:3} illustrates the comparison results on miniImagenet between different samplers. The results show that compared to Uniform sampler, the samplers that rely on task diversity do not achieve performance improvement on all models, while our ASr does. Table \ref{table:1} summarizes more standard few-shot learning results for image classification. We find that different meta-learning models do not show a clear preference for task diversity, for example, ProtoNet performs better with high-diversity task samplers on tieredImagenet and CIFAR-FS datasets, but prefers Uniform sampler on miniImagenet. No sampler can achieve a universal effect improvement. Our ASr achieves state-of-the-art (SOTA) performance on all datasets and frameworks, demonstrating its high effectiveness and compatibility.

%%%%%%%%% 8.2.2 %%%%%%%%%%%%%%%
\subsubsection{Regression}
\label{sec:8.2.2}
\noindent{\bf Datasets}.
We calculate the Mean Square Error (MSE) on two regression datasets: (i) Sinusoid dataset \citep{finn2017model} is a simple synthetic dataset where the data points are generated according to the law of the sine function. The data for each task in our experiment is generated in the form of $A\sin \omega x + b + \epsilon $, where $A \in \left [ 0.1, 5.0 \right ] $, $\omega \in \left [ 0.5, 2.0 \right ]$, and $b \in \left [ 0,2\pi \right ] $; and (ii) Harmonic dataset \citep{lacoste2018uncertainty} is also a simple synthetic dataset, but sampled from a sum of two sine waves with different phase, amplitude, and a frequency ratio of 2: $f(x)=a_1 sin(\omega x+b_1)+a_2 sin(2 \omega x+b_2)$, where $y\sim \mathcal{N} (f(x), \sigma_y^2)$. Each task in Harmonic is sampled with $\omega \sim \mathcal{U} (5,7)$, $(b_1, b_2)\sim \mathcal{U} (0,2\pi )^2$, and $(a_1, a_2)\sim  \mathcal{N} (0,1)^2$.

\noindent{\bf Frameworks}.  
Following \citep{lacoste2018uncertainty}, we only use optimization-based methods, i.e., MAML and Reptile as backbone frameworks.

\noindent{\bf Results}.
Table \ref{table:regression} shows the performance of meta-learning models with different samplers on regression. Our ASr demonstrates excellent performance on both Sinusoid and Harmonic datasets, leading to similar conclusions as our classification experiments.

%%%%%%%%%%%% standard %%%%%%%%%%%%%%%%
\begin{table}
\begin{center}
\caption{Few-shot regression results (MSE) on Sinusoid and Harmonic. The best results of each benchmark dataset are highlighted in \textbf{bold}.}
\label{table:regression}
\begin{tabular}{l|l|ccc}
\toprule[1.2pt]
{\textbf{Datasets}}& {\textbf{Samplers}} & {\textbf{MAML}} &  \textbf{Reptile} \\
\midrule
\multirow{10}{*}{Sinusoid} & Uniform & 0.94 $\pm$ 0.04 & 0.41 $\pm$ 0.06 \\
        & NDT & 0.81 $\pm$ 0.04 & 0.37 $\pm$ 0.05 \\
        & NDE & 0.76 $\pm$ 0.05 & 0.37 $\pm$ 0.04 \\
        & NDTE & 0.95 $\pm$ 0.06 & 0.39 $\pm$ 0.04 \\
        & SEU & 1.89 $\pm$ 0.05 & 1.02 $\pm$ 0.13 \\
        & OHTM & 1.28 $\pm$ 0.09 & 0.83 $\pm$ 0.08 \\
        & OWHTM & 1.36 $\pm$ 0.11 & 0.83  $\pm$ 0.08 \\
        & GCP & 1.01 $\pm$ 0.03 & 0.59 $\pm$ 0.06 \\
        & DATS & 0.94 $\pm$ 0.10 & 0.39 $\pm$ 0.07 \\
        & ASr & \textbf{0.69 $\pm$ 0.02} & \textbf{0.36 $\pm$ 0.05} \\
\midrule
\multirow{10}{*}{Harmonic} & Uniform & 0.99 $\pm$ 0.05 & 1.19 $\pm$ 0.11 \\
        & NDT & 1.01 $\pm$ 0.04 & 1.22 $\pm$ 0.12 \\
        & NDE & 0.98 $\pm$ 0.05 & 1.18 $\pm$ 0.09 \\
        & NDTE & 0.97 $\pm$ 0.06 & 1.14 $\pm$ 0.08  \\
        & SEU & 1.10 $\pm$ 0.08 & 1.17 $\pm$ 0.08 \\
        & OHTM & 1.06 $\pm$ 0.09 & 1.16 $\pm$ 0.12 \\
        & OWHTM & 1.07 $\pm$ 0.09 & 1.16 $\pm$ 0.12 \\
        & GCP & 0.99 $\pm$ 0.09 & 1.12 $\pm$ 0.03 \\
        & DATS & 0.97 $\pm$ 0.05 & 1.16 $\pm$ 0.09 \\
        & ASr & \textbf{0.95 $\pm$ 0.05} & \textbf{1.11 $\pm$ 0.09} \\
\hline
\end{tabular}
\end{center}
\end{table}

\begin{table*}
\begin{minipage}{0.65\textwidth}
\begin{center}
\caption{Standard few-shot learning results (accuracy $\pm $ 95\% confidence interval) on drug activity prediction. ``Mean", ``Mde.", and ``$> 0.3$" are the mean, the median value of $R^2$, and the number of $R^2> 0.3$.}
\label{table:sfsl-drug}
\resizebox{\linewidth}{!}{
\begin{tabular}{l|l|ccc|ccc|ccc|ccc}
\toprule[1.2pt]
& \multirow{2}{*}{\textbf{Samplers}} & \multicolumn{3}{c|}{\textbf{Group 1}} & \multicolumn{3}{c|}{\textbf{Group 2}} & \multicolumn{3}{c|}{\textbf{Group 3}} & \multicolumn{3}{c}{\textbf{Group 4}} \\ 
  &  & \textbf{Mean} & \textbf{Med.} & \textbf{$>$ 0.3} & \textbf{Mean} & \textbf{Med.} & \textbf{$>$ 0.3} & \textbf{Mean} & \textbf{Med.} & \textbf{$>$ 0.3} & \textbf{Mean} & \textbf{Med.} & \textbf{$>$ 0.3} \\
\midrule
% \multirow{10}{*}{\rotatebox{90}{miniImagenet\to CUB}}
\multirow{10}{*}{\rotatebox{90}{MAML}}
& \textbf{Uniform} & 0.372 & 0.310 & 50 & 0.371 & 0.240 & 45 & 0.322 & 0.254 & 43 & 0.341 & 0.280 & 47 \\
& \textbf{NDT} & 0.355 & 0.281 & 47 & 0.362 & 0.233 & 40 & 0.311 & 0.238 & 42 & 0.318 & 0.271 & 45 \\
& \textbf{NDE} & 0.356 & 0.285 & 48 & 0.366 & 0.227 & 41 & 0.305 & 0.244 & 41 & 0.327 & 0.264 & 45 \\
& \textbf{NDTE} & 0.358 & 0.246 & 45 & 0.348 & 0.226 & 41 & 0.298 & 0.215 & 38 & 0.316 & 0.254 & 44 \\
& \textbf{SEU} & 0.354 & 0.255 & 45 & 0.354 & 0.211 & 40 & 0.302 & 0.228 & 40 & 0.312 & 0.265 & 44 \\
& \textbf{OHTM} & 0.396 & 0.315 & 50 & 0.384 & 0.254 & 45 & 0.327 & 0.250 & 45 & 0.335 & 0.287 & 48 \\
& \textbf{OWHTM} & 0.397 & 0.320 & 51 & 0.389 & 0.254 & 45 & 0.325 & 0.254 & 45 & 0.341 & 0.294 & 48 \\
& \textbf{s-DPP} & 0.384 & 0.318 & 50 & 0.376 & 0.235 & 42 & 0.312 & 0.238 & 42 & 0.336 & 0.274 & 44 \\
& \textbf{d-DPP}& 0.384 & 0.324 & 50 & 0.369 & 0.248 & 44 & 0.318 & 0.245 & 42 & 0.334 & 0.286 & 45 \\
& \textbf{ASr} & \textbf{0.405} & \textbf{0.323} & \textbf{54} & \textbf{0.386} & \textbf{0.257} & \textbf{46} & \textbf{0.335} & \textbf{0.286} & \textbf{48} & \textbf{0.354} & \textbf{0.284} & \textbf{48} \\
\midrule
\multirow{10}{*}{\rotatebox{90}{ANIL}}
& \textbf{Uniform} & 0.354 & 0.297 & 50 & 0.301 & 0.252 & 46 & 0.317 & 0.300 & 49 & 0.324 & 0.297 & 48 \\
& \textbf{NDT} & 0.345 & 0.284 & 48 & 0.277 & 0.241 & 44 & 0.302 & 0.289 & 47 & 0.305 & 0.281 & 46 \\
& \textbf{NDE} & 0.356 & 0.285 & 49 & 0.294 & 0.246 & 44 & 0.311 & 0.298 & 48 & 0.312 & 0.283 & 48 \\
& \textbf{NDTE} & 0.342 & 0.281 & 47 & 0.296 & 0.245 & 45 & 0.315 & 0.291 & 48 & 0.320 & 0.300 & 48 \\
& \textbf{SEU} & 0.337 & 0.280 & 47 & 0.291 & 0.250 & 45 & 0.302 & 0.289 & 47 & 0.325 & 0.294 & 48 \\
& \textbf{OHTM} & 0.346 & 0.302 & 50 & 0.301 & 0.267 & 47 & 0.315 & 0.302 & 48 & 0.336 & 0.300 & 49 \\
& \textbf{OWHTM} & 0.351 & 0.306 & 51 & 0.304 & 0.268 & 47 & 0.320 & 0.309 & 50 & 0.334 & 0.302 & 49 \\
& \textbf{s-DPP} & 0.358 & 0.309 & 51 & 0.299 & 0.259 & 46 & 0.321 & 0.297 & 50 & 0.328 & 0.300 & 49 \\
& \textbf{d-DPP}& 0.356 & 0.302 & 51 & 0.301 & 0.257 & 47 & 0.319 & 0.296 & 49 & 0.321 & 0.298 & 48 \\
& \textbf{ASr} & \textbf{0.361} & \textbf{0.315} & \textbf{53} & \textbf{0.311} & \textbf{0.258} & \textbf{48} & \textbf{0.331} & \textbf{0.304} & \textbf{50} & \textbf{0.326} & \textbf{0.304} & \textbf{50} \\
\bottomrule
\end{tabular}}
\end{center}
\end{minipage}%
\hfill
\begin{minipage}{0.3\textwidth}
\begin{center}
\caption{Standard few-shot learning results (MSE) on pose prediction.}
\label{table:sfsl-pose}
\resizebox{\linewidth}{!}{
\begin{tabular}{l|l|c|c}
\toprule[1.2pt]
& \textbf{Samplers} & \textbf{10-shot} & \textbf{15-shot}\\
\midrule
% \multirow{10}{*}{\rotatebox{90}{miniImagenet\to CUB}}
\multirow{10}{*}{\rotatebox{90}{MAML}}
& \textbf{Uniform} & 3.121 $\pm$ 0.239 & 2.501 $\pm$ 0.194 \\
& \textbf{NDT} & 3.321 $\pm$ 0.252 & 2.630 $\pm$ 0.202 \\
& \textbf{NDE} & 3.197 $\pm$ 0.230 & 2.596 $\pm$ 0.194 \\
& \textbf{NDTE} & 3.478 $\pm$ 0.166 & 2.832 $\pm$ 0.210 \\
& \textbf{SEU} & 3.570 $\pm$ 0.214 & 2.932 $\pm$ 0.226 \\
& \textbf{OHTM} & 3.012 $\pm$ 0.206 & 2.435 $\pm$ 0.216 \\
& \textbf{OWHTM} & 3.007 $\pm$ 0.197 & 2.394 $\pm$ 0.201 \\
& \textbf{s-DPP} & 2.913 $\pm$ 0.165 & 2.247 $\pm$ 0.184 \\
& \textbf{d-DPP}& 2.881 $\pm$ 0.194 & 2.203 $\pm$ 0.206 \\
& \textbf{ASr} & \textbf{2.455 $\pm$ 0.209} & \textbf{2.132 $\pm$ 0.189} \\
\midrule
\multirow{10}{*}{\rotatebox{90}{ANIL}}
& \textbf{Uniform} & 6.851 $\pm$ 0.385 & 6.532 $\pm$ 0.392 \\
& \textbf{NDT} & 6.932 $\pm$ 0.296 & 6.627 $\pm$ 0.341 \\
& \textbf{NDE} & 6.917 $\pm$ 0.334 & 6.596 $\pm$ 0.280 \\
& \textbf{NDTE} & 7.202 $\pm$ 0.256 & 6.832 $\pm$ 0.269 \\
& \textbf{SEU} & 7.416 $\pm$ 0.239 & 7.019 $\pm$ 0.251 \\
& \textbf{OHTM} & 6.809 $\pm$ 0.350 & 6.472 $\pm$ 0.329 \\
& \textbf{OWHTM} & 6.794 $\pm$ 0.331 & 6.456 $\pm$ 0.301 \\
& \textbf{s-DPP} & 6.962 $\pm$ 0.305 & 6.746 $\pm$ 0.347 \\
& \textbf{d-DPP}& 6.918 $\pm$ 0.299 & 6.744 $\pm$ 0.342 \\
& \textbf{ASr} & \textbf{6.651 $\pm$ 0.277} & \textbf{6.448 $\pm$ 0.309} \\
\bottomrule
\end{tabular}}
\end{center}
\end{minipage}
\end{table*}

\begin{table}
\begin{center}
\caption{Standard few-shot learning results (nAP) on object detection. The ``-'' denotes that there is no need to select specific task samplers for the corresponding method.}
\label{tab:objective}
\resizebox{\linewidth}{!}{
\begin{tabular}{l|l|ccc}
\toprule[1.2pt]
{\textbf{Model}}& {\textbf{Samplers}} & {\textbf{5-way 1-shot}} &  \textbf{5-way 3-shot} & \textbf{5-way 5-shot} \\
\midrule
Faster RCNN & $-$ & 14.78 & 20.34 & 26.89 \\
LSTD & $-$ & 17.66 & 22.37 & 29.00 \\
FRCN-PN & $-$ & 16.78 & 21.51 & 26.01 \\
\midrule
\multirow{8}{*}{MetaRCNN} & Uniform & 19.03 & 24.51 & 31.23 \\
        & NDT & 17.48 & 23.00 & 29.83 \\
        & NDE & 17.96 & 23.88 & 30.06 \\
        & SEU & 18.02 & 24.01 & 30.15 \\
        & OHTM & 18.97 & 24.31 & 31.44 \\
        & GCP & 19.22 & 25.00 & 32.65 \\
        & DATS & 19.26 & 24.85 & 32.64 \\
        & ASr & \textbf{21.38} & \textbf{26.40} & \textbf{33.96} \\
\midrule
\multirow{6}{*}{MetaDet} & Uniform & 18.01 & 22.47 & 30.00 \\
        & NDT & 16.37 & 21.05 & 27.94 \\
        & NDE & 16.85 & 21.32 & 27.96 \\
        & OHTM & 18.58 & 23.01 & 31.88 \\
        & DATS & 20.13 & 23.38 & 31.44 \\
        & ASr & \textbf{21.00} & \textbf{24.89} & \textbf{32.97} \\
\hline
\end{tabular}}
\end{center}
\end{table}

\begin{table}
\begin{center}
\caption{Standard few-shot learning results (mean mIoU) on semantic segmentation. The ``-'' denotes that there is no need to select specific task samplers for the corresponding method.}
\label{tab:seg}
\resizebox{0.9\linewidth}{!}{
\begin{tabular}{l|l|ccc}
\toprule[1.2pt]
{\textbf{Model}}& {\textbf{Samplers}} & {\textbf{1-shot results}} &  \textbf{5-shot results} \\
\midrule
PANet & $-$ & 48.10 & 55.70 \\
FWB & $-$ & 51.90 & 55.10 \\
CRNet & $-$ & 55.20 & 58.50 \\
PFENet & $-$ & 59.70 & 64.10 \\
\midrule
\multirow{8}{*}{BAM} & Uniform & 64.41 & 68.76
 \\
        & NDT & 59.37 & 66.48 \\
        & NDE & 60.20 & 67.00 \\
        & SEU & 60.38 & 67.17 \\
        & OHTM & 64.20 & 68.79 \\
        & GCP & 65.00 & 69.38 \\
        & DATS & 65.16 & 69.24 \\
        & ASr & \textbf{65.82} & \textbf{70.03} \\
\hline
\end{tabular}}
\end{center}
\end{table}

%%%%%%%%% 8.2.3 %%%%%%%%%%%%%%%
\subsubsection{Drug Activity Prediction}
\label{sec:8.2.3}
\noindent{\bf Datasets}.
The dataset contains 4,276 tasks sampled from the large-scale database proposed in \citep{martin2019all}. It aims to predict the activity of compounds on specific target proteins. The evaluation metric is the squared Pearson correlation coefficient ($R^2$) between the predicted and actual values for each task, which is different from other tasks. We report the mean and median of $R^2$, and the count of $R^2$ values exceeding 0.3, which is a reliable pharmacological indicator.

\noindent{\bf Frameworks}.  
Following \citep{yao2021improving}, we use MAML and ANIL \citep{raghu2019rapid} as meta-learning baselines for evaluation.

\noindent{\bf Results}.
Table \ref{table:sfsl-drug} shows the standard few-shot learning results of drug activity prediction. ASr achieves better performance compared to SOTA samplers in all groups. For the reliability index $R^2 > 0.3$, our ASr achieves an obvious improvement effect, improving by 2. This achievement reflects the applicability of ASr in different domains, as well as the effectiveness of the three measurements mentioned in Subsection \ref{sec:5}.

%%%%%%%%% 8.2.4 %%%%%%%%%%%%%%%
\subsubsection{Pose Prediction}
\label{sec:8.2.4}
\noindent{\bf Datasets}.
The data in this experiment is generated using the Pascal 3D dataset \citep{xiang2014beyond}. We randomly select 50 objects for meta-training and additional 15 objects for meta-testing.

\noindent{\bf Frameworks}.  
Similar to drug activity prediction, we use MAML and ANIL as backbone frameworks.

\noindent{\bf Results}.
Table \ref{table:sfsl-pose} summarizes the standard few-shot learning results of pose prediction. Compared to the regression problem mentioned in Subsection \ref{sec:8.2.2}, this problem is more difficult due to reasons such as complex data or diverse classes. But our ASr still achieves excellent results, reducing MSE by an average of 0.12 more than other samplers, proving its compatibility.

\subsubsection{Object Detection}
\label{sec:8.2.5}

\noindent{\bf Datasets}.
The data in this experiment is generated using the Pascal VOC 2007 \citep{oquab2014learning}, which is a popular benchmark for few-shot object detection. It has 20 categories with 5k images for training and
5k images for testing. The categories for training and testing are without overlap. The mean Average Precision
(nAP) over the selected categories is used as the evaluation metric.
We use the same experimental settings as \citep{wu2020meta}.

\noindent{\bf Frameworks}.  
We use the Meta-RCNN \citep{wu2020meta} and MetaDet \citep{huang2022survey} as the meta-learning baselines. We compare their performance using different samplers with multiple methods, i.e., Faster-RCNN \citep{sun2018face}, LSTD \citep{chen2018lstd}, and FRCN-PN \citep{snell2017prototypical,wu2020meta}.

\noindent{\bf Results}.

Table \ref{tab:objective} shows the comparative experimental results. Compared with the aforementioned classification tasks, few-shot object detection needs to further consider the location information of the object and has more requirements for task features \citep{lang2022learning}. The experimental results show that ASr can still achieve stable improvement of the meta-learning method, and can even achieve similar effects to the SOTA object detection method with few updates.

\subsubsection{Semantic Segmentation}
\label{sec:8.2.6}
\noindent{\bf Datasets}.
We choose PASCAL-$5^i$ \citep{shaban2017one} as the benchmark dataset. PASCAL-$5^i$ is created from PASCAL VOC 2012 \citep{everingham2010pascal} with additional annotations from SDS \citep{hariharan2011semantic}. The object categories are equally distributed across four partitions, and the experiments are carried out using a cross-validation manner following \citep{lang2022learning}. In each partition, we randomly select 1,000 pairs of support and query images for validation purposes. In this experiment, we use mIoU \citep{lang2023base} as the evaluation metric.

\noindent{\bf Frameworks}. 
We use BAM \citep{lang2022learning} as the meta-learning baseline, and compare its performance using different samplers with multiple semantic segmentation methods, i.e., PANet \citep{wang2019panet}, FWB \citep{wang2019panet}, CRNet \citep{liu2020crnet}, and PFENet \citep{tian2020prior}.

\noindent{\bf Results}.

Table \ref{tab:seg} shows the performance of the meta-learning model using different samplers. From the results, we find that: (i) high-diversity samplers can often achieve better performance improvements; (ii) ASr achieves similar or even better results than the optimal sampler. Combining the above results, we can conclude that ASr achieves comparable robustness.

%%%%%%%%%%%% cross-domain %%%%%%%%%%%%%%
\begin{table*}
\begin{center}
\caption{Cross-domain few-shot classification results (accuracy $\pm $ 95\% confidence interval) with different task samplers on miniImagenet to CUB \citep{hilliard2018few} and miniImagenet to Places \citep{zhou2017places}. The best results are highlighted in \textbf{bold}.}
\label{table:cross-domain}
\resizebox{\linewidth}{!}{
\begin{tabular}{l|l|cc|cc|cc|ccc}
\toprule[1.2pt]
& \textbf{Samplers} & MAML & Reptile & ProtoNet & MatchingNet & CNAPs & SCNAP & Baseline++ & S2M2 & MetaQDA \\
\midrule
% \multirow{10}{*}{\rotatebox{90}{miniImagenet\to CUB}}
\multirow{10}{*}{\rotatebox{90}{miniImagenet $\to$ CUB}}
& \textbf{Uniform} & $33.62\pm 0.18$ & $36.58\pm 0.78$ & $33.37\pm  0.42$ & $38.19\pm 0.63$ & $32.72\pm 0.57$ & $41.98\pm 0.22$ & \textbf{39.23 $\pm$ 0.72} & $48.24\pm 0.84$ & $48.88\pm 0.64$ \\
& \textbf{NDT} & $28.46\pm 0.28$ & $32.85\pm 0.40$ & $24.75\pm  0.51$ & $31.24\pm 0.25$ & $21.83\pm 0.71$ & $35.82\pm 0.15$ & $24.84\pm 0.37$ & $29.23\pm 0.98$ & $34.89\pm 0.37$ \\
& \textbf{NDE} & $32.64\pm 0.79$ & $36.47\pm 0.72$ & $26.72\pm  0.62$ & $35.24\pm 0.56$ & $28.29\pm 0.48$ & $35.91\pm 0.44$ & $23.18 \pm 0.57$ &  $35.28\pm 0.90$ & $40.28\pm 0.39$ \\
& \textbf{NDTE} & $34.01\pm0.42$ & $37.98\pm 0.21$ & $29.18\pm  0.39$ & $36.93\pm 0.28$ & $30.47\pm 0.72$ & $38.98\pm 0.17$ & $25.96 \pm 0.46$ &  $37.82\pm 0.19$ & $42.28\pm 0.46$ \\
& \textbf{SEU} & $27.04\pm 0.71$ & $31.43\pm 0.89$ &  $20.04\pm  0.77$ & $36.20\pm 0.65$ & $31.12\pm 0.85$ & $35.82\pm 0.31$ & $25.83\pm 0.52$ & $30.18\pm 0.92$ & $35.09\pm 0.92$ \\
& \textbf{OHTM} & $32.74\pm 0.30$ & $34.84\pm 0.82$ &  $33.09\pm  0.40 $ & $33.42\pm 0.58$ & $32.40\pm 0.49$ & $38.29\pm 0.33$ & $38.60 \pm 0.47$ & $44.81\pm 0.32$ & $40.28\pm 0.39$ \\
& \textbf{OWHTM} & $32.93\pm 0.56$ & $34.27\pm 0.45$ &  $34.75\pm  0.92$ & $34.46\pm 0.84$ & $32.80\pm 0.90$ & $39.29\pm 0.20$ & $39.12\pm 0.37$ & $44.89\pm 0.32$ & $42.29\pm 0.25$ \\
& \textbf{s-DPP} & $33.41\pm 0.47$ & $35.10\pm 0.42$ &  $33.47\pm  0.78$ & $32.44\pm 0.65$ & $28.84\pm 0.92$ & $40.98\pm 0.20$ & $35.81\pm 0.90$ & $42.82\pm 0.73$ & $46.92\pm 0.92$ \\
& \textbf{d-DPP} & $33.55\pm 0.32$ & $34.90\pm 0.30$ &  $34.26\pm  0.42$ & $35.55\pm 0.34$ & $29.02\pm 0.51$ & $42.39\pm 0.20$ & $35.02\pm 0.50$ & $39.81\pm 0.83$ & $47.90\pm 0.82$ \\
% & \textbf{GCP} &  &  &  &  &  &  &  &  &  \\
% & \textbf{DATS} &  &  &  &  &  &  &  &  &  \\
& \textbf{ASr} & \textbf{35.98 $\pm$ 0.17} & \textbf{38.20 $\pm$ 0.27} & \textbf{35.33 $\pm$ 0.34} & \textbf{38.45 $\pm$ 0.17}  & \textbf{33.70 $\pm$ 0.38} & \textbf{44.82 $\pm$ 0.39} & 39.17 $\pm$ 0.39 & \textbf{48.29 $\pm$ 0.55} & \textbf{49.82 $\pm$ 0.46} \\
\midrule
\multirow{10}{*}{\rotatebox{90}{miniImagenet $\to$ Places}}
& \textbf{Uniform} & $29.84 \pm 0.56$ & $33.84\pm 0.48$ &  $30.91\pm  0.57$ & \textbf{33.28 $\pm$ 0.30} & $25.86\pm 0.45$ & $37.29\pm 0.82$ & $42.24\pm 0.37$ & $46.19\pm 0.24$ & $48.89\pm 0.49$ \\
& \textbf{NDT} & $20.27 \pm 0.48$ & $23.92 \pm 0.38$ &  $19.34\pm  0.58$ & $25.19\pm 0.41$ & $14.02\pm 0.45$ & $29.83\pm 0.82$ & $26.82\pm 0.90$ & $37.10\pm 0.42$ & $32.32\pm 0.59$ \\
& \textbf{NDE} & $25.58 \pm 0.76$ & $27.98 \pm 0.40$ &  $21.84\pm  0.45$ & $26.02\pm 0.62$ & $23.51\pm 0.65$ & $32.65\pm 0.37$ & $41.80\pm 0.78$ & $44.23\pm 0.87$ & $44.38\pm 0.59$ \\
& \textbf{NDTE} & $21.25 \pm 0.45$ & $31.98 \pm 0.43$ &  $25.44\pm  0.46$ & $26.25\pm 0.71$ & $20.86\pm 0.92$ & $35.27\pm 0.82$ & $35.58\pm 0.58$ & $41.72\pm 0.38$ & $48.83\pm 0.72$ \\
& \textbf{SEU} & $19.72 \pm 0.80$ & $28.29 \pm 0.47$ &  $15.24\pm  0.81$ & $25.46\pm 0.58$ & $22.79\pm 0.71$ & $33.28\pm 0.39$ & $30.73\pm 0.37$ & $38.83\pm 0.42$ & $46.38\pm 0.67$ \\
& \textbf{OHTM} & $26.78 \pm 0.27$ & $32.89 \pm 0.36$ &  $27.78\pm  0.89$ & $30.21\pm 0.40$ & $25.93\pm 0.75$ & $35.83\pm 0.41$ & $40.79\pm 0.40$ & $45.82\pm 0.35$ & $43.82\pm 0.28$ \\
& \textbf{OWHTM} & $26.58 \pm 0.78$ & $31.98 \pm 0.55$ &  $28.92\pm  0.50$ & $30.89\pm 0.42$ & $26.29\pm 0.56$ & $36.28\pm 0.39$ & $42.48\pm 0.37$ & $46.32\pm 0.58$ & $43.82\pm 0.74$ \\
& \textbf{s-DPP} & $29.28\pm 0.53$ & $29.23\pm 0.49$ &  $28.79\pm  0.23$ & $26.44\pm 0.34$ & $24.83\pm 0.29$ & $38.23\pm 0.52$ & $40.32\pm 0.88$ & $45.23\pm 0.18$ & $45.82\pm 0.35$ \\
& \textbf{d-DPP} & $27.70 \pm 0.75$ & $30.10\pm 0.43$ &  $28.57\pm  0.91$ & $28.15\pm 0.18$ & $24.27\pm 0.93$ & $37.89\pm 0.48$ & $41.75\pm 0.83$ & $43.75\pm 0.68$ & $42.84\pm 0.23$ \\
% & \textbf{GCP} &  &  &  &  &  &  &  &  &  \\
% & \textbf{DATS} &  &  &  &  &  &  &  &  &  \\
& \textbf{ASr} & \textbf{30.41 $\pm$ 0.65} & \textbf{35.98 $\pm$ 0.19} &  \textbf{32.85 $\pm$  0.19} & 32.93  $\pm$ 0.09 & \textbf{26.24 $\pm$ 0.55} & \textbf{40.89 $\pm$ 0.23} & \textbf{42.23 $\pm$ 0.39} & \textbf{46.90 $\pm$ 0.33} & \textbf{48.91 $\pm$ 0.44} \\
\bottomrule
\end{tabular}}
\end{center}
\end{table*}

%%%%%%%%%%% multi-domain %%%%%%%%%%%%%
\begin{table*}
\begin{center}
\caption{Multi-domain few-shot classification results (accuracy $\pm $ 95\% confidence interval) with different task samplers on Meta-Dataset. The values in the brackets indicate the improvement of performance using our ASr compared to the SOTA sampler. The overall results are not the average of ID (in-domain) results and OOD (out-of-domain) results, but rather obtained by training on all ten datasets of Meta-Dataset. The best results are highlighted in \textbf{bold}.}
\label{table:multi-domain}
\resizebox{\linewidth}{!}{
\begin{tabular}{l|l|c|cccc|cccc|cc|c}
\toprule[1.2pt]
& \multirow{2}{*}{\textbf{Model}} & \multicolumn{11}{c}{\textbf{Samplers}}\\  \cline{3-14}
 &  & \textbf{Uniform} & \textbf{NDT} & \textbf{NDE} & \textbf{NDTE} & \textbf{SEU} & \textbf{OHTM} & \textbf{OWHTM} & \textbf{s-DPP} & \textbf{d-DPP} & \textbf{GCP} & \textbf{DATS} & \textbf{ASr}\\
\midrule
\multirow{5}{*}{\rotatebox{90}{\textbf{overall}}}
& MAML & 24.51 $\pm$ 0.13 & 24.68 $\pm$ 0.14 & 24.65 $\pm$ 0.11 & 23.10 $\pm$ 0.17 & 22.41 $\pm$ 0.09 & 23.02 $\pm$ 0.10 & 23.46 $\pm$ 0.11 & 24.74 $\pm$ 0.08 & 23.67 $\pm$ 0.13 & 24.73 $\pm$ 0.12 & 24.36 $\pm$ 0.10 & \textbf{26.21 $\pm$ 0.18} (+1.47) \\
& Reptile & 59.78 $\pm$ 0.27 & 42.67 $\pm$ 0.30 & 59.65 $\pm$ 0.32 & 60.16 $\pm$ 0.24 & 53.64 $\pm$ 0.28 & 52.45 $\pm$ 0.25 & 52.37 $\pm$ 0.23 & 54.98 $\pm$ 0.28 & 49.05 $\pm$ 0.27 & 60.01 $\pm$ 0.22 & 59.82 $\pm$ 0.28 & \textbf{60.40 $\pm$ 0.25} (+0.62) \\
& ProtoNet & 37.92 $\pm$ 0.19 & 31.41 $\pm$ 0.15 & 34.78 $\pm$ 0.15 & 35.20 $\pm$ 0.17 & 35.63 $\pm$ 0.16 & 38.91 $\pm$ 0.18 & 39.22 $\pm$ 0.11 & 40.75 $\pm$ 0.14 & 38.83 $\pm$ 0.19 & 38.11 $\pm$ 0.17 & 37.99 $\pm$ 0.21 & \textbf{40.49 $\pm$ 0.16} (+1.27) \\
& MatchingNet & 60.27 $\pm$ 0.28 & 50.47 $\pm$ 0.25 & 55.69 $\pm$ 0.31 & 54.41 $\pm$ 0.27 & 52.02 $\pm$ 0.25 & 62.76 $\pm$ 0.24 & 62.04 $\pm$ 0.25 & 62.17 $\pm$ 0.27 & 61.56 $\pm$ 0.28 & 60.28 $\pm$ 0.26 & 59.97 $\pm$ 0.30 & \textbf{62.82 $\pm$ 0.30} (+0.06) \\
& CNAPs & 65.92 $\pm$ 0.32 & 61.15 $\pm$ 0.25 & 58.73 $\pm$ 0.21 & 57.71 $\pm$ 0.22 & 54.53 $\pm$ 0.22 & 49.89 $\pm$ 0.26 & 50.11 $\pm$ 0.31 & 47.38 $\pm$ 0.29 & 48.70 $\pm$ 0.24 & 64.93 $\pm$ 0.28 & 65.12 $\pm$ 0.29 & \textbf{65.97 $\pm$ 0.27} (+0.05) \\
\midrule
\multirow{5}{*}{\rotatebox{90}{\textbf{ID}}}
& MAML & 31.37 $\pm$ 0.09 & 32.73 $\pm$ 0.10 & 32.89 $\pm$ 0.11 & 29.83 $\pm$ 0.10 & 27.89 $\pm$ 0.09 & 28.39 $\pm$ 0.07 & 29.90 $\pm$ 0.09 & 30.43 $\pm$ 0.10 & 30.49 $\pm$ 0.11 & 32.00 $\pm$ 0.10 & 31.88 $\pm$ 0.07 & \textbf{33.17 $\pm$ 0.09} (+0.28) \\
& Reptile & 63.78 $\pm$ 0.28 & 58.90 $\pm$ 0.25 & 62.28 $\pm$ 0.26 & 63.47 $\pm$ 0.25 & 60.39 $\pm$ 0.27 & 59.89 $\pm$ 0.28 & 60.33 $\pm$ 0.28 & 62.19 $\pm$ 0.26 & 61.82 $\pm$ 0.26 & 64.46 $\pm$ 0.29 & 64.35 $\pm$ 0.19 & \textbf{65.01 $\pm$ 0.24} (+1.23) \\
& ProtoNet & 42.18 $\pm$ 0.17 & 35.92 $\pm$ 0.22 & 38.47 $\pm$ 0.19 & 39.38 $\pm$ 0.17 & 37.12 $\pm$ 0.17 & 41.89 $\pm$ 0.20 & 43.38 $\pm$ 0.21 & 42.37 $\pm$ 0.21 & 42.81 $\pm$ 0.19 & 42.66 $\pm$ 0.20 & 42.41 $\pm$ 0.17 & \textbf{43.78 $\pm$ 0.21} (+0.97) \\
& MatchingNet & 65.73 $\pm$ 0.20 & 58.56 $\pm$ 0.27 & 62.50 $\pm$ 0.28 & 62.18 $\pm$ 0.23 & 60.45 $\pm$ 0.26 & 64.62 $\pm$ 0.26 & 64.45 $\pm$ 0.25 & 64.52 $\pm$ 0.19 & 64.01 $\pm$ 0.24 & 65.38 $\pm$ 0.30 & 65.26 $\pm$ 0.15 & \textbf{65.82 $\pm$ 0.27} (+0.09) \\
& CNAPs & 68.47 $\pm$ 0.31 & 63.82 $\pm$ 0.27 & 64.89 $\pm$ 0.29 & 62.82 $\pm$ 0.24 & 59.25 $\pm$ 0.26 & 63.47 $\pm$ 0.27 & 65.17 $\pm$ 0.30 & 66.24 $\pm$ 0.22 & 65.40 $\pm$ 0.21 & 68.49 $\pm$ 0.30 & 68.26 $\pm$ 0.23 & \textbf{68.83 $\pm$ 0.25} (+0.36) \\
\midrule
\multirow{5}{*}{\rotatebox{90}{\textbf{OOD}}}
& MAML & 19.19 $\pm$ 0.10 & 17.31 $\pm$ 0.12 & 21.56 $\pm$ 0.09 & 20.78 $\pm$ 0.07 & 16.90 $\pm$ 0.11 & 16.81 $\pm$ 0.12 & 16.79 $\pm$ 0.10 & 22.82 $\pm$ 0.13 & 20.32 $\pm$ 0.11 & 19.81 $\pm$ 0.08 & 20.01 $\pm$ 0.12 & \textbf{24.45 $\pm$ 0.11} (+1.63) \\
& Reptile & 49.28 $\pm$ 0.11 & 39.73 $\pm$ 0.12 & 48.83 $\pm$ 0.09 & 51.29 $\pm$ 0.08 & 46.78 $\pm$ 0.12 & 48.82 $\pm$ 0.10 & 48.05 $\pm$ 0.11 & 47.82 $\pm$ 0.09 & 47.31 $\pm$ 0.10 & 49.85 $\pm$ 0.11 & 51.30 $\pm$ 0.10 & \textbf{55.09 $\pm$ 0.08} (+3.80) \\
& ProtoNet & 30.89 $\pm$ 0.11 & 21.39 $\pm$ 0.09 & 25.56 $\pm$ 0.12 & 22.56 $\pm$ 0.10 & 19.23 $\pm$ 0.14 & 27.78 $\pm$ 0.13 & 27.40 $\pm$ 0.12 & 23.10 $\pm$ 0.13 & 25.89 $\pm$ 0.10 & 31.82 $\pm$ 0.15 & 31.00 $\pm$ 0.12 & \textbf{35.28 $\pm$ 0.12} (+4.39) \\
& MatchingNet & 55.56 $\pm$ 0.19 & 43.20 $\pm$ 0.21 & 50.56 $\pm$ 0.21 & 47.15 $\pm$ 0.19 & 45.62 $\pm$ 0.20 & 55.57 $\pm$ 0.20 & 55.16 $\pm$ 0.23 & 49.46 $\pm$ 0.22 & 50.12 $\pm$ 0.22 & 55.28 $\pm$ 0.17 & 56.00 $\pm$ 0.20 & \textbf{59.71 $\pm$ 0.19} (+4.14) \\
& CNAPs & 55.51 $\pm$ 0.24 & 46.87 $\pm$ 0.25 & 49.24 $\pm$ 0.24 & 51.60 $\pm$ 0.23 & 48.66 $\pm$ 0.22 & 44.19 $\pm$ 0.27 & 44.38 $\pm$ 0.25 & 44.27 $\pm$ 0.27 & 43.45 $\pm$ 0.22 & 55.28 $\pm$ 0.20 & 55.36 $\pm$ 0.20 & \textbf{58.21 $\pm$ 0.26} (+2.70) \\
\bottomrule
\end{tabular}}
\end{center}
\end{table*}

%%%%%%%%%%%%%%%%%%%%%%%%%%%%%%%%%%
%            8.3
%%%%%%%%%%%%%%%%%%%%%%%%%%%%%%%%%%
\subsection{Cross-domain Few-shot Learning}
\label{sec:8.3}
The outstanding performance of meta-learning models with ASr has been confirmed in standard few-shot learning from above, which can yield good results with limited data. However, in scenarios where data collection is not feasible in real life, such as skin diseases and satellite images \citep{zheng2015methodologies, tang2012cross}, the situation becomes complicated. Therefore, the performance in cross-domain tasks is crucial as it determines the applicability of the learned models, and the key challenge is to learn more precise representations from the source domain and apply them to the target domain \citep{guo2020broader}. In order to learn a robust representation, we suppose that the selection of training tasks is crucial for cross-domain learning.

\noindent{\bf Datasets}.
Two benchmark datasets are chosen for cross-domain few-shot classification. CUB \citep{hilliard2018few} contains 11,788 images of 200 categories, split into 100/50/50 classes for meta-training/validation/testing. Places \citep{zhou2017places} contains more than 2.5 million images across 205 categories, split into 103/51/51 classes for meta-training/validation/testing. The models in this experiment are trained on miniImagenet, and evaluated on CUB and Places.

\noindent{\bf Frameworks}. 
To evaluate the generalization ability of meta-learning models, it is important that they transfer well and do not significantly degrade when deployed to an unseen domain \citep{luo2020generalizing}. Therefore, in addition to the models mentioned in Section \ref{sec:8.2.1}, we introduce Baseline++ \citep{chen2019closer}, S2M2 \citep{mangla2020charting}, and MetaQDA \citep{zhang2021shallow} for cross-domain classification.

\noindent{\bf Results}.
Table \ref{table:cross-domain} presents the performance of the models with different task samplers in cross-domain few-shot learning. ASr outperforms other samplers by at least 1\% across almost all datasets, highlighting the potential of task sampling to address domain shift. Besides, some meaningful conclusions are obtained: the effects of different samplers on the models varied, with no clear preference; Uniform sampler generally outperform other task-diversity-based samplers except our ASr. These findings suggest that enhancing cross-domain few-shot learning performance through adaptive task sampling is an effective way.

%%%%%%%%%%%%%%%%%%%%%%%%%%%%%%%%%%
%            8.4
%%%%%%%%%%%%%%%%%%%%%%%%%%%%%%%%%%
\subsection{Multi-domain Few-shot Learning}
\label{sec:8.4}
In order to assess the effectiveness of our ASr in multi-domains, we conduct experiments on the challenging Meta-dataset \citep{triantafillou2019meta} for evaluation.

\noindent{\bf Datasets}.
Meta-Dataset \citep{triantafillou2019meta} is a large benchmark for few-shot learning that includes 10 datasets from diverse domains. It is designed to represent a more realistic scenario as it does not restrict few-shot tasks to having fixed ways and shots. This dataset includes 10 diverse domains, with the first 8 in-domain (ID) datasets used for meta-training, i.e., ILSVRC, Omniglot, Aircraft, Birds, Textures, Quick Draw, Fungi, and VGG Flower, and the remaining 2 reserved for testing out-of-domain (OOD) performance, i.e., Traffic Signs and MSCOCO. We evaluate the performance of the meta-learning models in 10 domains with different samplers across all 10 datasets.

\noindent{\bf Frameworks}.
We use the three types of frameworks described in Subsection \ref{sec:8.2.1}.

\noindent{\bf Results}.
Table \ref{table:multi-domain} presents the multi-domain few-shot learning results. It includes accuracies for both ID and OOD datasets, which refer to seen and unseen datasets during meta-training, respectively. The results show that our ASr achieves an average improvement of 1.2\% compared to the SOTA samplers, with a particularly significant improvement of nearly 3\% for out-of-domain tasks. It shows ASr's ability to balance distributions between classes and tasks. Meanwhile, previously proposed adaptive samplers that only rely on task diversity do not achieve such significant improvements. It indicates the importance of the three measurements we propose for task quality. These findings highlight the importance of sampling tasks in few-shot learning.

%%%%%%%%%%%%%%  Class-incremental few-shot learning %%%%%%%%%%%%%%%
\begin{table*}
\begin{center}
\caption{Class-incremental few-shot learning results (accuracy $\pm $ 95\% confidence interval) on miniImagenet. The values in the brackets indicate the performance improvement of our ASr compared to SOTA samplers.} 
\label{table:7}
\resizebox{\linewidth}{!}{
\begin{tabular}{l|cccccc}
\toprule[1.2pt]
\textbf{Model} & session 0 (50) & session 1 (60)  & session 2 (70) & session 3 (80) & session 4 (90) & session 5 (100)\\
\midrule
\textbf{AL-MML} & 71.42 $\pm$ 0.15 & 60.18 $\pm$ 0.24 (+1.89) & 44.16 $\pm$ 0.23 (+1.61) & 37.94 $\pm$ 0.17 (+0.93) & 30.45 $\pm$ 0.14 (+0.67) & 21.08 $\pm$ 0.14 (+0.44) \\
\textbf{MetaQDA}  & 66.49 $\pm$ 0.18 & 57.68 $\pm$ 0.17 (+1.96) & 50.20 $\pm$ 0.21 (+1.01) & 44.64 $\pm$ 0.17 (+0.75) & 39.47 $\pm$ 0.16 (+0.20) & 31.01 $\pm$ 0.22 (+0.39) \\
\bottomrule
\end{tabular}}
\end{center}
\end{table*}

%%%%%%%%%%%%%%%%% ablation study %%%%%%%%%%%%%%%%%%
\begin{figure*}[t]
     \centering
     \subfigure[miniImagenet]{\includegraphics[width=0.245\textwidth]{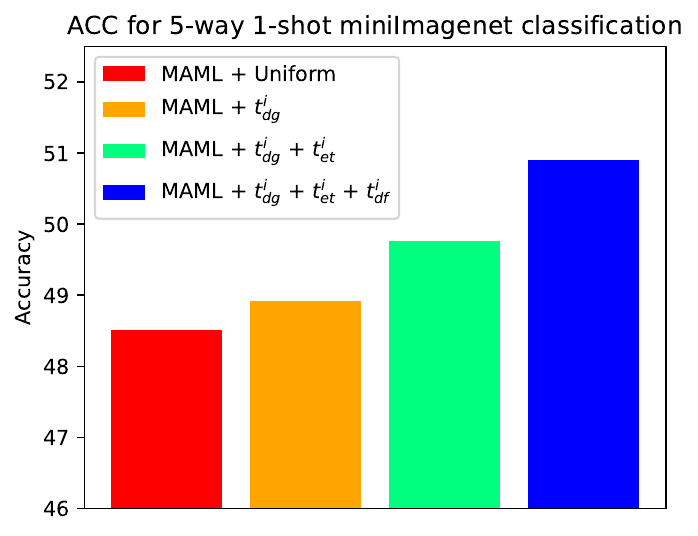}\label{fig:4.1}}
     \subfigure[Omniglot]{\includegraphics[width=0.245\textwidth]{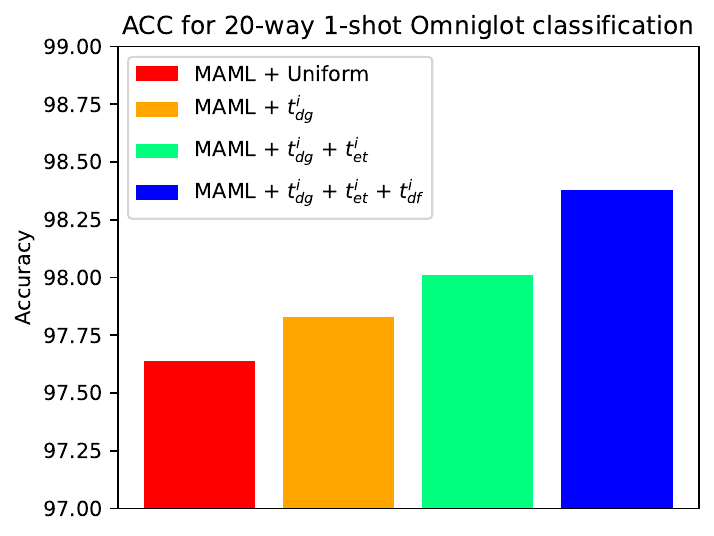}\label{fig:4.2}}
     \subfigure[Sinusoid]{\includegraphics[width=0.245\textwidth]{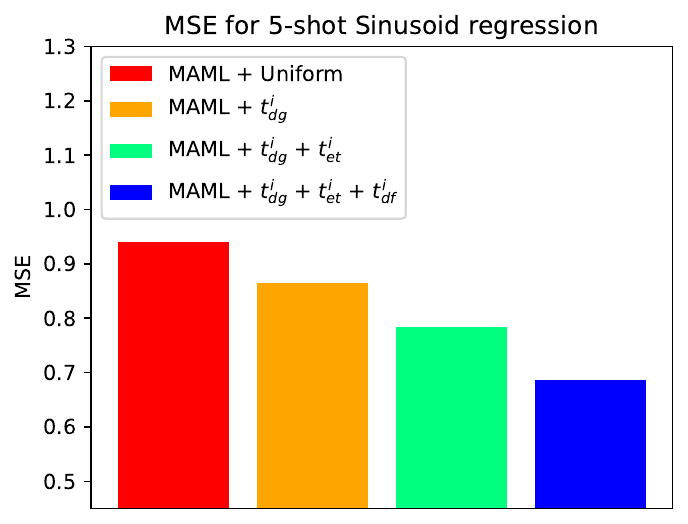}\label{fig:4.3}}
     \subfigure[Harmonic]{\includegraphics[width=0.245\textwidth]{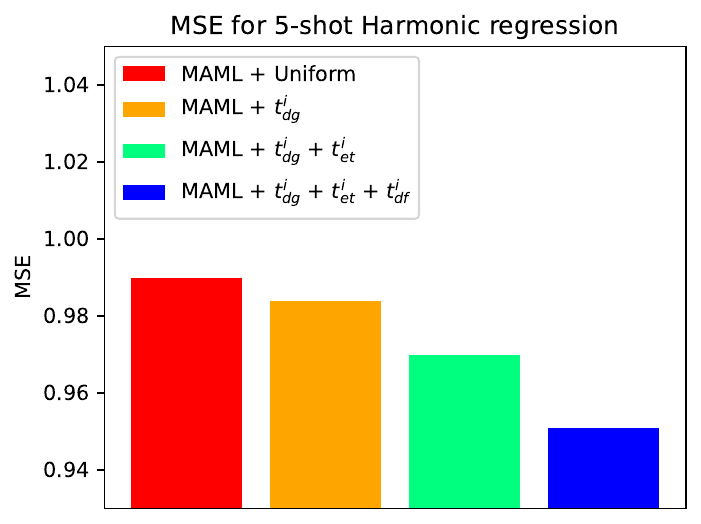}\label{fig:4.4}}
    \caption{Ablation study of ASr on 4 benchmarks. The meta-learning model in this experiment is MAML.}
    \label{fig:4}
\end{figure*}

\begin{figure}[t]
     \centering
     \subfigure[MAML]{\includegraphics[width=0.235\textwidth]{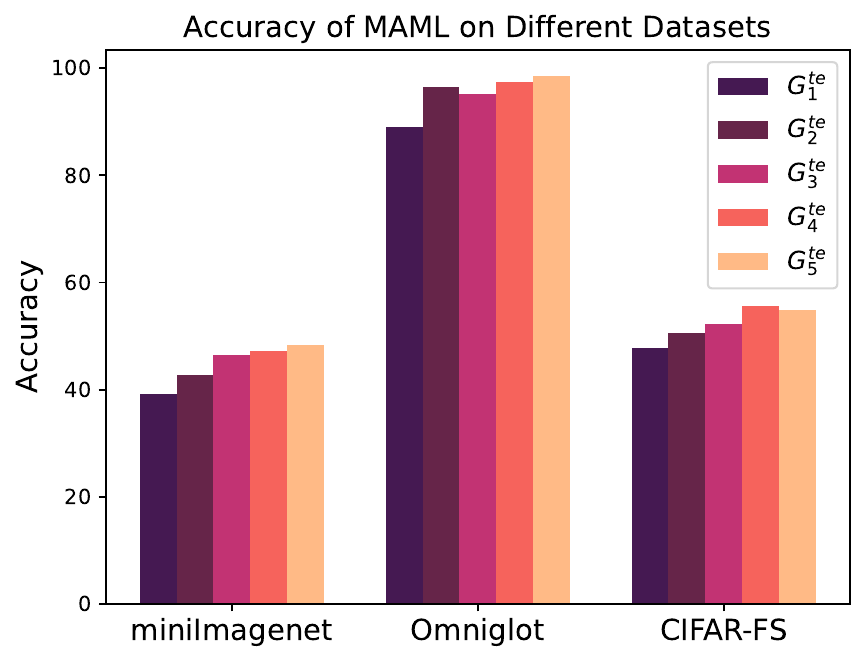}\label{fig:maml_te}}
     \subfigure[ProtoNet]{\includegraphics[width=0.235\textwidth]{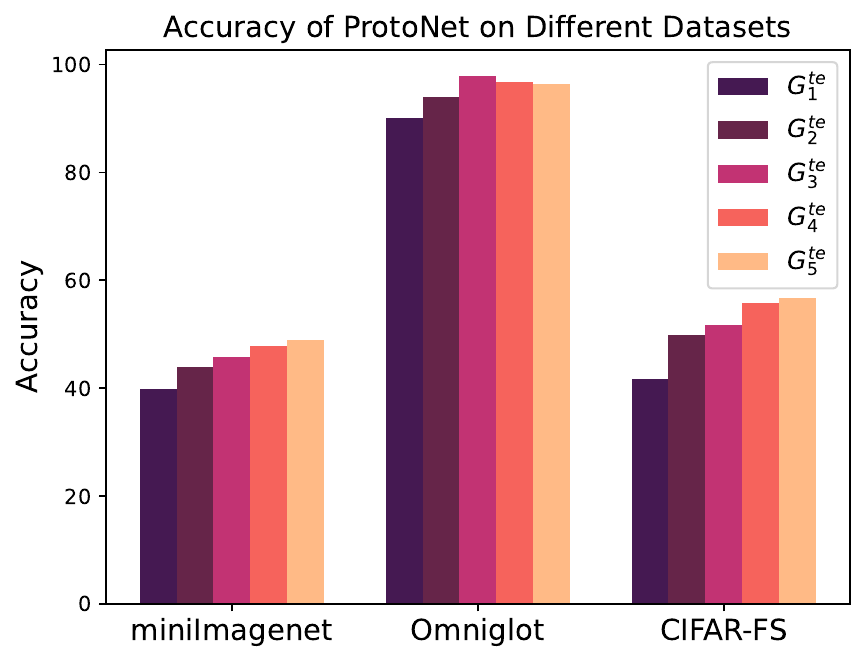}\label{fig:protonet_te}}
    \caption{The results of meta-learning models trained on tasks with different task entropy levels.}
    \label{fig:4.2.2}
\end{figure}

\begin{figure}[t]
     \centering
     \subfigure[MAML]{\includegraphics[width=0.235\textwidth]{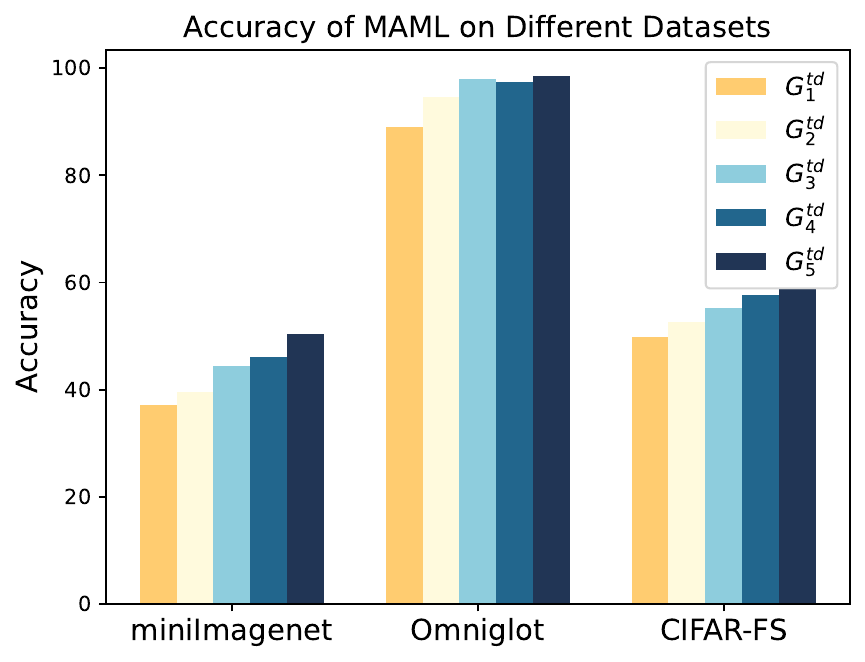}\label{fig:maml_td}}
     \subfigure[ProtoNet]{\includegraphics[width=0.235\textwidth]{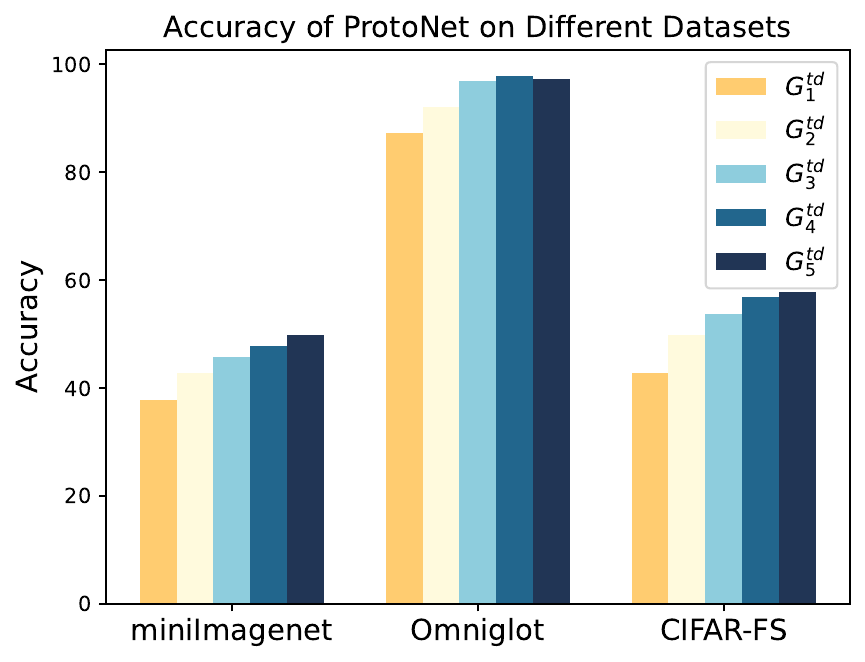}\label{fig:protonet_td}}
    \caption{The results of meta-learning models trained on tasks with different task difficulty levels.}
    \label{fig:4.2.3}
\end{figure}

\begin{figure}[t]
     \centering
     \subfigure[MAML]{\includegraphics[width=0.235\textwidth]{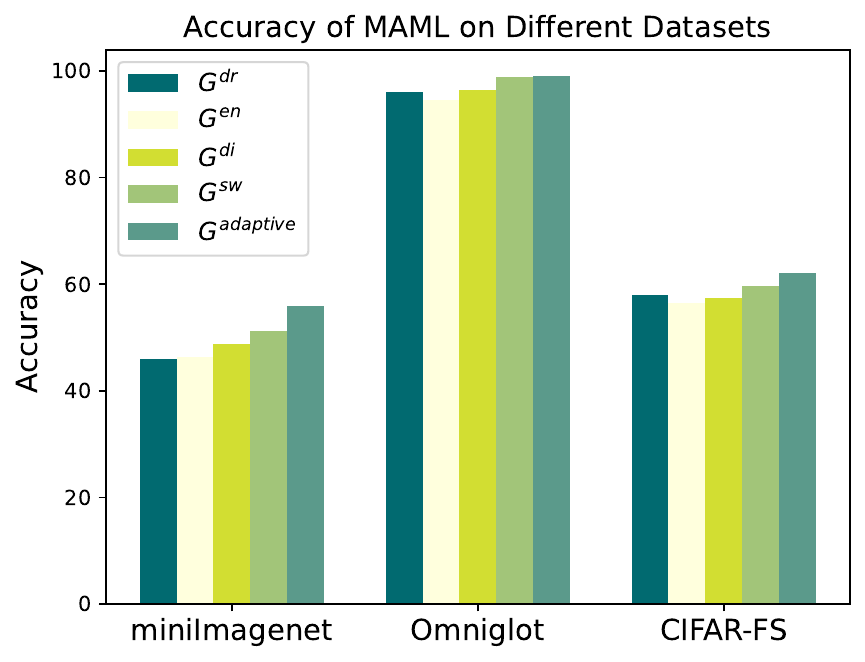}\label{fig:maml_fur}}
     \subfigure[ProtoNet]{\includegraphics[width=0.235\textwidth]{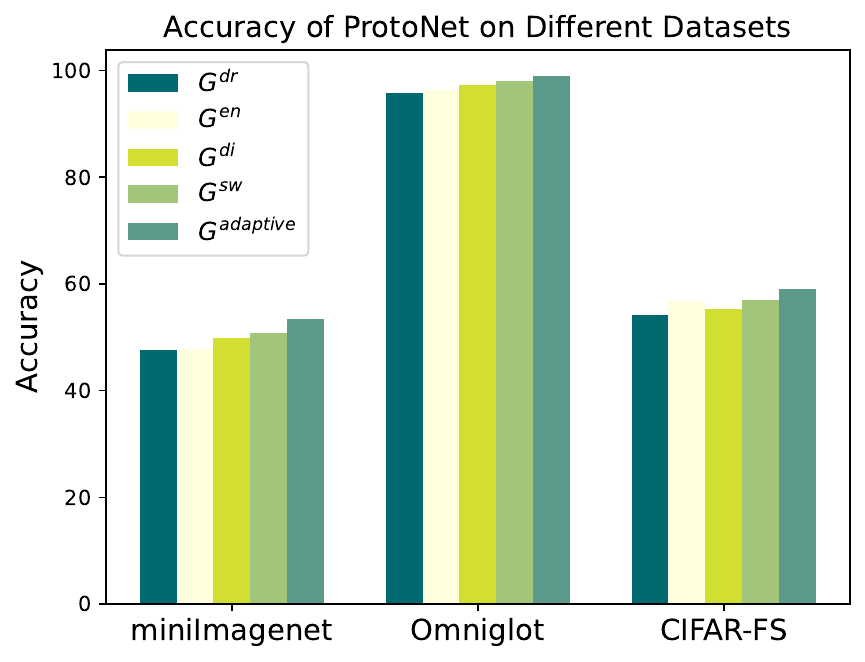}\label{fig:protonet_fur}}
    \caption{The performance of meta-learning models trained on tasks selected with different measurements.}
    \label{fig:4.2.4}
\end{figure}

%%%%%%%%%%%%%%%%%%%%%%%%%%%%%%%%%%
%            8.5
%%%%%%%%%%%%%%%%%%%%%%%%%%%%%%%%%%
\subsection{Few-shot Class-incremental Learning}
\label{sec:8.5}
\noindent{\bf Problem Setup}.
Few-Shot Class-Incremental Learning (FSCIL) requires models to incrementally learn new classes from limited data without forgetting the previously learned knowledge \citep{tao2020few, ren2019incremental}. In this experiment, we do not perform all data during meta-training, but instead use base sessions for training \citep{ren2019incremental}. Following \citep{tao2020few}, we continuously add 10 tasks to session 0 (originally containing 50 tasks) to simulate an incremental learning environment. We use MetaQDA \citep{zhang2021shallow} and AL-MML \citep{zhang2021few} as backbones to explore the effect of ASr on FSCIL.

\noindent{\bf Results}.
Table \ref{table:7} presents the results with different task samplers on FSCIL. Compared to the SOTA sampler, ASr can improve the performance to nearly 2\% when adding new classes, and this improvement persists as new classes are added. It demonstrates that ASr reduces knowledge forgetting, and obtains a better task sampling strategy to reduce the trade-off dilemmas.

%%%%%%%%%%%%%%%%%%%%%%%%%%%%%%%%%%
%            8.6
%%%%%%%%%%%%%%%%%%%%%%%%%%%%%%%%%%
\subsection{Ablation Study}
\label{sec:8.6}
To enhance the interpretability of ASr, we conduct ablation studies to evaluate the effectiveness of the three proposed measurements, i.e., task diversity $t_{dg}^i$, task entropy $t_{et}^i$, and task difficulty $t_{df}^i$, and their roles in our ASr. We evaluate the effects of these three measurements sequentially and in combination.

\subsubsection{Effect of Task Diversity}
\label{sec:8.6.1}
Regarding task diversity, we have demonstrated its effect in the previous experiments mentioned in Subsection \ref{sec:4.1}. It is worth noting that this effect is limited. To intuitively obtain the generalization situations under different task diversity levels, we analyze the training curve of MAML. We apply three samplers with different levels of task diversity, and evaluate their performance on different few-shot learning problems.

The results are shown in Figure \ref{fig:2}. It can be seen that all subfigures present two types of curves: the first type shows that the performance of the model initially increases, reaches a turning point and then decreases, while the second type shows that the performance of the model increases and then remains stable, but there is a large gap from the convergence results of other curves. The former mostly occurs in the groups corresponding to high-diversity samplers, that is, increasing diversity forces the model to over-fitting. On the other hand, the latter mostly occurs in the groups corresponding to low-diversity samplers, that is, under-fitting occurs when reducing diversity.

To sum up, task diversity affects model generalization ability, but over-constraining it may lead to a decrease in model generalization performance.

%%%%%%%%%%%%%%%%% further analysis %%%%%%%%%%%%%%%%%%
\begin{figure*}[t]
     \centering
     \subfigure[Standard]{\includegraphics[width=0.245\textwidth]{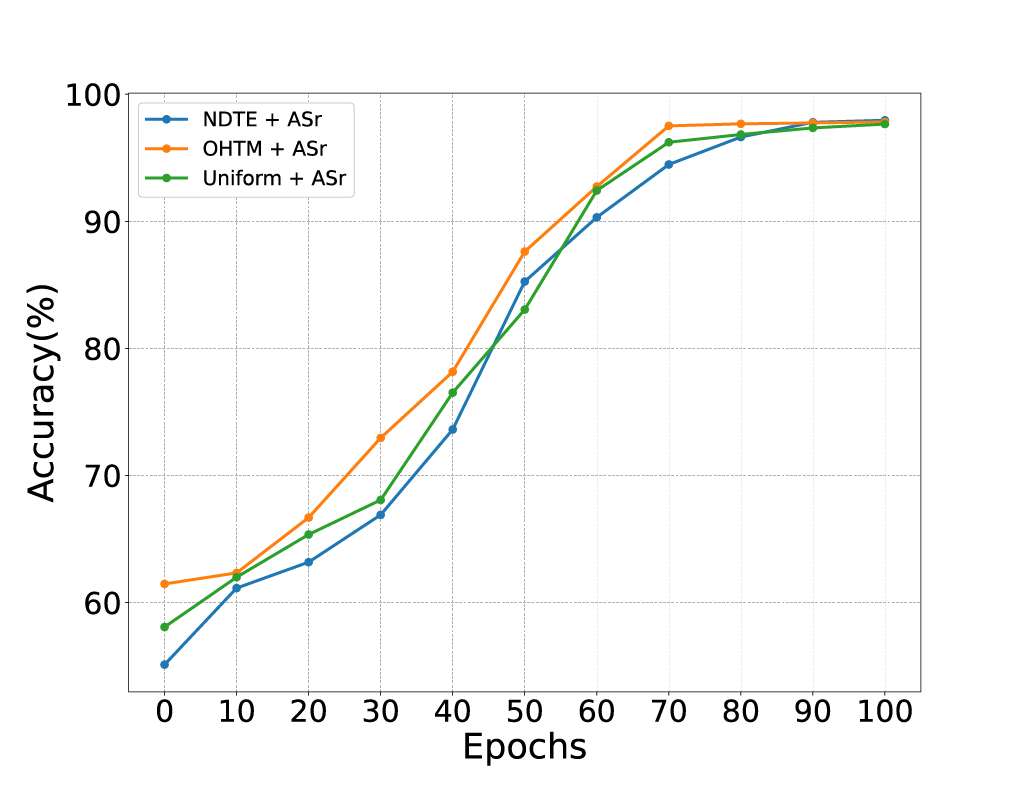}\label{fig:standard_after}}
     \subfigure[Cross-domain]{\includegraphics[width=0.245\textwidth]{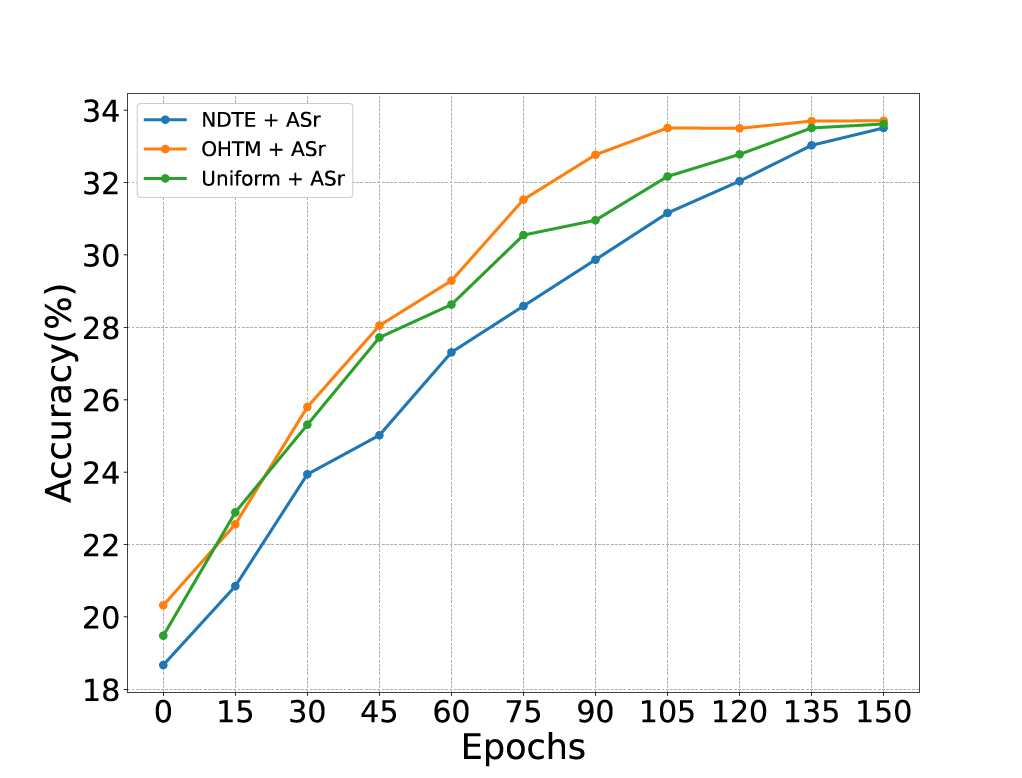}\label{fig:cross_domain_after}}
     \subfigure[Multi-domain]{\includegraphics[width=0.245\textwidth]{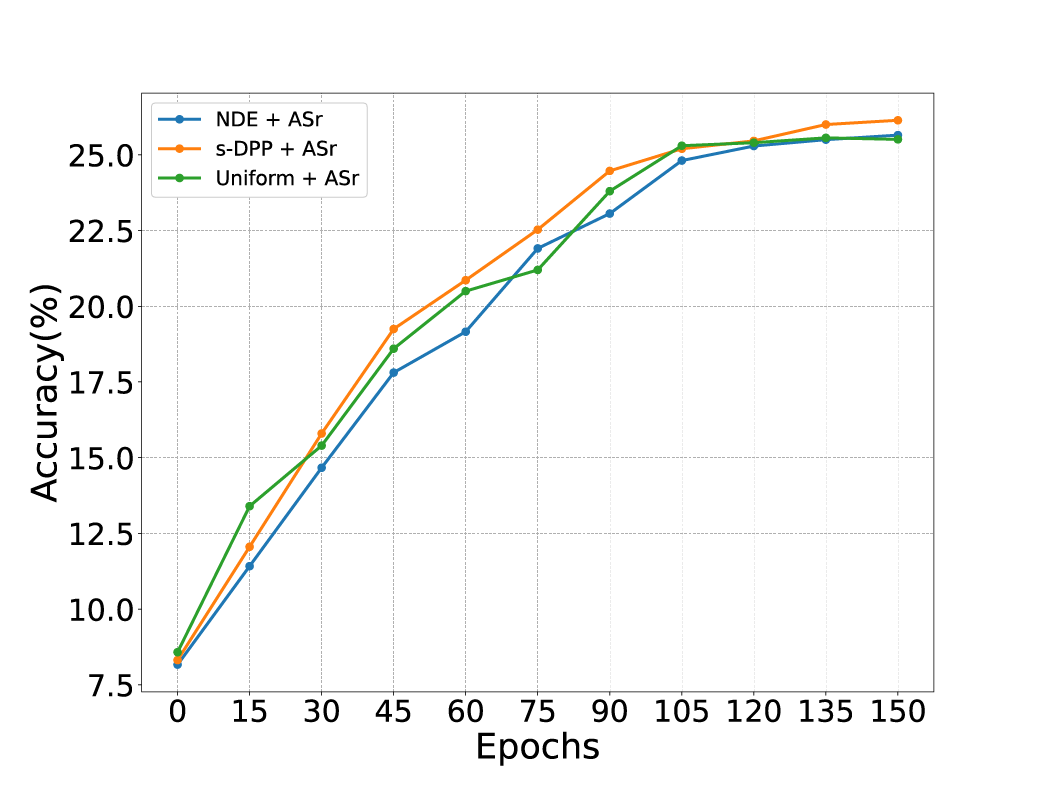}\label{fig:multi_domain_after}}
     \subfigure[Regression]{\includegraphics[width=0.245\textwidth]{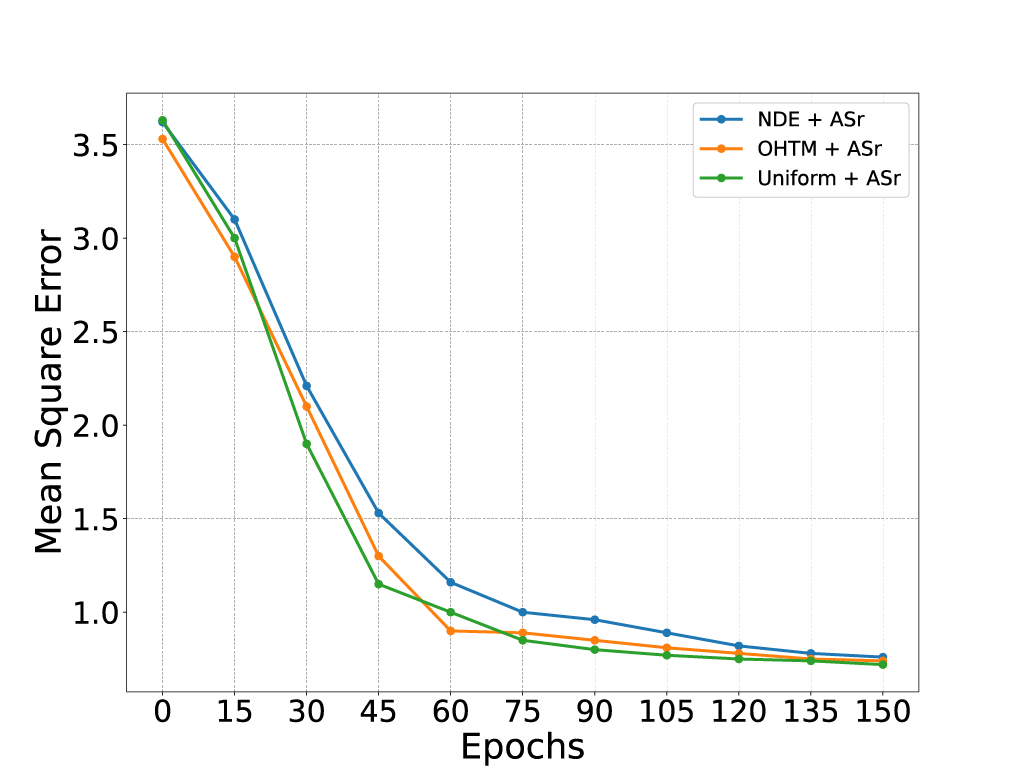}\label{fig:Regression_after}}
    \caption{The training process of MAML with ASr on various meta-learning settings.}
    \label{fig:5}
\end{figure*}

\begin{figure}
    \centering
    \includegraphics[width=0.35\textwidth]{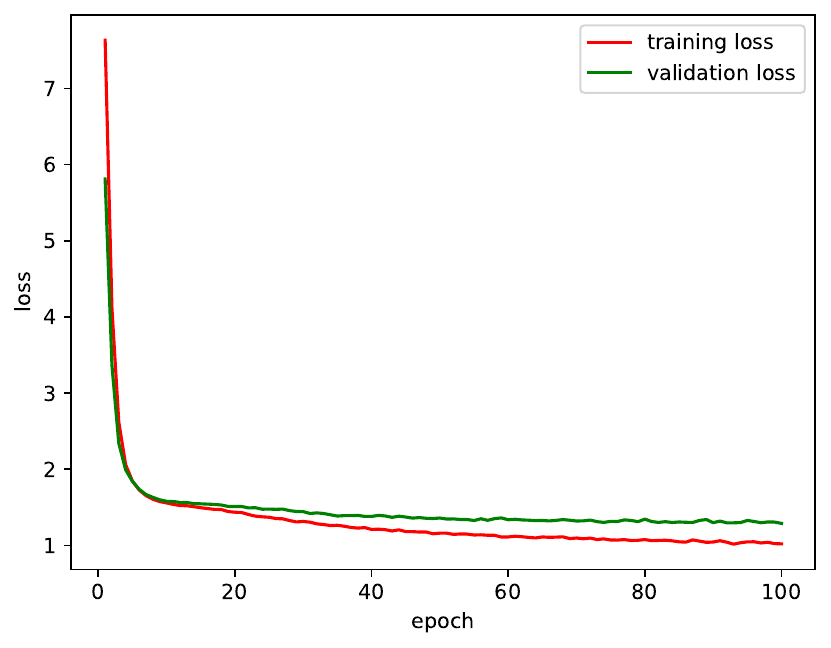}
    \caption{Loss of MAML with ASr on miniImagenet. }
    \label{fig:6}
\end{figure}

\begin{figure}
    \centering
    \includegraphics[width=0.45\textwidth]{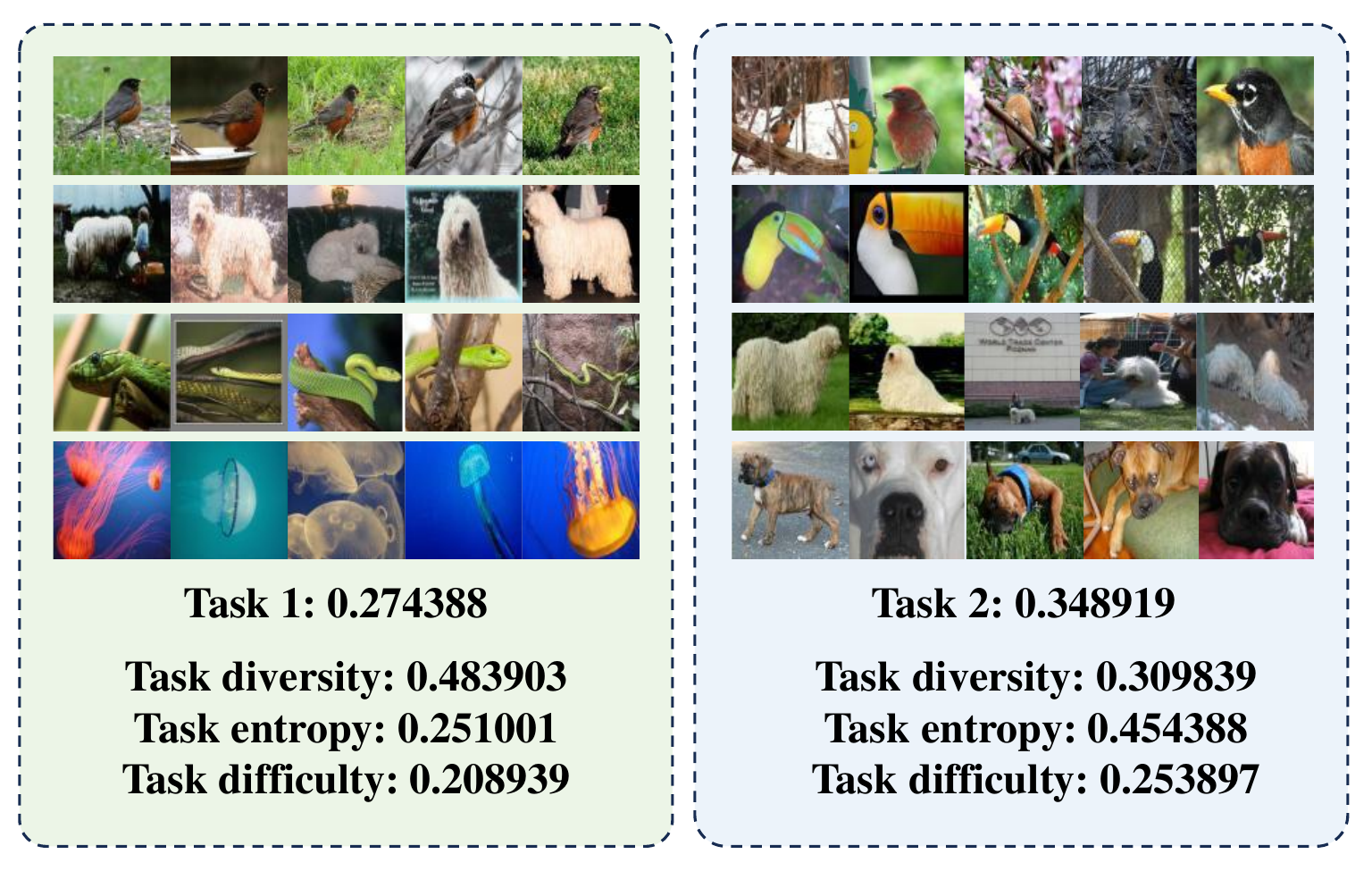}
    \caption{An example of pretext task sampling. The downstream task in this example is fine-grained classification.}
    \label{fig:7}
\end{figure}

\begin{figure}
    \centering
    \includegraphics[width=0.45\textwidth]{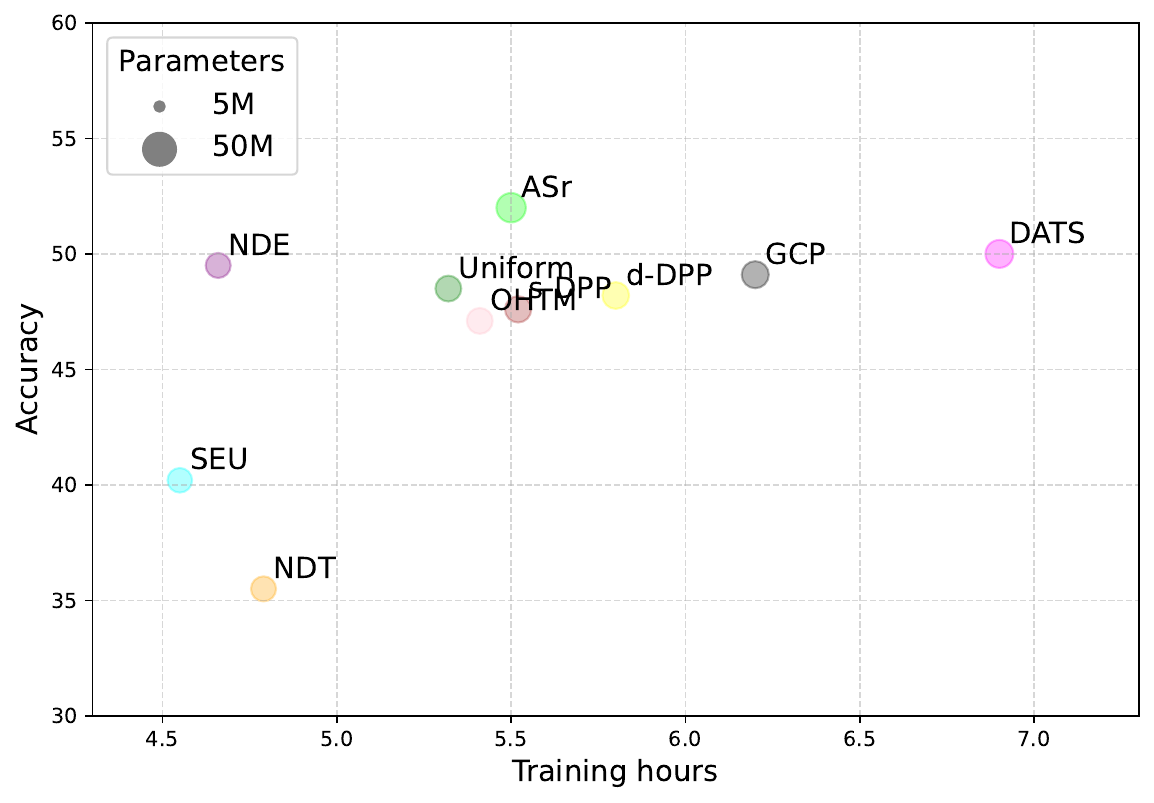}
    \caption{The trade-off performance of MAML using different task samplers. The horizontal axis represents the training time (hours), and the vertical axis represents the accuracy on miniImagenet. The area of the circle represents the model size of MAML with the corresponding labeled task sampler.}
    \label{fig:efficiency}
\end{figure}

\begin{figure}
    \centering
    \includegraphics[width=0.5\textwidth]{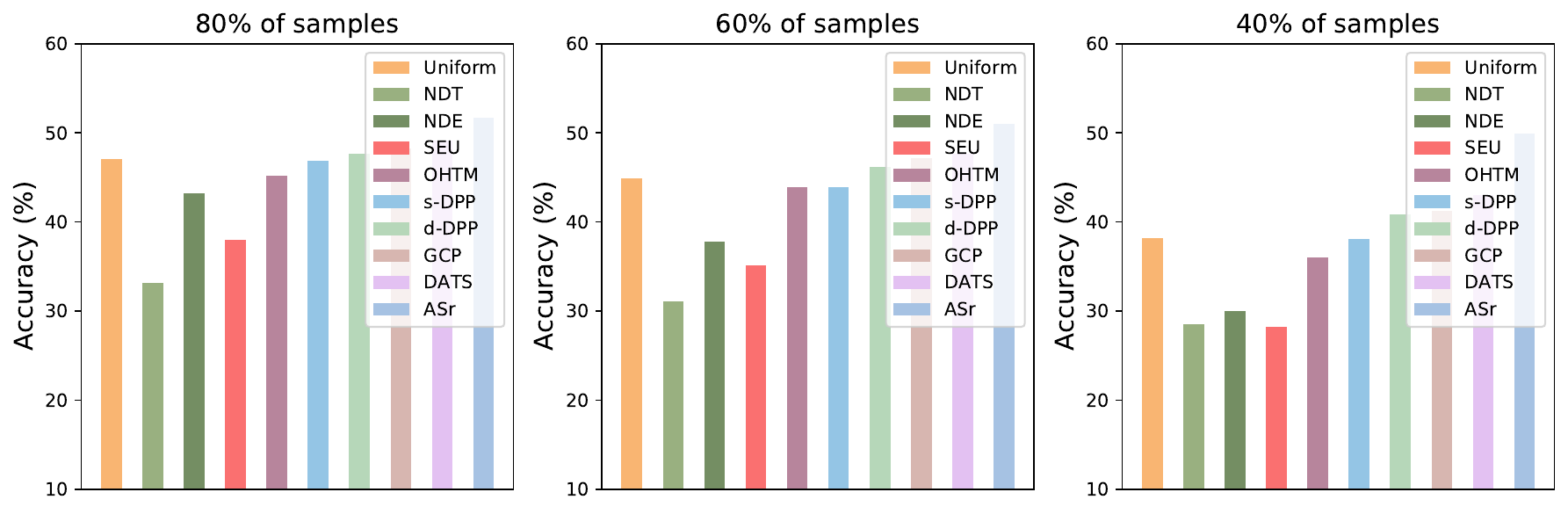}
    \caption{Performance of MAML using different samplers on small-scale datasets.}
    \label{fig:small}
\end{figure}

%%%%%%%%% 6.2.1 %%%%%%%%%%%%%%%
\subsubsection{Effect of Task Entropy}
\label{sec:8.6.2}
In order to evaluate the role of task entropy, we use MAML and ProtoNet as backbones, and build different training task groups for comparative experiments on three benchmark datasets: miniImagenet, Omniglot, and CIFAR-FS. Specifically, for each episode, we first randomly sample $N$ tasks to build a large candidate pool, where $N$ is related to the specific dataset, i.e., $N=40$ in miniImagenet. Then, we calculate the entropy of each task, and divide the tasks into five groups by task entropy scores in ascending order. Next, we use these groups to train the two backbones, and record the test results of the models. The tasks corresponding to the five groups are represented as $G_1^{te}\sim G_5^{te}$, where the task diversity score of $G_5^{te}$ is the highest.

The results are shown in Figure \ref{fig:4.2.2}. We observe that on multiple datasets, the higher the task entropy, the model achieves better results. For example, on miniImagenet and CIFAR-FS, the test accuracy of MAML trained on $G_5^{te}$ is the highest. However, task entropy is not always valid. For example, on Omniglot, ProtoNet trained on $G_5^{te}$ has a lower test accuracy than $G_3^{te}$. 

Therefore, task entropy affects model generalization, but hard to have positive effects on all domains.

%%%%%%%%% 6.2.1 %%%%%%%%%%%%%%%
\subsubsection{Effect of Task Difficulty}
\label{sec:8.6.3}
In this experiment, we use almost the same experimental setting as mentioned in Subsection \ref{sec:8.6.2}, and the only two differences are: (i) we divide the tasks of each episode by task difficulty in descending order, denoted as $G_1^{td}\sim G_5^{td}$; and (ii) we reduce the amount of sampled tasks for each episode to $N/2$ tasks, i.e., only 20 tasks are sampled for each episode in miniImagenet, and each group contains 4 tasks.

The results are shown in Figure \ref{fig:4.2.3}. The model trained on the group with the lowest task difficulty, $G_5^{td}$, achieves better results on most datasets, but not consistently. For example, the performance of the models trained on different groups fluctuates on Omniglot: MAML trained on $G_3^{td}$ outperforms MAML trained on $G_4^{td}$, while for ProtoNet, the model trained on $G_4^{td}$ achieves the best results.

Therefore, task difficulty can affect model generalization, but its impact fluctuates.

\subsubsection{Effect of Their Combination}
\label{sec:8.6.4}
\noindent{\bf The roles of the three measurements in ASr}.
From the above results, task diversity, task entropy, and task difficulty can all affect meta-learning generalization, but they are all limited. There is no one metric that can have a consistent impact on all domains. This is consistent with our theoretical analysis as mentioned in Subsection \ref{sec:6}, that is, all three are indispensable for task evaluation. To verify this, we sequentially remove these three measurements within ASr and evaluate their roles in ASr. We conduct experiments on two classification and two regression datasets, i.e., miniImagenet, Omniglot, Sinusoid, and Harmonic.

The results are depicted in Figure \ref{fig:4}. Each of the three components plays a positive role in enhancing the model's performance. Notably, the largest improvements are observed when these components are combined, i.e., the group ``MAML+$t_{dg}^i$+$t_{et}^i$+$t_{df}^i$" achieves the best results across all the datasets. Consequently, our design showcases robustness in its performance.

\noindent{\bf The advantage of adaptive sampling}.
To further demonstrate the superiority of our design, we compare the performance of simple weighted summation with that of our adaptive strategy. For 
$\mathcal{T}_i $, the score calculated through simple weighted summation is:
\begin{equation}
    S_{sw}(\mathcal{T}_i)= \tilde{t} _{dg}^i+ \tilde{t}_{et}^i-\tilde{t}_{df}^i
\end{equation}
where $\tilde{t} _{dg}^i$, $\tilde{t}_{et}^i$, and $\tilde{t}_{df}^i$ respectively represent the normalized ${t}_{dg}^i$, ${t}_{et}^i$, and ${t}_{df}^i$, and the value range is $\left [ 0,1 \right ] $.

Specifically, for each episode, we first randomly sample $N$ tasks to build a candidate pool and then calculate five indicators of all tasks, i.e., task diversity, task entropy, task difficulty, the calculation of $S_{sw}(\mathcal{T}_i)$, and our adaptive calculation. Next, we select five groups of tasks, each of which contains $N/5$ tasks with the best (highest/lowest) values in each indicator, denoted as $G^{dr}$, $G^{en}$, $G^{di}$, $G^{sw}$, and $G^{adaptive}$. $N$ is related to the specific dataset, i.e., $N=40$ in miniImagenet. There may be overlap in different groups of tasks. Finally, we record the test performance of MAML and ProtoNet trained on these five groups of tasks.

Figure \ref{fig:4.2.4} shows the comparison results. The model trained on $G^{sw}$ achieves better results than $G^{dr}$, $G^{en}$ and $G^{di}$ on multiple settings, but worse than $G^{adaptive}$. Therefore, our design of ASr is visionary.

%%%%%%%%%%%%%%%%%%%%%%%%%%%%%%%%%%
%            8.7
%%%%%%%%%%%%%%%%%%%%%%%%%%%%%%%%%%
\subsection{Further Analysis and Visualization}
\label{sec:8.7} 
Based on the above experimental results, we obtain four interesting conclusions: (i) different models exhibit varying tendencies towards sampling diversity strategies; (ii) the assumption that increasing task diversity will always improve model performance is limited; (iii) task diversity, task entropy, and task difficulty can well evaluate the quality of meta-learning tasks; and (iv) ASr improves accuracies across all datasets and meta-learning models. Furthermore, this section delves deeper into the intriguing issues about the effect of task sampling and our ASr.

%%%%%%%%% 8.7.1 %%%%%%%%%%%%%%%
\subsubsection{Generalization Analysis}
\label{sec:8.7.1}
In order to visualize the effect of ASr on improving generalization, we apply ASr to the four scenarios, i.e., standard few-shot classification, cross-domain few-shot classification, multi-domain few-shot classification, and few-shot regression. Specifically, we evaluate the effectiveness of ASr by recording its impact on addressing the issues of under-fitting and over-fitting shown in Figure \ref{fig:2}. We adopt the experimental settings as mentioned in Subsection \ref{sec:8.6.1}, but utilize ASr to assign weights to the selected tasks from the origin samplers. Subsequently, we document the performance during training as shown in Figure \ref{fig:5}. Compared to Figure \ref{fig:2}, we can observe that ASr eliminates the inflection point that exists after achieving optimal results. Moreover, ASr allows the model to converge to optimal performance. Therefore, ASr can eliminate both over-fitting and under-fitting caused by the original sampler, showing its effectiveness in enhancing generalization performance.

%%%%%%%%% 8.7.2 %%%%%%%%%%%%%%%
\subsubsection{Convergence Analysis}
\label{sec:8.7.2}
The convergence of the model is crucial for deep learning applications, as it enables the model to achieve higher performance faster and obtain a stable solution. In order to analyze the impact of introducing ASr, we visualize the training loss of MAML with ASr. Figure \ref{fig:6} shows that the training and validation losses of the model achieve good stability, and reach a level close to that of convergence within less than 10 epochs. Combining the experiment mentioned in Section \ref{sec:8.7.4}, the results demonstrate that ASr does not reduce the convergence efficiency of MAML, on the contrary, ASr reduces the loss of MAML and improves the performance.

%%%%%%%%% 8.7.3 %%%%%%%%%%%%%%%
\subsubsection{Visualization of Task Weights}
\label{sec:8.7.3}
ASr adapts the weights of each task according to three indicators: task diversity, task entropy, and task difficulty, and obtains the probability distribution that most suit the current scenario. We visualize the task weights to show the advantage of ASr in task sampling. In this experiment, we choose fine-grained classification as the downstream task to sample pretext tasks. 

Figure \ref{fig:7} shows an example of pretext task sampling. The values represent the weight of the tasks and the scores of task diversity, task entropy, and task difficulty. We can see that Task 1 on the left has higher task diversity and task difficulty, but lower task entropy than Task 2. If a simple weight summation is used, the sampler will tend to choose Task 1 with a larger value of ${t}_{dg}^i+{t}_{et}^i-{t}_{df}^i$. However, we find that using Task 2 as the training task can achieve better results through experiments. This may be because fine-grained classification requires detailed information, and Task 2 achieves better task entropy than Task 1 \citep{DBLP:conf/eccv/YangLWHGW18}. In Figure \ref{fig:7}, ASr can correctly judge this and give a higher probability to Task 2. Therefore, our design of ASr is foresighted. ASr does not simply sum up the three indicators we proposed, but integrates the optimization process into the meta-learning framework.

\subsubsection{Model Efficiency}
\label{sec:8.7.4}

According to the above analyses, ASr can perform well on any meta-learning framework with few update steps. In order to ensure the performance of ASr in practical applications, we evaluate the trade-off performance of the meta-learning model after introducing ASr. Specifically, we use MAML as the baseline and conduct experiments on the miniImagenet dataset. We record the training time and accuracy when the model reaches convergence using different samplers. Figure \ref{fig:efficiency} shows the trade-off performance. From the results, we can observe that: (i) ASr achieves the best performance improvement on meta-learning; (ii) the computing overhead introduced by ASr is similar to Uniform sampler, and is much lower than the previously proposed adaptive task samplers. This further proves that ASr is effective in advantages in practical applications.

\subsubsection{Effect on Small-scale Dataset}
\label{sec:8.7.5}

When data is scarce, additional model parameter adjustment will face various challenges, such as uneven data distribution \citep{daskalaki2006evaluation} and overfitting \citep{ying2019overview}. Taking this into account, we set ASr to be a lightweight network, and its learning process is embedded in the meta-learning process, so that it only needs one step of update to achieve superior performance improvement in few-shot learning. The experimental results described in Section \ref{sec:8.2}-\ref{sec:8.6} prove this. In this subsection, to further evaluate the effectiveness of ASr, we conduct experiment on small-scale datasets with uneven data distribution. Specifically, we adopt the same experimental settings as the image classification in Section \ref{sec:8.2}, but reduce the data candidate pool to the original 80\%, 60\%, and 40\%. Meanwhile, the samples and classes used for testing remain the same. Figure \ref{fig:small} shows the performance of different samplers on these smaller datasets. From the results, we can observe that ASr achieves stable performance improvements, which is difficult for other samplers to achieve. This illustrates the robustness of ASr and its practical significance to meta-learning.

%%%%%%%%%%%%%%%%%%%%%%%%%%%%%%%%%%%%%%%%%%%%%%%%%%%%%%%%%%%%%%%%%%%%%%%%%%%%%%%%%%%%%%%%%%%%%%%%%%%
%                                         7
%%%%%%%%%%%%%%%%%%%%%%%%%%%%%%%%%%%%%%%%%%%%%%%%%%%%%%%%%%%%%%%%%%%%%%%%%%%%%%%%%%%%%%%%%%%%%%%%%%%
\section{Conclusion}
\label{sec:9}
In this paper, we obtain three conclusions through empirical and theoretical analyses. Firstly, there are no universally optimal task sampling strategies to guarantee the performance of meta-learning models. Secondly, over-constraining task diversity may lead to under-fitting or over-fitting during meta-training. Lastly, the generalization performance of meta-learning models is affected by task diversity, task entropy, and task difficulty. Based on these findings, we propose the Adaptive Sampler (ASr). ASr is a plug-and-play task sampler that can be compatible with any meta-learning model. It dynamically adjusts task weights according to task diversity, task entropy, and task difficulty, obtaining the optimal probability distribution for meta-training tasks without introducing extra computational costs. We conduct extensive experiments on various meta-learning scenarios and frameworks, demonstrating the effectiveness and applicability of the proposed ASr. We also conduct ablation studies and further analyses to explore the interpretability of ASr.

\section*{Acknowledgements}
The authors would like to thank the anonymous reviewers for their valuable comments. This work was supported in part by the Postdoctoral Fellowship Program of CPSF (Grant No. GZB20230790), the China Postdoctoral Science Foundation (Grant No. 2023M743639), the Special Research Assistant Fund, Chinese Academy of Sciences (Grant No. E3YD590101), the Science and Technology Planning Project of Guangdong Province (Grant No. 2023A0505050111), and the Guangzhou-HKUST (GZ) Joint Funding Program (Grant No.2023A03J0008).

\section*{Data Availability}

The benchmark datasets can be downloaded from the literature cited in Subsection \ref{sec:8.2.1}.

\section*{Conflict of interest}

The authors declare no conflict of interest.

% \newpage
% \clearpage

% The conclusion goes here.

% or


% \appendices

% % you can choose not to have a title for an appendix
% % if you want by leaving the argument blank
% \section{}

% \ifCLASSOPTIONcompsoc
%   % The Computer Society usually uses the plural form
%   \section*{Acknowledgments}
% \else
%   % regular IEEE prefers the singular form
% \fi

\bibliographystyle{spbasic}      % basic style, 
\bibliography{reference.bib} 

\clearpage
	
\clearpage
\appendix

This appendix first provides the theoretical proofs of the theorems in Subsection \ref{sec:6}. Next, we introduce the details and experimental settings of the meta-learning models.

\section{Proofs}
\label{sec:appendix_1}
In this section, we provide the proofs of Theorem \ref{theorem:2}, Theorem \ref{theorem:3}, and Theorem \ref{theorem:4} in Appendix \ref{sec:appendix_1_1}, Appendix \ref{sec:appendix_1_2}, and Appendix \ref{sec:appendix_1_3}, respectively.

\vspace{0.1in} 
\noindent{\bf Notations}. Throughout this section, we use $Z_i$ to denote the representation of task $\mathcal{T}_i$, use $\textbf Z_i$ to denote the representation of the optimal $\mathcal{T}_i $, use $\mathbb{A}_+^n $, $\mathbb{R}_+ $, and $\mathbb{Z}_+ $ to denote the collection of $n \times n$ symmetric positive definite matrices, non-negative real numbers, and positive integers, respectively. The task $\mathcal{T}_i $ contains $n$ samples and $k$ classes, and class $j$ contains $n_j$ samples. The dimension of representation $Z_i$ is $d$, $Z_i\in \mathbb{R}^d$.

\subsection{Proof of Theorem 1}
\label{sec:appendix_1_1}
Theorem \ref{theorem:2}, also the Theorem \ref{theorem:5} mentioned below, gives the upper bound of task diversity. The condition for the upper bound being tight is consistent with \textit{Maximally Feature Space} in Corollary \ref{theorem:1}.

\begin{theorem}\label{theorem:5}
    \textit{Let $Z_i=\left [ Z_i^1,...,,Z_i^k \right ]\in \mathbb{R}^{d\times n}$ be the representation of task $\mathcal{T}_i $, which has $k$ classes and $n= {\textstyle \sum_{j=1}^{k}n_j} $ samples. For any representations $Z_i^j\in \mathbb{R}^{d \times n_j}$ of class $j$ and any $\sigma>0$, we have:}
    \begin{equation}
    \begin{array}{l}
    \frac{n}{2}\log\mathrm {det}(\mathcal{I} +\frac{d}{n\sigma ^2} Z_i^*Z_i) \\[10pt]
    \le \sum_{j=1}^{k}\frac{n}{2}\log\mathrm {det}(\mathcal{I} +\frac{d}{n\sigma ^2} {(Z_i^j)}^*{(Z_i^j)})
    \end{array}
    \end{equation}
    the equality holds if and only if:
    \begin{equation}
    \begin{array}{l}
    {(Z_i^{j_1})}^*(Z_i^{j_2})=0,\quad s.t.\quad 1\le j_1 \le j_2 \le k
    \end{array}
    \end{equation}
\end{theorem}

\noindent{\bf Proof}. According to \citep{chan2022redunet}, $\log\mathrm {det}(\cdot):\mathbb{A}_+^n\to \mathbb{R}  $ is strictly concave. For any $\beta \in (0,1)$ and $\left \{ Z_{j_1},Z_{j_2} \right \} \in \mathbb{A}_+^n $:
\begin{equation}
    \log\mathrm {det}((1-\beta )Z_{j_1}+\beta Z_{j_2})\ge (1-\beta )\log\mathrm {det}(Z_{j_1})+\beta \log\mathrm{det} (Z_{j_2})
\end{equation}
with equality holds if and only if $Z_{j_1}=Z_{j_2}$. Then for all $\left \{ A_a, A_b  \right \}\in \mathbb{A}_+^n$, we have:
\begin{equation}
    \log\mathrm(A_a)\le \log\mathrm(A_b)+\left \langle \bigtriangledown \log\mathrm(A_b),A_a-A_b \right \rangle 
\end{equation}
According to \citep{boyd2004convex}, let $A_b^{-1}=\bigtriangledown \log\mathrm(A_b)$ and $A_b^{-1}=(A_b^{-1})^*$, we have:
\begin{equation}\label{eq:a.1.1}
        \log\mathrm(A_a)\le \log\mathrm(A_b)+\mathrm {tr}(A_b^{-1}A_a)-n
\end{equation}
we now let:
\begin{equation}\label{eq:a.1.2}
\begin{array}{l}
    A_a= \mathcal{I}+\frac{d}{n\sigma ^2}(Z_i)^*Z_i\\[10pt]
    =\mathcal{I}+\frac{d}{n\sigma ^2}\begin{bmatrix}
 (Z_i^1)^*Z_i^1 & (Z_i^1)^*Z_i^2 & \dots  & (Z_i^1)^*Z_i^k \\
 (Z_i^2)^*Z_i^1 & (Z_i^2)^*Z_i^2 & \dots  & (Z_i^2)^*Z_i^k \\
 \vdots  & \vdots & \ddots  & \vdots \\
 (Z_i^k)^*Z_i^1 & (Z_i^k)^*Z_i^2 & \dots  & (Z_i^k)^*Z_i^k
\end{bmatrix}
\end{array}
\end{equation}
From the property of determinant for block diagonal matrix \citep{lu2018subspace}, we let:
\begin{equation}\label{eq:a.1.3}
\begin{array}{l}
    A_b= \mathcal{I}+\frac{d}{n\sigma ^2}(Z_i^j)^*Z_i^j\\[10pt]
    =\mathcal{I}+\frac{d}{n\sigma ^2}\begin{bmatrix}
 (Z_i^1)^*Z_i^1 & 0 & \dots  & 0 \\
 0 & (Z_i^2)^*Z_i^2 & \dots  & 0\\
 \vdots  & \vdots & \ddots  & \vdots \\
 0 & 0 & \dots  & (Z_i^k)^*Z_i^k
\end{bmatrix}
\end{array}
\end{equation}
Then, for $\mathrm {tr}(A_b^{-1}A_a)$:
\begin{equation}\label{eq:a.1.4}
\begin{array}{l}
\mathrm {tr}(A_b^{-1}A_a) = \mathrm {tr}\begin{bmatrix}
 \mathcal{I}  & \dots  & \chi  \\
 \vdots  & \ddots  & \vdots \\
 \chi  & \dots  & \mathcal{I}
\end{bmatrix}
=n
\end{array}
\end{equation}
bring Eq.\ref{eq:a.1.2}, Eq.\ref{eq:a.1.3}, and Eq.\ref{eq:a.1.4} back to Eq.\ref{eq:a.1.1}, we can get:
\begin{equation}
\begin{array}{l}
        \le \frac{n}{2}\log\mathrm {det}(\mathcal{I} +\frac{d}{n\sigma ^2} Z_i^*Z_i) \\[10pt]
    \le \sum_{j=1}^{k}\frac{n}{2}\log\mathrm {det}(\mathcal{I} +\frac{d}{n\sigma ^2} {(Z_i^j)}^*{(Z_i^j)})
\end{array}
\end{equation}
where the equality holds if and only if $A_a=A_b$, i.e., ${(Z_i^{j_1})}^*(Z_i^{j_2})=0,\quad s.t.\quad 1\le j_1 \le j_2 \le k$.

\subsection{Proof of Theorem 2}
\label{sec:appendix_1_2}
Theorem \ref{theorem:3}, also the Theorem \ref{theorem:6} mentioned below, gives that task entropy is maximized by the representations that are maximally discriminative between different classes and tight in each class. This is consistent with \textit{Maximally Discriminability} in Corollary \ref{theorem:1}, demonstrating that task entropy can well reflect intra-class compaction and inter-class separability.

\begin{theorem}\label{theorem:6}
    \textit{Let $Z_i=\left [ Z_i^1,..., Z_i^k \right ]$ be the representation of task $\mathcal{T}_i$, $\varsigma_j:= \left [ \varsigma_{1,j},..., \varsigma_{min(n_j,d),j} \right ] $ be the singular values of the representation $Z_i^j$ of class $j$, $\mathrm {C}_i=\left [ \mathrm {C}_i^1,...,\mathrm {C}_i^k \right ]$ is a collection of diagonal matrices, where the diagonal elements encode the $n$ samples into the $k$ classes. Given any $\epsilon >0$ and $d \ge d_j>0$, consider the optimization problem of task entropy:}
    \begin{equation}\label{eq:appen_2.1}
    \begin{array}{l}
    \underset{Z_i\in \mathbb{R}^{d\times n} }{\arg \max } (t_{et}^i)\\[10pt]
    s.t.\quad \left \| Z_i\mathrm {C}_i^j \right \|^2=\mathrm {tr}(\mathrm {C}_i^j ),\\
    \quad\qquad \mathrm {rank}(Z_i)\le  d_j, \forall j\in \left \{ 1,...,k \right \}  
    \end{array}
    \end{equation}
    Under the conditions where the error upper limit $\epsilon^4< \underset{j}{\min}\left \{ \frac{n_j}{n}\frac{d ^2}{d_j^2} \right \} $, and the dimension $d\ge  {\textstyle \sum_{j=1}^{k}d_j} $, the optimal solution $\textbf Z_i$ satisfies:
    \begin{itemize}
        \item Between-class: The representation $\textbf Z_i^{j_1}$ and $\textbf Z_i^{j_2}$ lie in orthogonal subspaces, i.e., $(\textbf Z_i^{j_1})^*(\textbf Z_i^{j_2})=0$, where $1\le j_1 \le j_2 \le k$.
        \item Within-class: each class $j$ achieves its maximal dimension $d_j$, i.e., $\mathrm {rank}(\textbf Z_i^j)= d_j$, and either $\left [ \varsigma_{1,j},..., \varsigma_{d_j,j} \right ]$ equal to $\frac{\mathrm {tr}(\mathrm {C}_i^j )}{d_j} $, or $\left [ \varsigma_{1,j},..., \varsigma_{d_j-1,j} \right ]$ equal to and have values larger than $\frac{\mathrm {tr}(\mathrm {C}_i^j )}{d_j} $.
    \end{itemize}
\end{theorem}

\noindent{\bf Proof}. Use singular value decomposition (SVD) to decompose $Z_i$ into $Z_i=U_i\Sigma_i V_i^*$, where $U_i$ and $V_i$ are unitary matrices, $\Sigma_i$ is a diagonal matrix, and its diagonal elements are singular values of $Z_i$. Since $\mathrm {rank}(Z_i)\le d_j, \forall j\in \left \{ 1,...,k \right \}$, assume the first $d_j$ diagonal elements of $\Sigma_i$ is not zero, and the subsequent diagonal elements are all zero. Therefore, $\Sigma_i$ will be:
    \begin{equation}
    \begin{array}{l}
    \Sigma_i=\left[\begin{array}{cc}
    \Sigma_{i,1} & 0 \\
    0 & 0
    \end{array}\right]
    \end{array}
    \end{equation}
Where $\Sigma_{i,1}$ is a diagonal matrix of $d_j\times d_j$, and its diagonal elements are $\varsigma_{1,j},..., \varsigma_{d_j,j}$. Similarly, for $U_i$ and $V_i$:
    \begin{equation}
    \begin{array}{l}
    U_i=\left[\begin{array}{cc}
    U_{i,1} & U_{i,2} \\
    \end{array}\right],\quad
    V_i=\left[\begin{array}{cc}
    V_{i,1} & V_{i,2} \\
    \end{array}\right]
    \end{array}
    \end{equation}
Among them, $U_{i,1}$ and $V_{i,1}$ are both matrices of $d\times d_j$, and $U_{i,2}$ and $V_{i,2}$ are both $ The matrix of d\times (n-d_j)$. Then, we get:
    \begin{equation}
    \begin{array}{l}
    Z_i=U_i\Sigma_i V_i^*\\[10pt]
    =\left[\begin{array}{cc}
    U_{i,1} & U_{i,2} \\
    \end{array}\right]\left[\begin{array}{cc}
    \Sigma_{i,1} & 0 \\
    0 & 0
    \end{array}\right]\left[\begin{array}{c}
    V_{i,1}^* \\
    V_{i,2}^*
    \end{array}\right]\\[10pt]
    =U_{i,1}\Sigma_{i,1} V_{i,1}^*
    \end{array}
    \end{equation}
Since $\mathrm {C}_i^j $ is a diagonal matrix, and only $n_j$ of its diagonal elements are 1, and the rest are 0, we have:
    \begin{equation}
    \begin{array}{l}
    \mathrm {tr}(\mathrm {C}_i^j )=n_j,\quad Z_i\mathrm {C}_i^j =U_{i,1}\Sigma_{i,1} V_{i,1}^ *\mathrm {C}_i^j
    \end{array}
    \end{equation}
Therefore, the constraint $\left \| Z_i\mathrm {C}_i^j \right \|^2=\mathrm {tr}(\mathrm {C}_i^j )$ can be equivalently written as:
    \begin{equation}
    \begin{array}{l}
    \left \| U_{i,1}\Sigma_{i,1} V_{i,1}^*\mathrm {C}_i^j  \right \|^2=n_j,\quad \forall j\in \left \{ 1,...,k \right \}    \end{array}
    \end{equation}
Without loss of generality, let $\textbf Z_i=\left [ \textbf Z_i^1 , ..., \textbf Z_i^k \right ] $ is the feature representation of the optimal task $\mathcal{T}_i $. To show that $\textbf Z_i^j,j\in \left \{ 1,...,k \right \} $, are pairwise orthogonal, suppose for the purpose of arriving at a contradiction that $(\textbf Z_i^{j_1})^*(\textbf Z_i^{j_2})=0$ for some $1\le j_1 \le j_2 \le k$. That is:
    \begin{equation}
    \begin{array}{l}
    \underset{Z_i\in \mathbb{R}^{d\times n} }{\arg \max } (t_{et}^i)\\[10pt]
    \le \frac{\mathrm {tr}(\mathrm {C}_i^j ) }{2n})\log\det(\mathcal{I} +\frac{d}{\mathrm {tr}(\mathrm {C}_i^j ) \epsilon^2}{{\textbf Z}_i^*}\mathrm {C}_i^j{{\textbf Z}_i})\\[10pt]
    s.t.\quad \left \| \textbf Z_i\mathrm {C}_i^j \right \|^2=\mathrm {tr}(\mathrm {C}_i^j ),\\
    \quad\qquad \mathrm {rank}(\textbf Z_i)=  d_j, \forall j\in \left \{ 1,...,k \right \}  
    \end{array}
    \end{equation}
According to the proof of Theorem \ref{theorem:5}, the strict inequality in \textit{Between-class} of Eq.\ref{eq:appen_2.1} holds for the optimal solution $\textbf Z_i$. On the other hand, since $\sum_{j=1}^{k} d_j\le n$, there exists $\left \{ Q_i^j\in \mathbb{R}^{d\times d_j}  \right \} _{j=1}^k$ such that the columns of the matrix $\mathcal{Q} $ are orthonormal. 

Since $Z_i=U_i\Sigma_i V_i^*$, $\Sigma_i\Sigma_i^*$ is a diagonal matrix and its diagonal element is $\varsigma _{l,j}^2 $, $\Sigma_{i,2}$ is a diagonal matrix of $d\times d$, and its diagonal element is $\varsigma _{l,j}^2 $, we have:
    \begin{equation}
    \begin{array}{l}
    Z_iZ_i^*=U_i\Sigma_i\Sigma_i^* U_i^*\\[10pt]
    =\left[\begin{array}{cc}
    U_{i,1} & U_{i,2} \\
    \end{array}\right]\left[\begin{array}{cc}
    \Sigma_{i,2} & 0 \\
    0 & 0
    \end{array}\right]\left[\begin{array}{c}
    U_{i,1}^* \\
    U_{i,2}^*
    \end{array}\right]\\[10pt]
    =U_{i,1}\Sigma_{i,2} U_{i,1}^*
    \end{array}
    \end{equation}
where the rank of $Z_iZ_i^*$ is equal to the rank of $\Sigma_{i,2}$, that is, $\mathrm {rank}(Z_iZ_i^*)=\mathrm {rank}(\Sigma_{i,2})=d_j $. This means that only $d_j$ of the eigenvalues of $Z_iZ_i^*$ are non-zero, and the rest are zero. Since $Z_iZ_i^*$ is a symmetric matrix, we can diagonalize it as:
    \begin{equation}
    \begin{array}{l}
    Z_iZ_i^*=Q_i\Lambda_i Q_i^*
    \end{array}
    \end{equation}
Since $(Z_i^{j_1})^* Z_i^{j_2}=V_{i,1}^{j_1*}\Sigma_{i,1}^{j_1*} U_{i,1}^{j_1* } U_{i,1}^{j_2}\Sigma_{i,1}^{j_2} V_{i,1}^{j_2}$, then:
    \begin{equation}
    \begin{array}{l}
    Z_i^{j_1}Z_i^{j_2*}\\[10pt]
    =Q_{i,1}^{j_1}\Lambda_{i,1}^{j_1} Q_{i,1}^{j_1*}Q_{i,1}^{j_2}\Lambda_{i,1}^{j_2} Q_{i,1}^{j_2*}\\[10pt]
    =Q_{i,1}^{j_1}\Lambda_{i,1}^{j_1}(Q_{i,1}^{j_1*}Q_{i,1}^{j_2})\Lambda_{i,1}^{j_2} Q_{i,1}^{j_2*}\\[10pt]
    =0
    \end{array}
    \end{equation}
That is, the matrices are pairwise orthogonal, i.e., $(\textbf Z_i^{j_1})^*(\textbf Z_i^{j_2})=0$, where $1\le j_1 \le j_2 \le k$.

Since $\det (Z_iZ_i^*)=\det (\Sigma_i\Sigma_i^*)=\det (\Sigma_{i,1}\Sigma_{i,1}^*)=\prod _{l=1} ^{d_j}\varsigma _{l,j}^2 $, we have:
    \begin{equation}
    \begin{array}{l}
    t_{et}^i=\sum _{j=1}^{k}\frac{n_j}{n}\log _2(\frac{n}{n_j})-\frac{n}{d}\log _2(\frac{n}{d})+\frac{2n}{d}\log _2(\prod _{l=1}^{d_j}\varsigma _{l,j})
    \end{array}
    \end{equation}
In order to maximize $t_{et}^i$, we need to maximize $\prod _{l=1}^{d_j}\varsigma _{l,j}$, subject to satisfying the constraints. Since $\left \| U_{i,1}\Sigma_{i,1} V_{i,1}^*\mathrm {C}_i^j \right \|^2=n_j$, we have:
    \begin{equation}
    \begin{array}{l}
    \left \| U_{i,1}\Sigma_{i,1} V_{i,1}^*\mathrm {C}_i^j  \right \|^2\\[10pt]
    =\mathrm {tr}(U_{i,1}\Sigma_{i,1} V_{i,1}^*\mathrm {C}_i^j V_{i,1}\Sigma_{i,1}^* U_{i,1}^*)\\[10pt]
    =\mathrm {tr}(\Sigma_{i,1} V_{i,1}^*\mathrm {C}_i^j V_{i,1}\Sigma_{i,1}^*)\\[10pt]
    =\sum _{l=1}^{d_j}\varsigma _{l,j}^2\left \| V_{i,l}^*\mathrm {C}_i^j V_{i,l} \right \|^2\\[10pt]
    =n_j
    \end{array}
    \end{equation}
Where $V_{i,l}$ represents the $l$th column of $V_{i,1}$. Since $\left \| V_{i,l}^*\mathrm {C}_i^j V_{i,l} \right \|^2\le 1$, we can get:
    \begin{equation}
    \begin{array}{l}
    \sum _{l=1}^{d_j}\varsigma _{l,j}^2\le n_j
    \end{array}
    \end{equation}
the equal sign holds true if and only if $\left \| V_{i,l}^*\mathrm {C}_i^j V_{i,l} \right \|^2= 1$. This means that $V_{i,l}$ must be an eigenvector of $\mathrm {C}_i^j $, and the corresponding eigenvalue is 1. Since only $n_j$ eigenvalues of $\mathrm {C}_i^j $ are 1 and the rest are 0. To maximize $\prod _{l=1}^{d_j}\varsigma _{l,j}$, we need to make $\varsigma _{l,j}$ as equal as possible, that is:
    \begin{equation}
    \begin{array}{l}
    \varsigma _{l,j}=n_j/d_j,\quad \forall l\in \left \{ 1,...,d_j \right \}
    \end{array}
    \end{equation}
Then, according to \citep{chan2022redunet}, the optimization problem in Eq.\ref{eq:appen_2.1} depends on $Z_i^j$ only through its singular values. We have:
    \begin{equation}
    \begin{array}{l}
    \left \| Z_i^j \right \|_{F}^2 =\sum_{l=1}^{\min\left \{ n_j,d \right \} } \varsigma _{l,j}^2\\[10pt]
    \mathrm {rank}(Z_i^j)=\left \| \varsigma_j  \right \|_0    
    \end{array}
    \end{equation}
Let $\varsigma_j^*:= \left [ \varsigma_{1,j}^*,..., \varsigma_{min(n_j,d),j}^* \right ]$ be an optimal solution to Eq.\ref{eq:appen_2.1}. Without loss of
generality we assume that the entries of $\varsigma_j^*$ are sorted in descending order. We define:
    \begin{equation}
    \begin{array}{l}
    f(z;d,\epsilon ,n_j,n)=\log (\frac{1 +\frac{d}{n \epsilon^2}{\varsigma_i}}{1 +\frac{d}{n_j \epsilon^2}{\varsigma_i^j}} )  
    \end{array}
    \end{equation}
Then apply the Lemma 13 in \citep{chan2022redunet} and conclude that the unique optimal solution to Eq.\ref{eq:appen_2.1}, we get:
\begin{itemize}
    \item $\varsigma_i^*=\left [ \frac{n_j}{d_j},...,\frac{n_j}{d_j}  \right ] $ or
    \item $\varsigma_i^*=\left [ \frac{n_j}{d_j},...,\frac{n_j}{{d_j}-1}, \varsigma_i^L \right ], \varsigma_i^L>0$
\end{itemize}

\subsection{Proof of Theorem 3}
\label{sec:appendix_1_3}
Theorem \ref{theorem:4}, also the Theorem \ref{theorem:7} mentioned below, gives the lower bound of task difficulty. The condition for the lower bound being tight is consistent with \textit{Minimally Effect Gap} in Corollary \ref{theorem:1}. This shows that task difficulty can well reflect causal invariance.

\begin{theorem}\label{theorem:7}
    \textit{The support set $\mathcal{D}_i^s$ and query set $\mathcal{D}_i^q$ are two different datasets of $\mathcal{T}_i$. For any $\mathcal{T}_i$ and $f$, we have:
    \begin{equation}\label{eq:a.3.1.1}
    \begin{array}{l}
    \sum\limits_{i,j} {\| {{\nabla _{x_{i,j}^s}}\mathcal{L}(\mathcal{D}_{i}^s,f) - {\nabla _{x_{i,j}^q}}\mathcal{L}(\mathcal{D}_{i}^q,f}) \|_2^2}\ge 0
    \end{array}
    \end{equation}
    the equality holds if and only if the gradients of the support set $\mathcal{D}_i^s$ and query set $\mathcal{D}_i^q$ are consistent: 
    \begin{equation}
        \sum\limits_{j}{{\nabla _{x_{i,j}^s}}\mathcal{L}(\mathcal{D}_{i}^s,f) = \sum\limits_{j}{\nabla _{x_{i,j}^q}}\mathcal{L}(\mathcal{D}_{i}^q,f}) 
    \end{equation}
    }
\end{theorem}

\noindent{\bf Proof}. 
we can deduce it directly from Eq.\ref{eq:a.3.1.1}:
    \begin{equation}
    \begin{array}{l}
    \sum\limits_{i,j} {\| {{\nabla _{x_{i,j}^s}}\mathcal{L}(\mathcal{D}_{i}^s,f) - {\nabla _{x_{i,j}^q}}\mathcal{L}(\mathcal{D}_{i}^q,f}) \|_2^2}\\[10pt]
    =\sum\limits_{i} {\| \sum\limits_{j}{{\nabla _{x_{i,j}^s}}\mathcal{L}(\mathcal{D}_{i}^s,f) - \sum\limits_{j}{\nabla _{x_{i,j}^q}}\mathcal{L}(\mathcal{D}_{i}^q,f}) \|_2^2}
    \end{array}
    \end{equation}
where $\left \| \cdot \right \| $ represents absolute value, always greater than zero, then:
    \begin{equation}
    \begin{array}{l}
    \sum\limits_{i} {\| \sum\limits_{j}{{\nabla _{x_{i,j}^s}}\mathcal{L}(\mathcal{D}_{i}^s,f) - \sum\limits_{j}{\nabla _{x_{i,j}^q}}\mathcal{L}(\mathcal{D}_{i}^q,f}) \|_2^2}\ge 0
    \end{array}
    \end{equation}
where the equality holds if and only if:
    \begin{equation}
        \sum\limits_{j}{{\nabla _{x_{i,j}^s}}\mathcal{L}(\mathcal{D}_{i}^s,f) = \sum\limits_{j}{\nabla _{x_{i,j}^q}}\mathcal{L}(\mathcal{D}_{i}^q,f}) 
    \end{equation}

In summary, the measurements we proposed can well evaluate the quality of meta-learning tasks.

\section{Meta-learning Models}
\label{sec:appendix_2}
In this section, we describe the details and experimental settings of the three types of meta-learning models mentioned in Subsection \ref{sec:8.2}.

\subsection{Overview}
To conduct a more comprehensive analysis of task diversity, we incorporate various meta-learning models. According to \citep{hospedales2021meta}, we classify them into three categories: 1) Optimization-based (i.e., MAML \citep{finn2017model}, Reptile \citep{nichol2018reptile}, and MetaOptNet \citep{lee2019meta}). These methods aim to learn a set of optimal initialization parameters that guide the model to quickly converge when learning new tasks. 2) Metric-based (i.e., ProtoNet \citep{snell2017prototypical}, MatchingNet \citep{vinyals2016matching}, and RelationNet \citep{sung2018learning}). These non-parametric methods are based on metric learning and are similar to nearest neighbor algorithms and kernel density estimation. 3) Bayesian-based models (i.e., CNAPs \citep{requeima2019fast}, SCNAP \citep{bateni2020improved}). These methods use conditional probability as the core of meta-learning computations and modify the classifier to pick up new classes using pre-trained networks.

With the development of meta-learning, many novel models and variants have emerged in recent years. Besides the above three types of frameworks, there are methods that particularly meta-learn features designed for few-shot learning and/or update features during meta-testing (i.e., SimpleShot \citep{wang2019simpleshot}, SUR \citep{mangla2020charting}, and PPA \citep{qiao2018few}).  We consider these models to be non-direct frameworks for the study of task sampling. For cross-domain analysis, we use Baseline++ \citep{chen2019closer} and S2M2 \citep{mangla2020charting} that use linear classifiers, and MetaQDA \citep{zhang2021shallow} which is a Bayesian meta-learning generalization of the classic quadratic discriminant analysis as base frameworks.

\subsection{Optimization-based}
\label{Optimization-based}

\subsubsection{MAML}
MAML \citep{finn2017model} is a meta-learning approach that is agnostic to specific models, making it compatible with any model trained using gradient descent. It can be applied to a variety of learning problems, with the explicit goal of training parameters that will generalize well to new tasks with only a small amount of training data and a few gradient steps.

For meta-learning with MAML, the method first initializes meta-parameters $\theta$ and samples tasks $\mathcal{T}_{i}$ from a task distribution $p(\mathcal{T})$. In the inner loop, the adaptive parameters are calculated for each task $\mathcal{T}_{i}$ using gradient descent, as follows: 
\begin{equation}
   \theta_{i}^{'}=\theta-\alpha \bigtriangledown_{\theta }\mathcal{L}_{\mathcal{T}_{i} }(f_{\theta })     
\end{equation}
In the outer loop, the meta-parameter $\theta$ is updated based on the accumulated gradient, as follows:
\begin{equation}
    \theta\gets \theta-\beta  \bigtriangledown_{\theta } {\textstyle \sum_{\mathcal{T}_{i}\sim p(\mathcal{T})}} \mathcal{L}_{\mathcal{T}_{i} }(f_{\theta_{i}^{'} })
\end{equation}
where $\alpha$ and $\beta$ are the step size hyperparameters of the inner loop and outer loop, respectively. 

In the experiments, we set the parameters as follows: the size of the running epoch is set to 150; batch size is 32 or 16; the meta-learning rate of Adam optimizer \citep{kingma2014adam} is 0.001; and the internal adaptation number is 1 with a step size of 0.4.

\subsubsection{Reptile}
Reptile \citep{nichol2018reptile} is a meta-learning approach that extends MAML by learning the initialization of neural network model parameters. It operates by repeatedly sampling a task, training it, and moving the initialization of the model parameters toward the trained weights for that task.

For meta-learning with Reptile, the method first initializes the meta-parameter $\theta$. For each iteration, a task $\mathcal{T}$ is sampled, corresponding to loss $\mathcal{L}_{\mathcal{T}}$ and a set of trained weights $\tilde{\theta}$. For a specific task, Reptile computes $\tilde{\theta}=U_{\mathcal{T}^{k} }(\theta ) $ denoting $k$ steps of SGD or Adam. The meta-parameter $\theta$ is then updated using the following equation:
\begin{equation}
    \theta\gets \theta+\epsilon (\tilde{\theta}-\theta )
\end{equation}
In the last step, instead of simply updating $\theta$, Reptile treats $\tilde{\theta}-\theta$ as a gradient and plugs it into an adaptive algorithm, such as Adam \citep{kingma2014adam}.

In the experiments, we set the parameters as follows: For \emph{mini}ImageNet, the running epoch size is set to 150, the batch size is 32, the learning rate is 0.01, the meta-learning rate is 0.001, and the internal adaptation number is 5. The inner loop uses the SGD optimizer, and the outer loop uses the Adam optimizer. For \emph{tiered}ImageNet, we increase the number of internal adaptations to 10. For Omniglot, the meta-learning rate is set to 0.0005, and the number of internal adaptations is the same as that on \emph{tiered}ImageNet, while only running for 100 epochs.

\subsubsection{MetaOptNet}
MetaOptNet \citep{lee2019meta} is a meta-learning model proposed for few-shot learning, which aims to learn an embedding model that generalizes well for novel categories under a linear classification rule. To achieve this, it utilizes the implicit differentiation of the optimality conditions of the convex problem and the dual formulation of the optimization problem.

The learning objective of MetaOptNet is to minimize the generalization error across tasks given a base learner $\mathcal{A}$ and an embedding model $\phi$. The generalization error is estimated on a set of held-out tasks. The choice of the base learner has a significant impact on the objective as it has to be efficient since the expectation has to be computed over a distribution of tasks. 

Formally, the learning objective is:
\begin{equation}
\label{eq:metaoptnet}
    \underset{\phi }{\min} \mathbb{E}_{\mathcal{T}}\left [\mathcal{L}^{meta}(\mathcal{D}^{test};\theta;\phi  )   \right ] 
\end{equation}
Once the embedding model $f_{\phi}$ is learned, its generalization is estimated on a set of held-out tasks (often referred to as a meta-test set):
\begin{equation}
    \mathbb{E}_{S}[\mathcal{L}^{meta}(\mathcal{D}^{test};\theta ,\phi) ], \quad where \quad \theta=\mathcal{A}(\mathcal{D}^{train};\theta)    
\end{equation}
The above equation is greatly affected by the selection of the base learner $\mathcal{A}$. The chosen base learner must be efficient as the expectation is calculated across a task distribution. In this study, we explore base learners that rely on multi-class linear classifiers, which can be expressed in a simplified form as follows:
\begin{equation}
     \theta = \mathcal{A}(\mathcal{D}^{train};\phi )=\arg \underset{\left \{ w_{k} \right \} }{\min} \underset{\left \{ \zeta_{i}  \right \} }{\min}\frac{1}{2}\sum_{k}\left \| w_{k} \right \|_{2}^{2}+C\sum_{n}\zeta _{n} \\
\end{equation}
\begin{equation}
    w_{y_{n}} \cdot f_{\phi}(x_{n})-w_{k}\cdot f_{\phi}(x_{n})\ge 1-\delta_{y_{n,k}}-\zeta _{n},\forall n,k
\end{equation}
where $C$ is the regularization parameter, and $\delta_{.,.}$ denotes the Kronecker delta function. The official repository trains the model using a 5-way 15-shot approach and evaluates it using a 5-way 1-shot approach. However, to ensure a fair and accurate comparison with other models as outlined in \citep{kumar2022effect}, we train and test the model using a 5-way 1-shot approach in this study. It is worth noting that our focus is on comparing the performance of different samplers for a given model, and the aforementioned difference in training and testing approaches would not affect our examination of task diversity in any way.

In the experiments, we set the parameters as follows: the size of the running epoch is set to 60, the batch size is 32 or 16, the learning rate is 0.01, and the meta-learning rate is 0.001. We use an SGD optimizer with a momentum of 0.9 and a weight decay of 0.0001 to make gradient steps.

\subsection{Metric-based Models}
\label{Metric-based}
\subsubsection{ProtoNet}
Prototypical networks (ProtoNet) \citep{snell2017prototypical} is a method proposed for few-shot classification tasks. This involves a classifier that must generalize to new classes not seen in the training set, with only a few examples available for each new class. ProtoNet addresses this problem by learning a metric space, where classification is performed by calculating distances to prototype representations of each class. Compared to other few-shot learning approaches, ProtoNet's simpler inductive bias is advantageous in the limited-data regime and achieves outstanding results.

To generate $M$-dimensional prototype representations $c_{k} \in \mathbb{R}^{M}$ for each class, ProtoNet employs an embedding function $f_{\phi}:\mathbb{R}^{D}\to \mathbb{R}^{M} $ with learnable parameters $\phi$. Each prototype is computed as the mean vector of the embedded support points belonging to its corresponding class:
\begin{equation}
    c_k=\frac{1}{\left | S \right | }\sum_{(x_i,y_i)\in S_{k}}f_{\phi}(x_i)  
\end{equation}

Once a prototype is constructed for each class, ProtoNet classifies query examples by determining the nearest prototype to them in the metric space using Euclidean distance. Specifically, the probability that a query example $x^*$ belongs to class $k$ is calculated as follows:
\begin{equation}
    p(y^*=k|x^*,S)=\frac{\mathrm {exp}(-\left \| g(x^*)-c_{k} \right \|_2^2 ) }{  {\textstyle \sum_{k^{'}\in \left \{ 1,...,N \right \} }\mathrm {exp}(-\left \| g(x^*)-c_{k^{'}} \right \|_2^2 )} } 
\end{equation}

In our experiments, we set the parameters as follows: For \emph{mini}ImageNet, Omniglot, and \emph{tiered}ImageNet, we use a batch size of 32 and run for 100 epochs in a 5-way-1-shot setting. However, we use a batch size of 16 rather than 32 in a 20-way-1-shot setting to accommodate the longer training time and memory constraints. We set the meta-learning rate to 0.001, use an Adam optimizer for gradient steps, and set the step size of the StepLR scheduler to 0.4 with a gamma value of 0.5.

\subsubsection{MatchingNet}
Matching Networks (MatchingNet), as described in \citep{vinyals2016matching}, leverages the concepts of metric learning based on deep neural features and the latest developments that enhance neural networks with external memories. This approach trains a network that maps a small labeled support set and an unlabelled example to its label, eliminating the need for fine-tuning to adapt to new class types.

The crucial point is that once trained, MatchingNet can generate sensible test labels for unseen classes without modifying the network. More precisely, MatchingNet aims to map a support set of $k$ image-label pairs, denoted as $S=\left \{ (x_i,y_i) \right \} ^k_{i=1}$, to a classifier $c_{S}(x^* )$. Given a test example $x^*$, the classifier produces a probability distribution over possible outputs $y^*$. The parametric neural network defined by $p$ is used to predict the appropriate label $\hat{y}$ for each test example $x^*$. MatchingNet assigns labels to each query example based on a cosine distance-weighted linear combination of the support labels:
\begin{equation}
    p(y^*=k|x^*,S)=\sum_{i=1}^{\left | S \right | }a(x^*,x_i) \Psi_{y_{i}=k} 
\end{equation}
where $a(\cdot,\cdot)$ denotes cosine similarity, $\Psi$ is the indicator function, and the output is softmax normalized over all support examples $x_i$.

In the experiments, we set the model parameters as follows: For standard few-shot learning under 5-way-1-shot settings, we run the epoch for 100 times with a batch size of 32. To make gradient steps, we use an Adam optimizer with a meta-learning rate of 0.001 and a weight decay of 0.0001. For training on CUB and Meta-Dataset under a 5-way 1-shot setting, we use the same parameters as the \emph{mini}ImageNet, except for the batch size and learning rate, which were set to 16 and 0.005, respectively.

\subsubsection{RelationNet}
Relation Network (RelationNet), presented in \citep{sung2018learning}, is a flexible and general framework for few-shot learning that is conceptually simple. The framework involves learning a deep distance metric to compare a small number of images within episodes, which is trained end-to-end from scratch.

The RelationNet framework comprises two modules: an embedding module, $f_{\varphi}$, and a relation module, $g_{\phi}$. The embedding module produces feature maps, $f_{\varphi}(x_{i})$ and $f_{\varphi}(x_{j})$, where $x_{i}$ and $x_{j}$ are samples in the support set $S$, and query set $Q$. These feature maps are combined using the operator $C(f_{\varphi}(x_{i}),f_{\varphi}(x_{j}))$ and fed into the relation module, $g_{\phi}$, for the next stage. The relation module produces a scalar value between 0 and 1 that represents the similarity between $x_{i}$ and $x_{j}$. The relation scores $r_{i,j}$ in $C$-way-1-shot settings ($C$ relation scores) are generated using the following equation:
\begin{equation}
    r_{i,j}=g_{\phi}(C(f_{\varphi}(x_i),f_{\varphi}(x_j))),i=1,2,...,C
\end{equation}
For $K$-shot settings, where $K>1$, we sum the embedding module outputs of all samples from each training class element-wise to form the feature map for that class. The model is trained using Mean Square Error (MSE) loss:
\begin{equation}
    \varphi,\phi=\underset{\varphi,\phi}{\arg\min}\sum_{i=1}^{m}\sum_{j=1}^{n}(r_{i,j}-1(y_i==y_j))^2
\end{equation}
Conceptually, this framework predicts relation scores, which can be considered a regression problem.

In the experiments, we set the model parameters as follows: For Omniglot, \emph{mini}ImageNet, and \emph{tiered}ImageNet under a 5-way-1-shot setting, the method is run for 100 epochs with a batch size of 32. An Adam optimizer is used to make gradient steps with a meta-learning rate of 0.001 and a weight decay of 0.0005. The same hyperparameters are used for training the model on Omniglot under a 20-way-1-shot setting.

In our experiments on Omniglot and \emph{mini}ImageNet, and \emph{tiered}ImageNet under a 5-way-1-shot setting, we run the epoch 100 times with a batch size of 32. We use an Adam optimizer to make gradient steps with a meta-learning rate of 0.001 and a weight decay of 0.0005. The same hyperparameters are used for training our model on Omniglot under a 20-way 1-shot setting.

\subsection{Bayesian-based Models}
\label{Bayesian-based}
\subsubsection{CNAPs}
The Conditional Neural Adaptive Processes (CNAPs) \citep{requeima2019fast} approach is designed to handle multi-task classification problems. It is based on a conditional neural process that employs an adaptation network to modulate the classifier's parameters based on the current task's dataset, without requiring additional tuning. This feature enables the model to handle a variety of input distributions.

The data for task $\tau $ includes a context set $D^\tau =\left \{ (x_{n}^\tau,y_{n}^\tau) \right \} _{n=1}^{N_{\tau}}$ and a target set $\left \{ (x_{m}^\tau,y_{m}^\tau) \right \} _{n=1}^{M_{\tau}}$. The former is with inputs and outputs observed while the latter is used to make predictions ($y^{\tau *}$ are only observed during training). CNAPs construct predictive distributions given $x^*$ as:
\begin{equation}
    p(y^*|x^*,\theta, D^\tau )=p(y^*|x^*,\theta ,\psi ^\tau =\psi_\phi(D^\tau ))
\end{equation}
where $\theta$ are global classifier parameters shared across tasks, $\phi$ are adaptation network parameters used in the function $\psi_\phi(\cdot)$ that acts on $D^\tau$, and $\psi ^\tau $ are local task-specific parameters produced by $\psi_\phi(\cdot)$. 

In the experiments, we set the model parameters as follows: In the standard few-shot learning setting, we run the epoch ten times with a batch size of 16 and a meta-learning rate of 0.005. In multi-domain few-shot learning, the meta-learning rate is set to 0.01.

\subsubsection{SCNAP}
Simple CNAPS (SCNAP) \citep{bateni2020improved}) is an architecture that performs better than CNAPs with up to 9.2\% fewer trainable parameters. It hypothesizes that a class-covariance-based distance metric, specifically the Mahalanobis distance, can be adopted into CNAPs. In contrast to CNAPs, SCNAP directly computes the conditional probability $p(\cdot)$ of a sample belonging to a class using a deterministic, fixed distance metric $d_k$, as follows:
\begin{equation}
\begin{split}
    p(y_i^*=k|f_{\theta}^{\tau}(x_i^*),S^\tau)=\mathrm {softmax}(-d_k(f_{\theta}^{\tau}(x_i^*)),\mu _k) \\
    d_k(x,y)=\frac{1}{2}(x-y)^T(Q_k^\tau)^{-1}(x-y) \qquad \quad
\end{split}
\end{equation}
where $Q_k^\tau$ is a covariance matrix specific to the task and class. 

The parameters of this model are consistent with CNAPs.

\end{document}